\documentclass{article}

\usepackage{microtype}
\usepackage{graphicx}
\usepackage{subcaption}
\usepackage{tabularx}
\usepackage{booktabs} 

\usepackage{hyperref}

\usepackage[preprint]{icml2026}

\usepackage{amsmath}
\usepackage{amssymb}
\usepackage{mathtools}
\usepackage{amsthm}

\usepackage{times}
\usepackage{latexsym}
\usepackage[T1]{fontenc}
\usepackage[utf8]{inputenc}
\usepackage{inconsolata}
\usepackage{algorithm}

\usepackage{tcolorbox}
\usepackage{natbib}
\usepackage[textsize=tiny]{todonotes}

\theoremstyle{plain}

\theoremstyle{definition}

\theoremstyle{remark}

\begin{document}

\icmltitlerunning{Tracing Multilingual Representations in LLMs with Cross-Layer Transcoders}

\twocolumn[
  \icmltitle{Tracing Multilingual Representations in LLMs with Cross-Layer Transcoders}



  \icmlsetsymbol{equal}{*}

  \begin{icmlauthorlist}
    \icmlauthor{Abir Harrasse}{equal,1,2,3}
    \icmlauthor{Florent Draye}{equal,1}
    \icmlauthor{Punya Syon Pandey}{4,5}
    \icmlauthor{Zhijing Jin}{1,4,5}
    \icmlauthor{Bernhard Schölkopf}{1,6}
  \end{icmlauthorlist}

  \icmlaffiliation{1}{Max Planck Institute for Intelligent Systems, Tübingen, Germany}
  \icmlaffiliation{2}{Mohammed VI Polytechnic University, Morocco}
  \icmlaffiliation{3}{Martian}
  \icmlaffiliation{4}{University of Toronto, Canada}
  \icmlaffiliation{5}{Vector Institute, Canada}
  \icmlaffiliation{6}{ELLIS Institute, Tübingen, Germany}

  \icmlcorrespondingauthor{Abir Harrasse}{aharrasse@cs.toronto.edu}
  \icmlcorrespondingauthor{Florent Draye}{florent.draye@tuebingen.mpg.de}
  \icmlcorrespondingauthor{Zhijing Jin}{zjin@cs.toronto.edu}


  \vskip 0.3in

]
\printAffiliationsAndNotice{}



%



\begin{abstract}
Multilingual Large Language Models (LLMs) can process many languages, yet how they internally represent this diversity remains unclear. Do they form shared multilingual representations with language-specific decoding, and if so, why does performance favor the dominant training language? To address this, we train models on different multilingual mixtures and analyze their internal mechanisms using \textit{Cross-Layer Transcoders} (CLT) and \textit{Attribution Graphs}. Our results reveal multilingual shared representations: the model employs highly similar features across languages, while language-specific decoding emerges in later layers.

Training models without English shows identical multilingual shared space structures. Decoding relies partly on a small set of high-frequency features in the final layers, which linearly encode language identity from early layers. Intervening on these features allows one language to be suppressed and another substituted. Finally, to explain non-English failures, we perform a \emph{Model-Diffing} experiment: underperformance arises from dim late-layer features, weak middle-layer clusters, and tokenizer bias toward English that forces early layers to specialize in word reassembly. Finetuning strengthens these features and their links, improving token assembly and language-specific decoding, providing a mechanistic explanation for multilingual gaps.

\end{abstract}

\begin{figure}[h!]
    \centering
    \includegraphics[width=\linewidth, trim=0 2cm 0 0, clip]{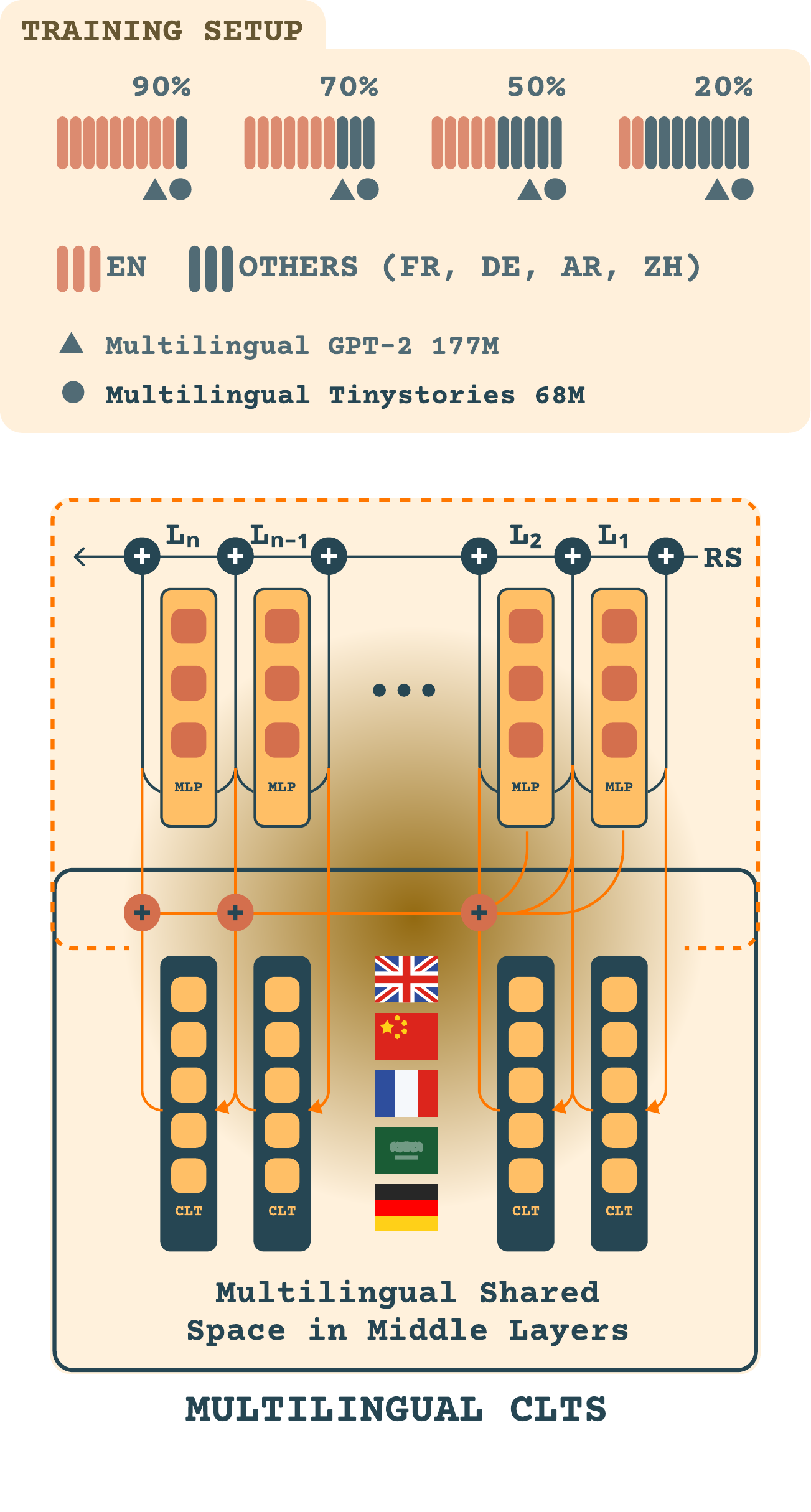}
    \caption{
   Cross-Layer Transcoders (CLTs) \cite{ameisen2025circuit} (bottom) reveal that multilingual LLMs (middle) consistently form a shared language-agnostic space in middle layers (gradient middle region) where all languages converge, regardless of English dominance in the training mixtures (top).}
    \label{fig:main_figure}
\end{figure}
\section{Introduction}

The inclusion of diverse languages in the large-scale training data of LLMs \cite{grattafiori2024llama3herdmodels, gemmateam2024gemma2improvingopen, brown2020languagemodelsfewshotlearners, chowdhery2022palmscalinglanguagemodeling} has led to remarkable multilingual capabilities \cite{shi2022language,conneau2020unsupervised,workshop2023bloom176bparameteropenaccessmultilingual} 
However, the underlying mechanisms driving their multilingual behavior remain poorly understood. For example, do LLMs form shared multilingual representations 
in their internal layers, so, why does performance still favor the dominant training language, such as English? Prior mechanistic investigations have yielded \textit{conflicting findings} about multilingual processing. While some work maps  inputs into English-aligned latent spaces and suggests that LLMs process non-English inputs through English-biased representations \cite{wendler2024llamasworkenglishlatent, schut2025multilingualllmsthinkenglish}, other studies find evidence of shared cross-lingual grammatical structures encoded in common feature directions \cite{brinkmann2025largelanguagemodelsshare}. Moreover, all existing approaches rely primarily on \textit{static} probing methods that cannot capture the \textit{dynamic transformation} of linguistic information across model layers.

To address the aforementioned limitations, we derive attribution graphs from our self-trained cross-layer transcoders \cite{ameisen2025circuit}, a recent advance in mechanistic interpretability, which can track feature interactions across layers and provide the first mechanistic account of how LLMs transition between shared and language-specific processing modes. 
Our results suggest that all languages have a shared representation in LLMs' internal layers, which we call as a ``pivot language.'' We also discover language-specific decoding in later layers to handle their multilingual output generation. 

Depending on our CLT-driven insights, we conduct two causal studies. First, we are able to change the output language by intervening on a small set of high-frequency language features in the final layers decoding from the pivot language to the target language.
Further, we investigate the causal effect of
training data composition by varying English-to-multilingual ratios. We analyze our re-trained models on different language ratios, and reveal how language distribution during pretraining shapes internal multilingual representations. Our contributions are as follows:

\begin{enumerate}

\item We find that multilingual LLMs form shared multilingual representations, relying on consistent internal circuits to process equivalent queries across languages. These structures emerge even without English or a dominant language, indicating that pivot mechanisms are an architectural property rather than a consequence of language dominance. While English-dominant models may still preferentially align with English, it is not strictly required for the emergence of shared representations.
\item We show that language decoding primarily depends on a small set of high-frequency language features in the late layers that read of information from early layer features and tokens embeddings. 
\item We identify and characterize language-specific failure modes, where models succeed in English but fail on equivalent inputs in other languages, linking these disparities to underlying mechanisms.

\item We release nine multilingual models\footnotemark[1] (two sizes × four data mixtures + non-English model), three finetuned models 
and their corresponding cross-layer transcoders (CLTs)\footnotemark[2], 
trained on balanced language distributions to support cross-lingual analysis and transfer. We also release our codebase\footnotemark[3].

\end{enumerate}

\footnotetext[1]{Models: \href{https://huggingface.co/collections/CausalNLP/multilingual-tinystories-6862b6562414eb84d183f82a}{Multilingual TinyStories} and \href{https://huggingface.co/collections/CausalNLP/multilingual-gpt2-models-684ad70e5fb3c84962306af3}{Multilingual GPT-2}}
\footnotetext[2]{CLTs: \href{https://huggingface.co/collections/CausalNLP/multilingual-clts}{Multilingual CLTs}}
\footnotetext[3]{Code: \href{https://github.com/abirharrasse/MultilingualCLTs}{https://github.com/abirharrasse/MultilingualCLTs}}

\section{Related Work}

\subsection{Multilingual Language Models}

Multilingual language models have evolved from early architectures like mBERT \cite{devlin2019bertpretrainingdeepbidirectional} to sophisticated models such as XLM-R \cite{conneau2020unsupervisedcrosslingualrepresentationlearning}, LLaMA \cite{touvron2023llamaopenefficientfoundation}, and Gemma \cite{gemmateam2024gemma2improvingopen}. Despite their impressive capabilities, systematic performance disparities persist across languages \cite{piresmultilingual, hu2020xtrememassivelymultilingualmultitask}. English consistently outperforms other languages, particularly for low-resource languages and complex reasoning tasks \cite{lauscher2020zeroherolimitationszeroshot, shi2022language}. Benchmarks like XTREME \cite{hu2020xtrememassivelymultilingualmultitask} and XNLI \cite{conneau2020unsupervisedcrosslingualrepresentationlearning} have documented these gaps across language families. The English-dominant training data composition has been hypothesized as a primary factor \cite{parrot}, though the underlying mechanisms driving these disparities remain poorly understood.

\subsection{Internal Representations and the Pivot Hypothesis}

Recent investigations have revealed that multilingual LLMs process non-English inputs through English-aligned latent representations. \citet{wendler2024llamasworkenglishlatent} used logit lens analysis to show that intermediate layers predict English tokens before target-language tokens. Similarly, \citet{schut2025multilingualllmsthinkenglish} demonstrated systematic bias toward English-like representations regardless of input language. These findings suggest models internally translate to English, perform reasoning, then translate back, a mechanism that has been leveraged through embedding alignment \cite{schuster2019crosslingual, Artetxe2018}, translation-based fine-tuning \cite{zhu2024extrapolating}, and romanization strategies for non-Latin scripts \cite{saji2025romanlensrolelatentromanization}.

Prompting strategies also exploit this English-pivot behavior, including direct prompt translation \cite{shi2022language, ahuja2023megamultilingualevaluationgenerative, etxaniz2023multilinguallanguagemodelsthink} and English chain-of-thought reasoning \cite{wei2022chain, vatsal2025multilingualpromptengineeringlarge}. While some grammatical features exhibit universal representations across languages \cite{brinkmann2025largelanguagemodelsshare}, lexical and semantic processing appears more language-specific \cite{chi2020finding}.

These observations align with the recently proposed ``Platonic hypothesis'' \cite{huh2024platonicrepresentationhypothesis}, which suggests that models converge toward abstract, unified representations capturing conceptual essences beyond surface forms. The shared multilingual space behavior, where models develop language-agnostic representations before language-specific decoding, provides some empirical support for this hypothesis in multilingual settings.

\subsection{Mechanistic Interpretability in LLMs}

Mechanistic interpretability seeks to understand transformer internals through circuit analysis \cite{olah2020zoom, elhage2021mathematical}. Early work identified specific circuits for behaviors like indirect object identification \cite{wang2022interpretabilitywildcircuitindirect} and induction heads \cite{olsson2022context}. Sparse Autoencoders (SAEs) decompose representations into interpretable features \cite{cunningham2023sparse, templeton2024scaling}, while transcoders directly model MLP outputs rather than autoencoding activations \cite{dunefsky2024transcodersinterpretablellmfeature}.

Cross-Layer Transcoders (CLTs) advance this approach by assigning distinct decoder matrices per downstream layer, significantly simplifying feature graphs and enabling tractable circuit analysis \cite{ameisen2025circuit, circuit-tracer}. However, CLT applications remain limited—primarily Anthropic's foundational work \cite{lindsey2025biology}, a ``greater-than'' mechanism study \cite{merullo2025replicating}, and recent open-source extensions \cite{lindsey2025landscape}. This scarcity reflects the considerable challenges in training CLTs at scale.

For multilingual models, mechanistic analyses are particularly sparse. Beyond \citet{wendler2024llamasworkenglishlatent}'s logit lens study and Anthropic's case study \cite{lindsey2025biology}, our work represents the first comprehensive CLT-based investigation of multilingual processing. Crucially, ensuring fair cross-language comparisons requires training CLTs on balanced multilingual distributions to avoid feature bias toward dominant languages.

\section{Experimental Setup and Methods}

\subsection{Mixture of multilingual LLMs}

We study the multilingual behavior of LLMs across five languages: English, German, French, Arabic, and Chinese. To investigate how training data composition affects internal representations, we train four GPT-2 style models (12 layers, $\sim$177.6M parameters) on a realistic web mixture (OpenWebText + FineWeb2 \cite{penedo2025fineweb}), and four tiny-stories style models (4 layers and $\sim$68.5M parameters) on a controlled narrative dataset (TinyStories translated into all five languages). Finally, we include an \textit{off-the-shelf} LLaMA-3.2-1B model \cite{meta-llama32-1b} for comparison. 

To facilitate our training data manipulation study, we investigate training data settings where all languages are evenly distributed vs those with one dominant language. To investigate the effect of a dominant language, we vary the English proportion in the data from dominant (90\%) to balanced (20\%), with the remaining data evenly split across the other four languages. All models use debiased tokenizers trained on evenly distributed five-language data, so that differences in internal representations reflect the training distribution rather than tokenization artifacts. Figure~\ref{fig:main_figure} illustrates the training configurations.


\subsection{Cross-Layer Transcoders (CLTs)}

To probe internal multilingual mechanisms, we employ CLTs, which map activations between layers to reveal how semantic representations evolve during processing. A CLT consists of neurons (features) divided into encoder and decoder components.

Formally, to run a cross-layer transcoder, let $\mathbf{h}_\ell \in \mathbb{R}^{d_\text{model}}$ be the input to the MLP at layer $\ell$ for single token position. We define
\begin{equation}
\mathbf{z}_{\ell} = \text{ReLU}(\mathbf{W}_\text{enc}^\ell \mathbf{h}_\ell + \mathbf{b}_\text{enc}^\ell ) \in \mathbb{R}^{d_\text{features}}
,
\end{equation}
where $\mathbf{W}_\text{enc}^\ell \in \mathbb{R}^{d_\text{features} \times d_\text{model}}$ is the encoder weight matrix and $\mathbf{b}_\text{enc}^\ell \in \mathbb{R}^{d_\text{features}}$ is the encoder bias. The CLTs then reconstructs the MLP output at layer $\ell'$ as
\begin{equation}
\hat{\mathbf{m}}_{\ell'} = \sum_{\ell \leq \ell'} \mathbf{W}_{\text{dec}}^{\ell \to \ell'}  \mathbf{z}_{\ell} +
\mathbf{b}_\text{dec}^{\ell'}
,
\end{equation}
where $\mathbf{W}_{\text{dec}}^{\ell \to \ell'} \in \mathbb{R}^{d_\text{model} \times d_\text{features}}$ is the decoder matrix from layer $\ell$ to $\ell'$.

Following Anthropic guidelines \cite{ameisen2025circuit}, the final training objective is
\begin{align}
\mathcal{L} = 
& \underbrace{\sum_{\ell'} \left\| \hat{\mathbf{m}}_{\ell'} - \mathbf{m}_{\ell'} \right\|_2^2}_{\text{MSE reconstruction}} \nonumber \\
&+ \lambda_{0} \underbrace{\sum_{\ell} \tanh\!\big(C \, (\mathbf{z}_\ell \odot \|\mathbf{W}_{\text{dec}}^{\ell}\|)\big)}_{\text{$L_0$ sparsity}} \nonumber \\
&+ \lambda_{\text{df}} \underbrace{\sum_{\ell} 
   \mathrm{ReLU}\!\Big(\exp(\tau) - \mathbf{h}_\ell^{\text{pre}}\Big)\|\mathbf{W}_{\text{dec}}^{\ell}\|}_{\text{dead feature penalty}}
   ,
\end{align}
where $\mathbf{W}_{\text{dec}}^{\ell}$ 
are concatenated decoder weights for layer $\ell$, $\mathbf{h}_\ell^{\text{pre}}$ are pre-activation values, 
$\tau$ is a threshold parameter, and $C$ is a scaling constant. 
$\lambda_{0}$ and $\lambda_{\text{df}}$ control the strength of the sparsity 
and dead-feature regularization terms.

We train CLTs on activation vectors sampled uniformly across languages until convergence. Training details, hyperparameters, and performance metrics appear in Appendix~\ref{appendix:clt_training}.


\subsection{Graph attribution}

Following \citet{ameisen2025circuit}, we compute the attribution score between every
feature $n$ at layer $\ell$, position $k$, and feature $n'$ at layer $\ell'$, 
position $k'$, as
\begin{equation}
a_{\ell, k, n}^{\ell', k', n'} 
= \sum_{\ell \leq s \leq \ell'} 
f_{k,n}^{\ell \to s} \;
J_{s, k}^{\ell', k'} \;
g_{k',n'}^{\ell'}
,
\end{equation}
where $f_{k,n}^{\ell \to s}$ denotes the vector for feature $n$ in the decoder 
matrix projecting from layer $\ell$ to $s$, 
$J_{s, k}^{\ell', k'}$ is the Jacobian between the MLP output at $(\ell, k)$ 
and the MLP input at $(\ell', k')$, computed during a forward pass where 
nonlinearities (normalization layers, attention computations, and MLPs) are 
frozen using stop-gradient operations, and 
$g_{k',n'}^{\ell'}$ is the corresponding encoder feature at layer $\ell'$ and 
position $k'$. The graph is then pruned to retain only features that cumulatively account for $80\%$ of the effect on the final logit, and edges that cumulatively account for $95\%$ of the edge effect on the final logit. We use the circuit-tracer library \cite{circuit-tracer}. 

\subsection{Multilingual Score}
\label{subsec:multilingual_score}

For each CLT feature $f$, we compute the multilingual scores on both the top 100 most activated sequences for that feature and on a general sample of 600k sequences balanced across the five languages (each of length 16). If a feature is active for at least one token in a sequence, it is considered active for that sequence. Since each sequence corresponds to a specific language, this allows us to compute a normalized activation distribution:
\begin{align}
p_l(f) = \frac{A_l(f)}{\sum_{l'} A_{l'}(f)}
,
\end{align}
where \(A_l(f)\) is the number of sequences in language \(l\) for which feature \(f\) is active. We then define the multilingual score for feature \(f\) as the entropy of this distribution:
\begin{align}
H(f) = -\sum_{l=1}^{L} p_l(f) \log p_l(f), \quad L=5
,
\end{align}
which measures multilinguality: low \(H(f)\) indicates language-specific features, while high \(H(f)\) indicates multilingual features. We normalize the entropy scores to have scores between 0 and 1.

\section{Results}

We organize our results around three core questions that progressively unpack how multilingual representations emerge and operate in large language models:
\begin{enumerate}
    \item Do models form shared multilingual representations or a pivot language across data mixtures? (Section \ref{sec:pivot_language})
    \item  How do models decode language identity and route information to the correct output language? (Section \ref{sec:language_decoding})
    \item Why do models still fail in non-English settings despite shared representations? (Section \ref{sec:why_models_fail})
\end{enumerate}

\subsection{Emergence of a Pivot Language}
\label{sec:pivot_language}





\subsubsection{Generalization Under Data Imbalance}
\label{sec:generalization_imbalance}

We begin by assessing whether models trained on imbalanced multilingual mixtures are still able to generalize across languages.  
Figure~\ref{fig:combined} reports the validation cross-entropy losses for models trained with varying proportions of English tokens, including extreme settings where English accounts for $90\%$ of the training data.  
Despite this imbalance, the models achieve competitive validation performance not only in English but also in minority languages such as Arabic, indicating that the learned representations capture transferable features rather than being restricted to high-resource languages.  
Together, these results suggest that multilingual circuits remain robust under imbalance and motivate a deeper investigation into how such robustness is organized within the model. More results are presented in Appendix~\ref{appendix:validation_losses}.

    


\subsubsection{Language Entropy Across Layers}
\begin{figure}
    \centering
    \includegraphics[width=0.85\linewidth]{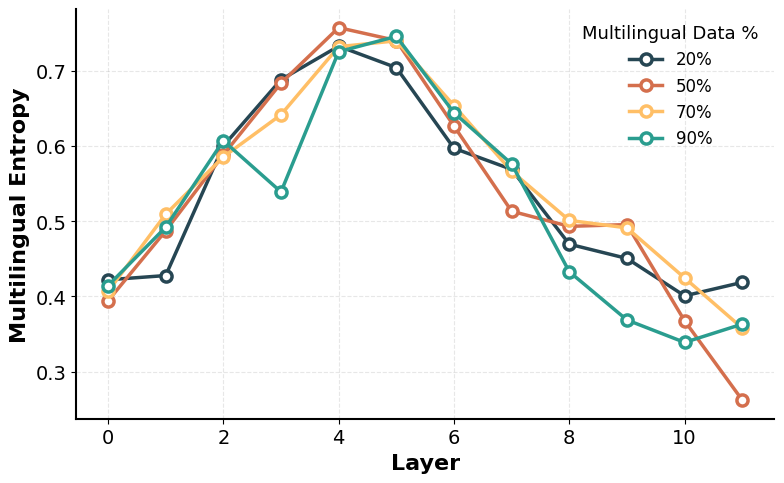}
    \caption{Rate-weighted multilingual score $H(f)$ across layers, showing that middle layers consistently form a multilingual space, while early and late layers are more language-specific. This pattern holds across model sizes and training data mixtures.}
    \label{fig:multilingual_score}
\end{figure}

To investigate how multilingual representations are distributed within the network, we analyze the language entropy of feature activations at each layer. For each feature, we compute the multilingual score 
$H(f)$ as defined in Section~\ref{subsec:multilingual_score}, and multiply it by its corresponding activation rate computed over a large subset of the training data (4M Tokens). Figure~\ref{fig:multilingual_score} shows the resulting average weighted entropy for the top100 score across layers for models trained on different data mixtures.

Here we summarize three major findings on the layerwise organization of multilingual representations:
\\
\paragraph{Finding 1.} Layerwise entropy follows a U-shaped trend. Entropy is relatively low in early layers, rises sharply in the middle layers, and decreases again toward the output layers. This indicates that early layers primarily encode language-specific features, middle layers integrate information into a shared multilingual space, and later layers partially re-specialize for language-specific decoding.
\paragraph{Finding 2.} The trend is robust across data mixtures and model scales. The emergence of a shared multilingual latent space in the middle layers is stable across different training data, and we also observe this well-defined behavior in the Llama-3.2-1B CLTs (see Figure~\ref{fig:llama_multilingual_score}), showing generalization to larger models.
\paragraph{Finding 3.} Model depth influences the behavior. For the 4-layer TinyStories model, the up-and-down pattern is absent and all layers have similar entropy scores (Figure~\ref{fig:llama_multilingual_score}), suggesting a minimum model size for this behavior to emerge.

The behavior is similar for the general entropy score. A discussion comparing the top100 score and the overall training distribution score is provided in Appendix~\ref{appendix:comparaison_multilingual_score}. These findings complement the validation loss results and provide quantitative evidence that the model organizes its internal representations to support cross-lingual generalization.

\paragraph{Shared Multilingual Space Independence from English.} To validate that shared multilingual space 
structures do not require English dominance, we trained GPT-2 on balanced non-English 
data (25\% French, 25\% German, 25\% Arabic, 25\% Chinese) and computed weighted 
multilingual entropy across layers. The resulting curve exhibits the identical 
characteristic rise-and-fall pattern observed in English-dominant models, with peak 
entropy in layers 5-6 (Figure~\ref{fig:pivot}). This demonstrates that the shared multilingual space 
emergence is a fundamental architectural property, not an artifact of English dominance.

\subsubsection{Multilingual Circuits Across Data Mixtures}
\label{subsec:multiling_circuits}
\begin{figure}[h!]
\centering
\begin{subfigure}[b]{0.45\textwidth}
\centering
\includegraphics[width=\textwidth,trim=0 0 0 2, clip]{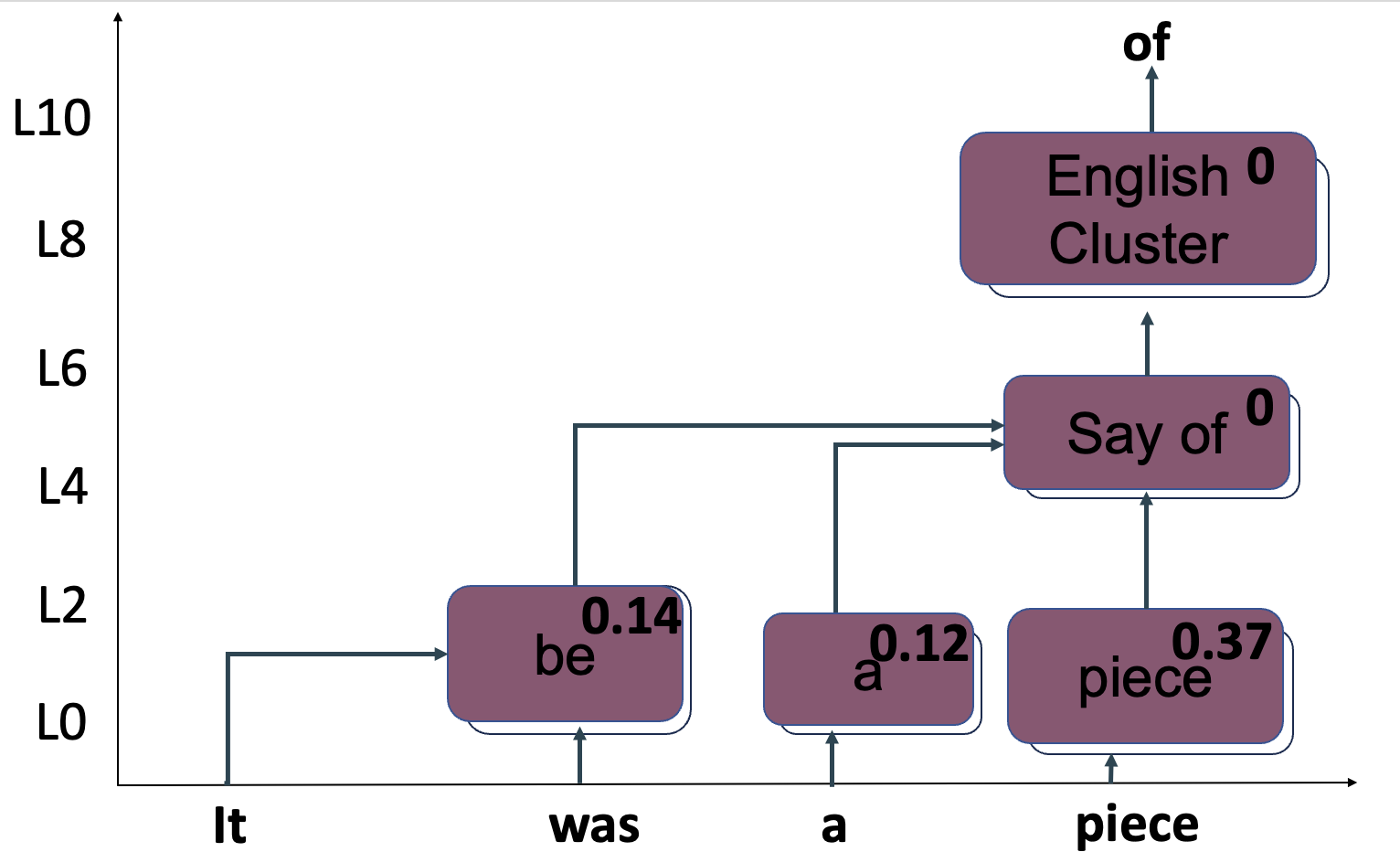}
\end{subfigure}
\hfill
\begin{subfigure}[b]{0.45\textwidth}
\centering
\includegraphics[width=\textwidth]{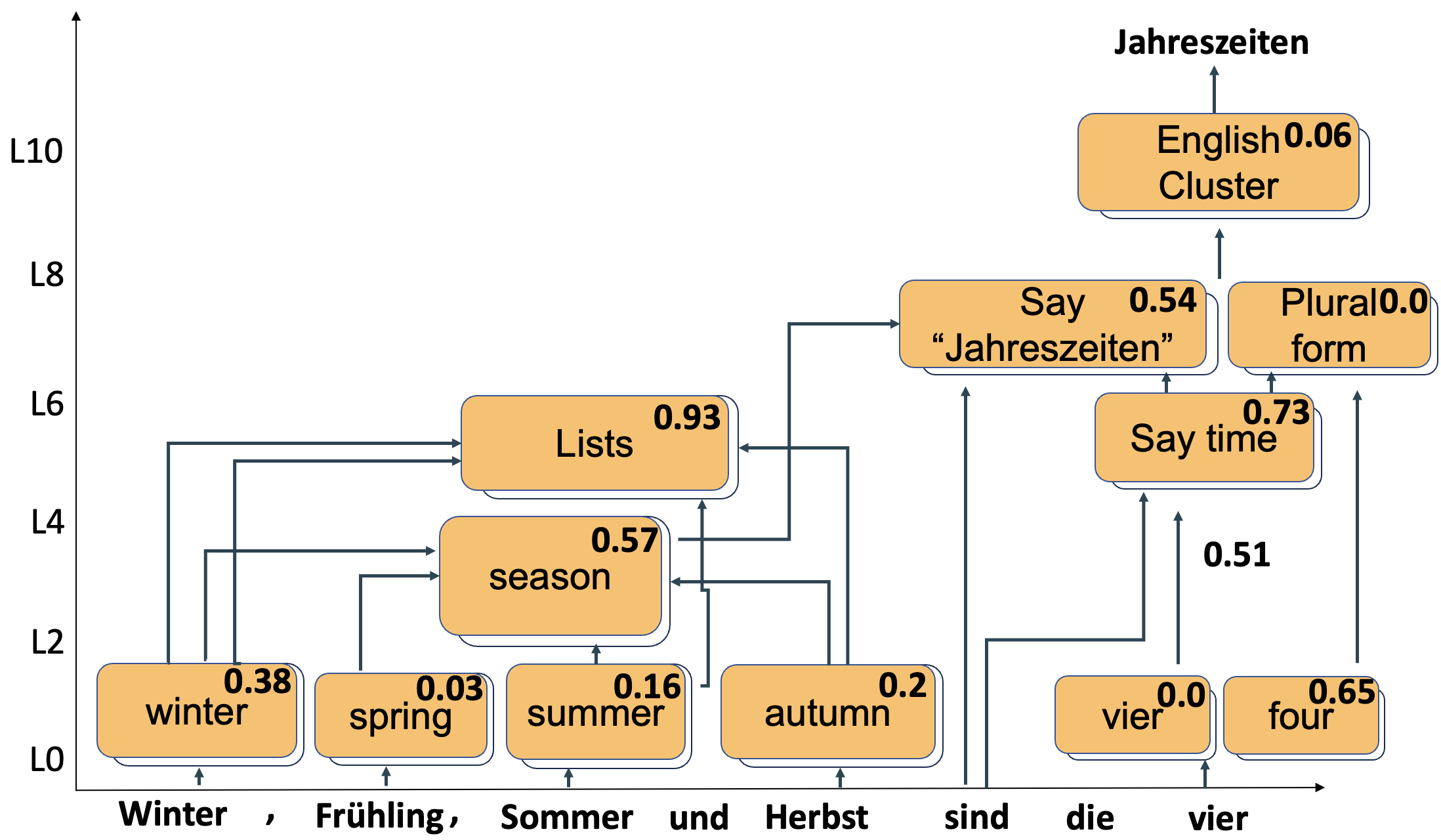}
\end{subfigure}

\caption{Circuits identified across different training
data mixtures. Middle layers consistently form multilingual
clusters, while early and late layers remain more
language-specific. The scores at the top right of each
cluster indicate language entropy (higher values correspond
to more multilingual clusters). The patterns are
stable across mixtures, sentence types, and languages.}
\label{fig:multiling_evidence}
\end{figure}

To complement the multilingual score analysis, we extract attribution graphs to identify circuits active across data mixtures. We focus on two sentence types: (i) preposition sentences, where the model predicts a function word such as in J'ai bu une tasse or It was a piece, and (ii) content word sentences, where it predicts a meaningful token such as I prefer drinking tea to drinking in the 90\% English mixture or Winter, spring, summer, and autumn are the four in the 20\% mixture. We also analyze examples involving calendar terms (Monday, Tuesday, Wednesday, Thursday) and analogy prompts (the opposite of 'men' is…). Figure~\ref{fig:multiling_evidence} shows two representative circuits, one for an English preposition and one for a German content word, with remaining examples in Appendix~\ref{app:circuits} and more details about the clustering and cluster's entropy calculation in Appendix~\ref{app:multi-case-study} .

\begin{figure*}
\centering
\includegraphics[width=\linewidth]{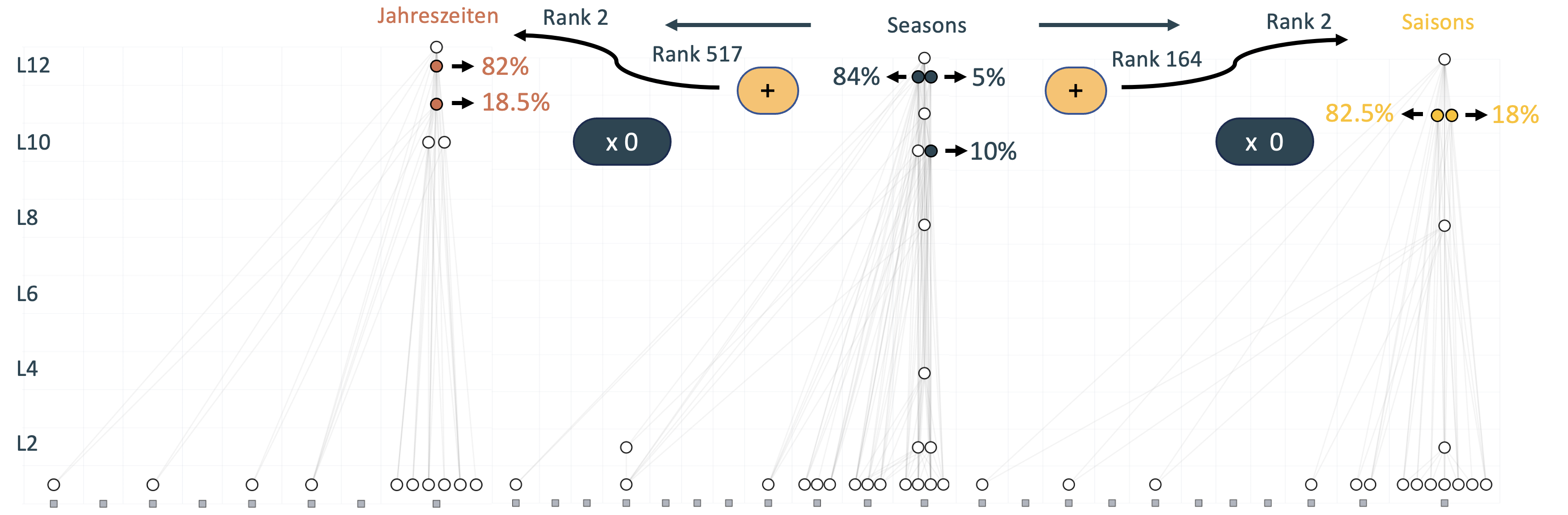}
\caption{Attribution graphs in French, English, and German for the 20\% model pruned at 50\% for the prompt ``Autumn, Winter, Fall, and Spring are the four''. Interventions are displayed and language features highlighted in color with their corresponding percentage of activation on their language.}
\label{fig:language_feature_graph}
\end{figure*}

Here we summarize three main findings on multilingual circuits across layers and data mixtures:

\paragraph{Finding 1.} Layerwise organization exhibits a three-phase structure: early and late layers are mostly language-specific, while middle layers form dense multilingual clusters. Circuits for determiners remain largely language-specific, confirming observations from \citet{schut2025multilingualllmsthinkenglish}.

\paragraph{Finding 2.} Semantic alignment can emerge early or late depending on the data. In some cases in the 90\% English model, the late-layer English cluster is missing, suggesting the model does not always rely on English for semantic grounding. Conversely, layer 0 sometimes already shows multilingual alignment. When tokenization is clean, the first attention and MLP blocks link semantically equivalent tokens across languages, indicating that semantic mapping begins from the very first layer. In the 20\% mixture, Arabic features link words sharing historical meanings, such as connecting ``qalb'' (“heart”) to its older sense of “change” or “transformation,” showing that the model recovers deep etymological relations.

\paragraph{Finding 3.} Performance and representation quality are affected by tokenization. The model’s weaker performance on Arabic partially results from tokenization. Even with a balanced multilingual tokenizer, Arabic words are frequently split into small fragments, forcing the first layers (up to layer 3) to focus on reassembling words rather than learning higher-level meaning (Figures~\ref{fig:arabic_fail_1} and~\ref{fig:arabic_fail_2}).

\subsection{Mechanisms of Language Decoding}
\label{sec:language_decoding}
A central question in multilingual models with pivot languages is how models encode language information and integrate it in final layers to predict the correct next token. Across all graphs, language-specific features appear predominantly in early and late layers. In early layers, these features often correspond to single-token content rather than language alone. As in \cite{lindsey2025biology}, we find early-layer features indicating output language for specific contexts (e.g., after quotation marks in ``\texttt{The opposite of ``men'' is ``}''), but these features are not consistently present in general cases.

\paragraph{Finding 1.}
Across all attribution graphs, we find that late-layer language clusters drive decoding. These clusters correspond to high-frequency features that activate for large proportions of tokens within their respective languages. As shown in Appendix \ref{appendix:language_features}, Figure \ref{fig:scatter}, most high-frequency CLT features are language-specific. Figure \ref{fig:language_feature_graph} shows that these late-layer activations can be linearly read from early-layer features and embedding nodes, indicating that language identity propagates hierarchically through the model.

\paragraph{Finding 2.}
Direct intervention experiments confirm the causal role of these clusters in controlling output language. Zeroing late-layer language features for one language and adding those from a target language (using activations from translated prompts) substantially increases the target-language logit rank. Perfect prediction typically requires additional feature modifications or sweeping intervention values.

\begin{figure}
\centering
\includegraphics[width=\linewidth]{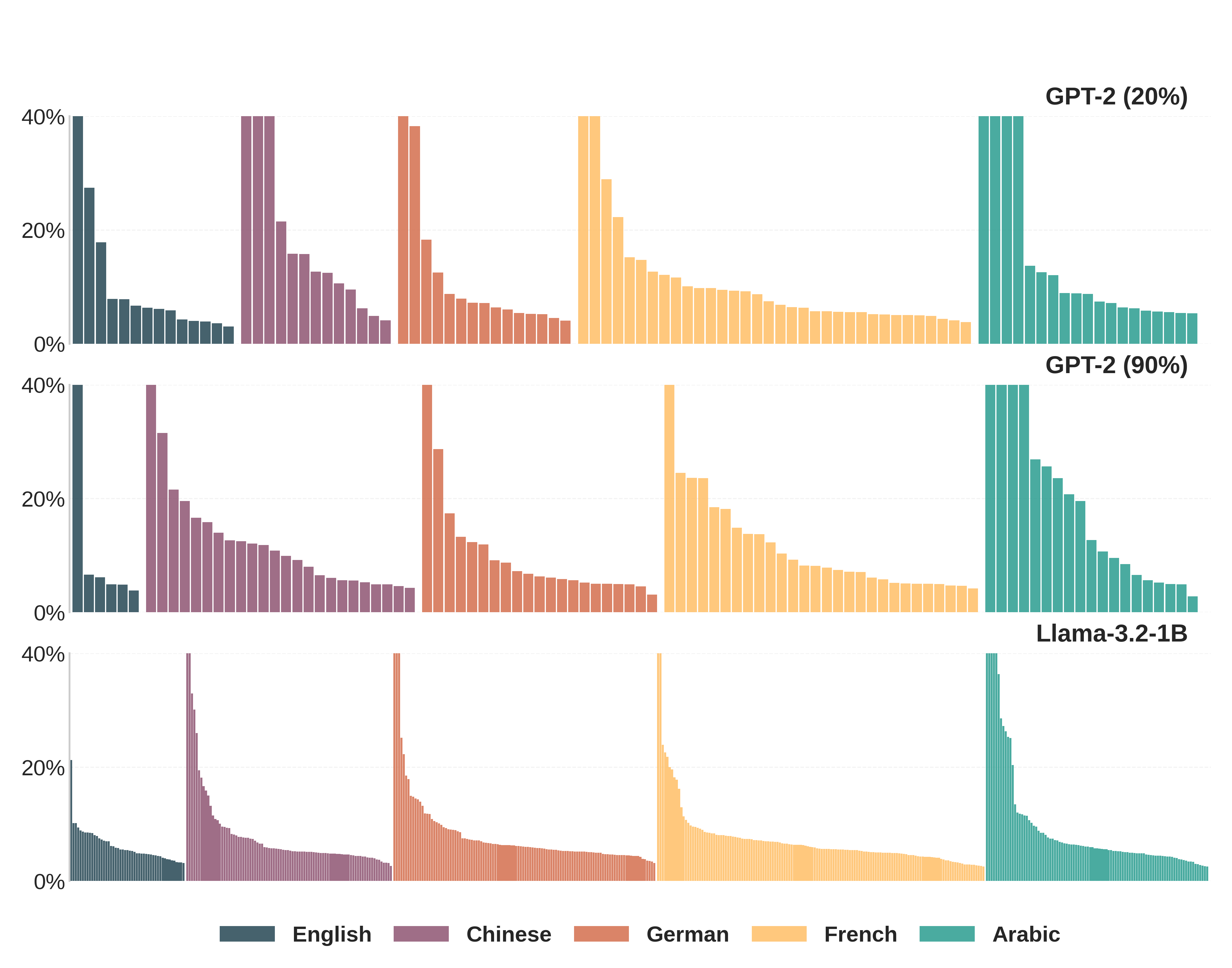}
\caption{Frequency of token activations over their respective languages of the language features for the $20\%$ model.}
\label{fig:frequency_comparaison}
\end{figure}

\paragraph{Finding 3.}
Finally, activation frequency patterns reveal both language ambiguity and dominance effects. As shown in Figure \ref{fig:frequency_comparaison}, a small number of features activate on 50–100\% of tokens within each language, while inactive tokens occur mostly at sequence beginnings where language identity is ambiguous (especially between French, English, and German). Dominant languages exhibit fewer high-frequency features than others across both GPT-2 and LLaMA models, suggesting that English functions as a default output language requiring fewer explicit high-frequency features \cite{lindsey2025biology}.

Language-specific features exhibit significant overlap with English features that increases with English training proportion (77\% to 84\%), while non-English languages maintain stronger mutual feature overlap ($89\%$) that progressively converges toward English at higher proportions. This pattern indicates that the shared multilingual space emergence is partly driven by training data composition, with English acting as a convergence point (Appendix~\ref{appendix:language_features}).



\subsection{Failure Mechanisms in Non-English Languages}
\label{sec:why_models_fail}

As shown in Section~\ref{subsec:multiling_circuits}, multilingual LLMs develop shared representations in their middle layers, indicating that equivalent meanings are processed via common internal mechanisms across languages. Despite this, models consistently perform better on English, and even models trained on uniformly distributed multilingual data show a persistent English bias. This raises a key question: if the model relies on shared circuits, why does this performance gap persist?

To address this, we analyze failure and repair mechanisms in non-English languages using a \textbf{model-diffing approach}. Intuitively, the features a model adjusts to perform better in a language reveal what it struggles with most. Let $\mathcal{M}_0$ denote the base multilingual model, and $\mathcal{M}_{\ell}$ a version finetuned on language $\ell$. We finetune the original CLT (trained on activations from $\mathcal{M}_0$) on the activations of $\mathcal{M}_{\ell}$ to identify the features the model modifies to improve performance. We denote the corresponding feature dictionaries as $\mathcal{D}_0$ and $\mathcal{D}_{\ell}$.

We present the findings from our model-diffing experiments on low-performing languages $\ell \in \{\texttt{Arabic, Chinese, German}\}$. Our goal is to understand which features and layers $\mathcal{M}_\ell$ adjusts relative to the base model $\mathcal{M}_0$, and how these changes propagate to token-level representations.

\subsubsection{Models Repair in Later Layers}
\label{subsec:feature_changes}

For each language $\ell$, we compute the cosine similarity between the feature dictionaries $\mathcal{D}_0$ and $\mathcal{D}_\ell$ across layers.  

\begin{figure*}
    \centering
    \includegraphics[width=\linewidth]{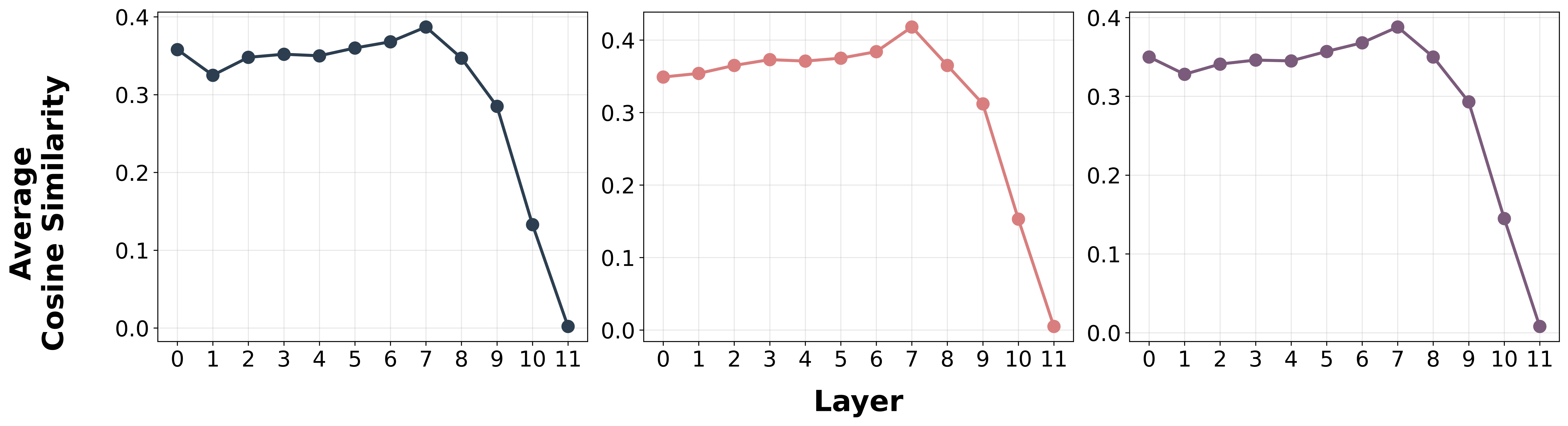}
    \caption{Cosine similarity between $\mathcal{D}_0$ and $\mathcal{D}_\ell$ across layers for \texttt{Arabic} (left). \texttt{German} (middle) and \texttt{Chinese} (right). The largest representation changes occur in the late layers, indicating that the model focuses its most transformative adaptations there.}
    \label{fig:cosine_similarity}
\end{figure*}

Figure~\ref{fig:cosine_similarity} shows that the largest changes occur in the later layers. This finding aligns with our previous observation that language-specific features are predominantly located in these layers. It is intuitive that the model focuses on refining these representations: by strengthening their connection to the multilingual shared space in the middle layers, $\mathcal{M}_\ell$ can more effectively translate and map language-specific inputs to shared high-level representations, ultimately improving performance.

\subsubsection{Repair Through Multilingual Feature Changes}
\label{subsec:feature_buckets}

To understand how finetuning affects multilingual representations, we split features in each layer into two buckets based on cosine similarity: high-change features ($\text{cos-sim}(\mathcal{D}_0, \mathcal{D}_\ell) < 0.5$) and low-change features ($\text{cos-sim}(\mathcal{D}_0, \mathcal{D}_\ell) \ge 0.5$). 

\begin{figure}
    \centering
    \includegraphics[width=\columnwidth]{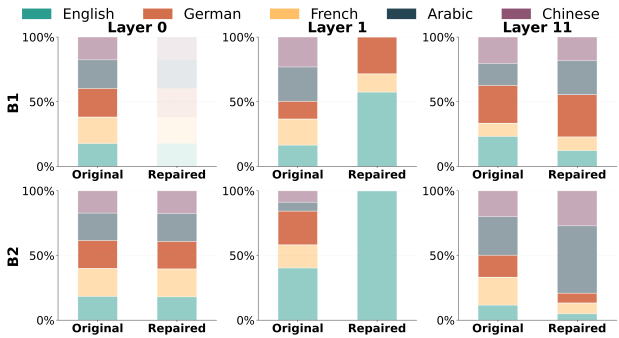}
    \caption{Language distribution comparison across layers for \texttt{German}. Middle layers show high entropy and low language distribution deltas and are therefore omitted from the figure. Results for \texttt{Arabic} and \texttt{Chinese} are presented in Appendix~\ref{app:multiling-buckets}}
    \label{fig:feature_buckets}
\end{figure}

Figure~\ref{fig:feature_buckets} shows the distribution of high-change features across layers. Across all languages, high-change features are concentrated in late layers, consistent with our previous analysis. Interestingly, we observe additional patterns in the early and late layers:

\begin{itemize}
    \item \textbf{Layer 0 (embedding layer):} Features remain largely unchanged, indicating that the model preserves the base token embeddings.
    \item \textbf{Layer 1:} Changes primarily eliminate reductions for Chinese, while English and French dominate the remaining high-change features. This suggests early-layer adjustments help align language-specific signals with shared representations.
    \item \textbf{Layer 11 (late layer):} Most transformative changes occur here. High-change features (\textit{bucket 1}) consistently increase probabilities for the target language across all languages, indicating that the model performs subtle adjustments—minor rotations or rescaling—to align language-specific features with the correct output while maintaining stable, well-behaved representations. 
\end{itemize}

These adjustments naturally push features from other languages into \textit{bucket 2} (low similarity), balancing the representation space. Notably, we also observe cross-language effects: finetuning German increases Arabic features in \textit{bucket 1}, and Chinese exhibits similar interactions with French and Arabic. This hints at overlapping subspaces between languages, where modifications in one language can partially transfer to others. More results and experimental setup are shown in Appendix~\ref{app:multiling-buckets}.

\subsubsection{Causal Impact of High-Change Features}
\label{subsec:causal_interventions}

To assess the functional role of the high-change features identified in $\mathcal{D}_\ell$, we perform a causal analysis using LogitLens. For a given high-change feature $f \in \mathcal{D}_\ell$ and $\mathcal{D}_0$, we compute its top aligned tokens via the unembedding matrix:

\[
\mathbf{t}^*(\mathbf{f}) = \arg\max_i \big[ W_U W_{\text{dec}} f \big]_i,
\]

where $W_{\text{dec}}$ is the decoder projection and $W_U$ is the unembedding matrix. This operation identifies the tokens most strongly influenced by $\mathbf{f}$.

We find that, after finetuning, these features contribute to increased token assembly in early layers (see Figure~\ref{fig:logitlens-features}). This suggests that late-layer adaptations in $\mathcal{D}_\ell$ propagate backward, enabling $\mathcal{M}_\ell$ to construct more abstract, compositional representations. Furthermore, the norms of high-change features in $\mathcal{D}_\ell$ are larger than in $\mathcal{D}_0$, reflecting stronger and more confident activations.

For further details on the features analyzed and the results of  the Logit Lens experiment, see Appendix~\ref{app:causal_features}.

\subsubsection{Circuit-Level Illustration}
\label{subsec:circuit_examples}

To illustrate our findings, we present over 40 representative circuit examples from \texttt{Arabic}, \texttt{Chinese}, and \texttt{German} (full details in Appendix~\ref{appendix:circuits_eng_better}). These examples show how interventions that activate clusters of high-change features can improve the model's retrieval of the correct answer. We further observe that sub-tokenization weakens edges to the most important clusters, and that the tokenizer—despite being trained on equal amounts of data per language—exhibits a bias toward English. This bias provides a mechanistic explanation for the performance gaps observed across languages.

\section{Conclusion}

We present a mechanistic account of multilingual processing in LLMs using Cross-Layer Transcoders, showing that models form shared multilingual representations in their middle layers while relying on small sets of high-frequency features for language-specific decoding. These features linearly read out language identity from earlier layers, and targeted interventions on them can reliably shift the model’s output language. Performance differences across languages seem to arise less from missing multilingual circuits and more from factors such as tokenization and the strength of downstream activations. Overall, our results show how CLTs make it possible to pinpoint where multilingual disparities emerge and provide a practical basis for more focused investigations of cross-language processing in LLMs.

\section*{Limitations}

\textbf{Model Scale and Architecture.} Our primary analysis focuses on relatively small models (68M and 177M parameters) trained from scratch. While we include LLaMA-3.2-1B for validation and observe consistent patterns, our findings may not fully generalize to production-scale models (10B+ parameters) or models trained with different architectural choices. Larger models may develop additional mechanisms or more nuanced multilingual representations that our current analysis does not capture.


\textbf{Tokenization Analysis.} While we identify tokenization as a critical factor in performance disparities, our analysis uses a single BPE-based tokenizer trained on balanced data. We do not systematically compare alternative tokenization strategies (e.g., character-level, morpheme-aware, or language-specific tokenizers) that might better preserve semantic information across scripts. The optimal tokenization approach for multilingual models remains an open question.

\textbf{Reproducibility Constraints.} The computational cost of training both base models and CLTs (requiring thousands of GPU-hours) may limit independent replication. While we release our models and CLTs, the specific training dynamics and their sensitivity to random seeds, hardware variations, and implementation details have not been exhaustively characterized.

\section*{Impact Statement}

CLTs make it possible to trace multilingual processing step by step—from early encoding through shared representations to language-specific decoding—revealing where mechanisms are shared across languages and where they diverge. This enables a concrete, mechanistic view of language disparities: rather than relying solely on task-level performance, we can directly identify circuits that are weaker, missing, or more fragmented for particular languages. Using this lens, we show that even under balanced training data, multilingual models exhibit systematic gaps; for example, Arabic is heavily sub-tokenized, requiring early layers to reconstruct fragmented words. We release our models and CLTs to support further research, while noting that training required thousands of GPU-hours and that intervention capabilities raise fairness and misuse considerations. Although our experiments focus on five languages and relatively small models, the approach provides a structured foundation for analyzing multilingual processing and developing more equitable multilingual AI systems.
\section*{Acknowledgment}
We thank Michael Hanna for insightful discussions on the added value of Cross-Layer Transcoders compared to Single-Layer Transcoders, as well as on the training challenges and implementation details of CLTs. We are grateful to Anson Lei for his assistance with the trained GPT-2 model generation issue and for sharing tips on prompt formatting. We also thank Francesco Ortu for his valuable feedback, which helped improve the clarity and flow of the paper and Antia Garcia for helping improve the design of the main figure.

This material is based in part upon work supported by the German Federal Ministry of Education and Research (BMBF): Tübingen AI Center, FKZ: 01IS18039B; by the Machine Learning Cluster of Excellence, EXC number 2064/1 – Project number 390727645; by Schmidt Sciences SAFE-AI Grant; by the Frontier Model Forum and AI Safety Fund; by Open Philanthropy;
by the Cooperative AI Foundation; and by the Survival and Flourishing Fund.
Resources used in preparing this research were provided, in part, by the Province of Ontario, the Government of Canada through CIFAR, and companies sponsoring the Vector Institute. F.D. acknowledges support through a fellowship from the Hector Fellow Academy. We gratefully acknowledge EleutherAI for providing compute resources that enabled the Cross-Layer Transcoder finetuning in our model-diffing experiments.

\bibliography{references}
\bibliographystyle{icml2026}

\cleardoublepage
\appendix
\onecolumn

\section*{\LARGE Supplementary Materials}
\addcontentsline{toc}{section}{Supplementary Material} 
\bigskip

{\LARGE \textbf{Table of Contents}}
\noindent\rule{\linewidth}{0.4pt}

\begin{flushleft}
\textbf{PART I: Additional Methodological Details} \dotfill \pageref{appendix:training} \\[6pt]

\begin{tabularx}{\linewidth}{@{}lXr@{}}
A & Multilingual Model Training & \pageref{appendix:training} \\
& A.1 \quad Tokenization Strategy & \pageref{app:token_strategy} \\
& A.2 \quad Model Architectures and Training & \pageref{app:model_arch_training} \\[4pt]

B & CLT Training & \pageref{appendix:clt_training} \\
& B.1 \quad Training Configuration & \pageref{app:clt_training_config} \\
& B.2 \quad Performance & \pageref{app:clt_peformance} \\[4pt]

C & Additional Validation Loss Curves Under Data Imbalance & \pageref{appendix:validation_losses} \\
& C.1 \quad GPT2 Models & \pageref{app:gpt2-loss} \\
& C.2 \quad TinyStories Models & \pageref{app:tinystories-loss} \\[4pt]

D & Language Feature Analysis & \pageref{appendix:language_features} \\
& D.1 \quad Methodology and Definitions & \pageref{app:method-def} \\
& D.2 \quad Feature Overlap Results & \pageref{app:lang-results} \\[4pt]
\end{tabularx}

\noindent\rule{\linewidth}{0.4pt}

\textbf{PART II: Complementary Analyses} \dotfill \pageref{appendix:comparaison_multilingual_score} \\[6pt]
\begin{tabularx}{\linewidth}{@{}lXr@{}}
E & Multilingual Score & \pageref{appendix:comparaison_multilingual_score} \\
F & Shared Multilingual Space Independence from English Dominance & \pageref{app:noenglish} \\
\end{tabularx}

\noindent\rule{\linewidth}{0.4pt}

\textbf{PART III: Case Studies} \dotfill \pageref{app:multi-case-study} \\[6pt]
\begin{tabularx}{\linewidth}{@{}lXr@{}}
G & Case Studies: Multilingual Circuits & \pageref{app:multi-case-study} \\
& G.1 \quad Multilingual Circuits: U-shaped Entropy & \pageref{app:circuits} \\
& G.2 \quad Multilingual Circuits: Failure Modes & \pageref{appendix:circuits_eng_better} \\
& G.3 \quad Intervention Validation & \pageref{app:intervention-validation} \\[4pt]

H & Extended Task Analysis: Translation and Cultural Prompts  & \pageref{app:complextasks} \\
& H.1 \quad Translation Prompts & \pageref{app:trans-prompts} \\
& H.2 \quad Cultural Context Prompts & \pageref{app:culture-context-prompts} \\[4pt]

I & Why English Performs Better: A Case Study & \pageref{app:why_eng_better} \\
& I.1 \quad Cluster Activation Analysis & \pageref{app:eng-better-cluster-activ} \\
& I.2 \quad Sub-tokenization Effects & \pageref{sec:subtokenization-effects} \\[4pt]
\end{tabularx}

\noindent\rule{\linewidth}{0.4pt}

\textbf{PART IV: Model Diffing and Tokenizer Analysis} \dotfill \pageref{app:model-diffing} \\[6pt]
\begin{tabularx}{\linewidth}{@{}lXr@{}}
M & Model Diffing: Additional Material & \pageref{app:model-diffing} \\
& M.1 \quad Multilinguality of Repair Features & \pageref{app:multiling-buckets} \\
& M.2 \quad Token Assembly in Early Layers & \pageref{app:causal_features} \\[4pt]

N & Tokenizer Analysis: BPE Constraints on Multilingual Processing & \pageref{app:tokenizer} \\
& N.1 \quad Vocabulary Allocation: Cannibalization Testing & \pageref{app:vocab-allocation} \\
& N.2 \quad Morphological Coherence & \pageref{app:morph-coherence} \\
& N.3 \quad Mechanistic Implications & \pageref{app:tokenizer-mech-imp} \\[4pt]
\end{tabularx}

\end{flushleft}

\newpage
\section{Multilingual Model Training}
\label{appendix:training}

\subsection{Tokenization Strategy}
\label{app:token_strategy}
All models employ a unified tokenization approach designed to eliminate tokenization-induced biases. We train a debiased BPE tokenizer on uniformly distributed five-language data, with each language contributing exactly the same number of tokens. The resulting vocabulary contains 119,547 tokens, ensuring balanced representation across languages and writing systems.

\paragraph{Tokenizer Training Protocol}
\begin{itemize}
    \item \textbf{Training Data}: Equal token counts per language (20\% each)
    \item \textbf{Vocabulary Size}: 119,547 tokens
    \item \textbf{Algorithm}: Byte-Pair Encoding (BPE)
    \item \textbf{Special Tokens}: BOS (beginning of sentence), EOS (end of sentence), PAD (padding), UNK (unknown)
\end{itemize}

\begin{table*}[H]
\centering
\begin{tabular}{lccccc}
\toprule
\textbf{Architecture} & \textbf{Layers} & \textbf{Hidden Dim} & \textbf{Attention Heads} & \textbf{Parameters} & \textbf{Context Length} \\
\midrule
GPT-2        & 12 & 768 & 12 & 177.6M & 1024 \\
TinyStories  & 4  & 768 & 16 & 68.5M  & 512  \\
\bottomrule
\end{tabular}
\caption{Model architecture specifications.}
\label{tab:model_architectures}
\end{table*}

\subsection{Model Architectures and Training}
\label{app:model_arch_training}
\subsubsection{Model Architecture Specifications}
We train transformer models following two architectures to enable comparisons across model families and scales:

\begin{table}[H]
\centering
\begin{tabular}{lcc}
\toprule
\textbf{Component} & \textbf{GPT-2} & \textbf{TinyStories} \\
\midrule
Vocabulary Size        & 119,547 & 119,547 \\
Embedding Dimension    & 768     & 768 \\
FFN Inner Dimension    & 3072    & 3072 \\
Attention Head Dim.    & 64      & 48 \\
Dropout Rate           & 0.1     & 0.1 \\
Layer Norm             & Pre-LN  & Pre-LN \\
Activation Function    & GELU    & GELU \\
Position Encoding      & Learned & Learned \\
Context Length         & 1024    & 512 \\
\bottomrule
\end{tabular}
\caption{Detailed architecture hyperparameters.}
\label{tab:detailed_hyperparams}
\end{table}

\subsubsection{Training Scale and Compute Optimization}
Following Chinchilla scaling laws \cite{hoffmann2022training}, we compute optimal training data sizes in Table~\ref{tab:training_scale}.

\begin{equation}
N_{\text{tokens}} = 20 \times N_{\text{parameters}}
\end{equation}

\begin{table}[H]
\centering
\setlength{\tabcolsep}{4pt} 
\begin{tabular}{lccc}
\toprule
Model & Tokens & Steps & FLOPs \\
\midrule
GPT-2       & 3.55B & 1.73M & $3.8\times 10^{12}$ \\
TinyStories & 1.37B & 670K  & $5.6\times 10^{11}$ \\
\bottomrule
\end{tabular}
\caption{Compute-optimal training specs.}
\label{tab:training_scale}
\end{table}

\subsubsection{Training Implementation}
We use the \texttt{nanoGPT} codebase \citep{Karpathy2022} for reproducible transformer training and use the hyperparameters presented in Table~\ref{tab:training_hyperparams}.

\paragraph{Optimization Configuration}
\begin{table}[H]
\centering
\begin{tabularx}{\linewidth}{lX lX}
\toprule
\textbf{Parameter} & \textbf{Value} & \textbf{Parameter} & \textbf{Value} \\
\midrule
Optimizer & AdamW & Learning Rate & 6e-4 \\
$\beta_1$ & 0.9 & $\beta_2$ & 0.95 \\
Weight Decay & 0.1 & Gradient Clipping & 1.0 \\
Warmup Steps & 2000 & LR Schedule & Cosine linear warmup \\
Batch Size & 64 & Gradient Accumulation & 8 steps \\
Effective Batch Size & 512 &  &  \\
\bottomrule
\end{tabularx}
\caption{Training hyperparameters (horizontal layout for one-column appendix).}
\label{tab:training_hyperparams}
\end{table}

\paragraph{Training Infrastructure} We train each model on 4×H100 80GB GPUs. Training is distributed using data parallelism with gradient synchronization, and we use mixed precision (FP16) with automatic loss scaling.

\subsubsection{Training Dynamics}
\begin{figure}[h!]
\centering
\includegraphics[width=0.7\linewidth]{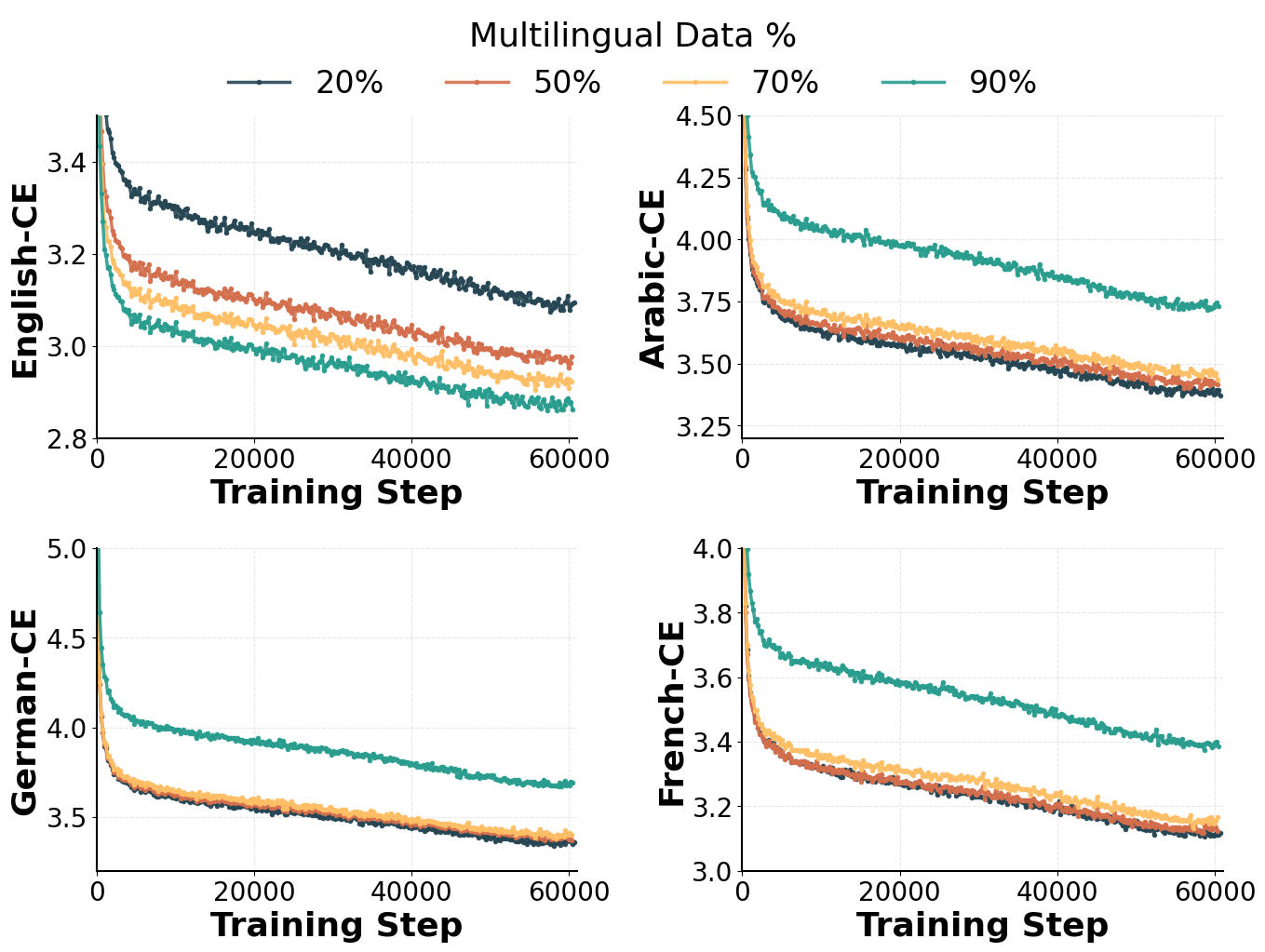}
\caption{Validation cross-entropy loss curves for models trained on multilingual mixtures with varying proportions of English. Despite extreme imbalance (up to 90\% English), the models maintain strong next-token prediction performance across languages.}
\label{fig:combined}
\end{figure}

Figures~\ref{fig:combined}, \ref{fig:appendix_val_chinese} and \ref{fig:appendix_val_tinystories} shows training and validation loss across English data proportions. Higher English proportions reduce English validation loss but can slightly degrade performance on other languages.

\subsubsection{Training Procedure and Models Trained}
Each architecture is trained on all data mixtures: $20\%, 50\%, 70\%$ and $90\%$. Furthermore, we train another model on non-English data ($25\%$ for each of Arabic, Chinese, German and French).

For comparative purposes, we also include a pre-trained LLaMA-3.2-1B \cite{meta-llama32-1b} model alongside our custom-trained models.

\section{CLTs training}
\label{appendix:clt_training}
\subsection{Training Configuration}
\label{app:clt_training_config}
We report in Table~\ref{tab:config_CLT}, the hyper-parameters of the CLT training. We mostly follow Anthropic guidelines \cite{ameisen2025circuit}. This setup is similar across Tinystories, GPT-2, and Llama CLTs.

\begin{table}[H]
\centering
\begin{tabularx}{\linewidth}{lX lX}
\toprule
\textbf{Parameter} & \textbf{Value} & \textbf{Parameter} & \textbf{Value} \\
\midrule
\multicolumn{4}{l}{\textit{Model Architecture}} \\
Input dimension ($d_{\text{in}}$) & 768 & Latent dimension ($d_{\text{latent}}$) & 24,576 \\
Expansion factor & 32 & Context size & 16 \\
\midrule
\multicolumn{4}{l}{\textit{Training Hyperparameters}} \\
Learning rate & $2 \times 10^{-4}$ & Adam $\beta_1$ / $\beta_2$ & 0.9 / 0.999 \\
Batch size (tokens) & 1,024 & LR warm-up steps & 1,000 \\
LR decay steps & 3,749 &  &  \\
\midrule
\multicolumn{4}{l}{\textit{Loss Coefficients}} \\
L0 coefficient & 2.0 & Optimal L0 & 10 \\
Dead penalty coefficient & $1 \times 10^{-5}$ &  &  \\
\midrule
\multicolumn{4}{l}{\textit{JumpReLU Configuration}} \\
Bandwidth & 1.0 & Initial threshold & 0.03 \\
\midrule
\multicolumn{4}{l}{\textit{Other}} \\
Seed & 42 & Precision & float32 (mixed) \\
\bottomrule
\end{tabularx}
\caption{Training Configuration (horizontal layout for one-column appendix).}
\label{tab:config_CLT}
\end{table}

However, for the Llama CLT, we use less features, as we are limited by GPU size with an expansion factor of around 4. We use torch.float32 for the training as well. 

For all CLTs, we stop training when the L0 sparsity coefficient goes below 10 and Explained Variance is beyond $70\%$. At this point, we save the full model activations and then load them in random chunks from the dataset for training. The CLTs contain roughly $78 \times 768 \times 32 \times 768$ parameters, though for the LLaMA CLT we use fewer features (expansion factor = 4) due to GPU memory constraints. Replacement scores are consistently in the range 0.75–0.8, which aligns with expected reconstruction performance. For circuit extraction, attribution graphs are pruned to retain 80\% of cumulative node scores and 85\% of cumulative edge scores using the circuit-tracing library. Training is conducted over approximately 200M tokens and follows standard CLT guidelines \cite{ameisen2025circuit}.

\subsection{Performance}
\label{app:clt_peformance}
\begin{figure*}[h!]
    \centering
    \includegraphics[width=1.\textwidth,trim=0 0 0 25, clip]{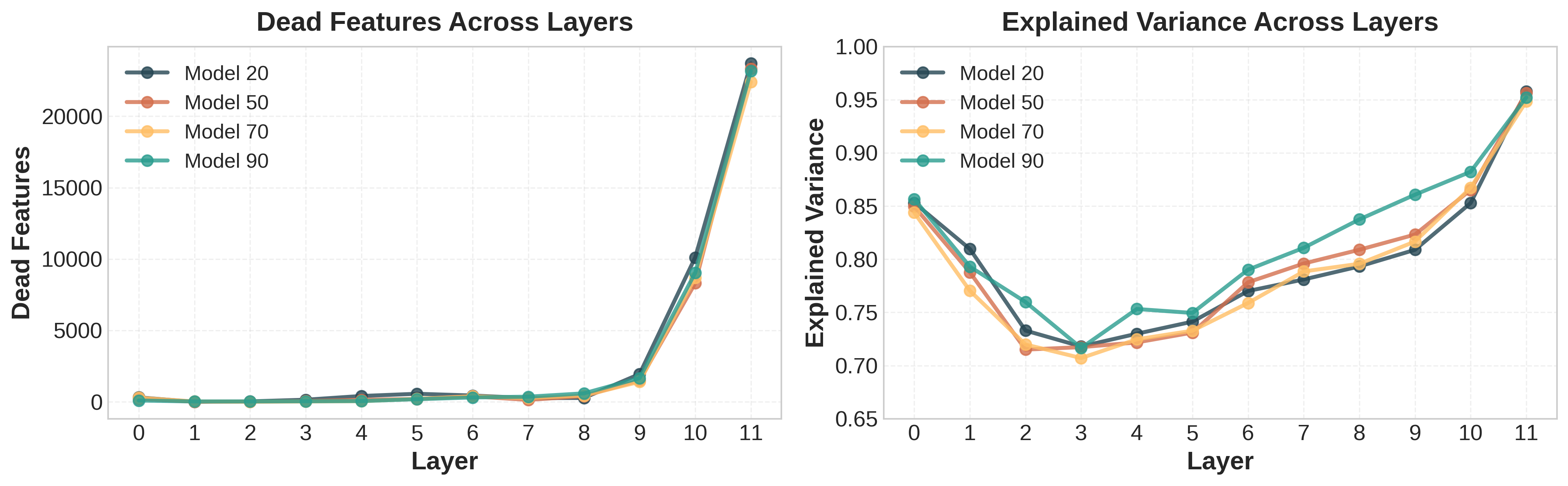}
    \caption{Dead feature count and explained variance across layers for the GPT-2 CLTs.}
    \label{fig:model_metrics_comparison}
\end{figure*}

\begin{figure}[H]
    \centering
    \includegraphics[width=0.7\linewidth,trim=0 0 0 25, clip]{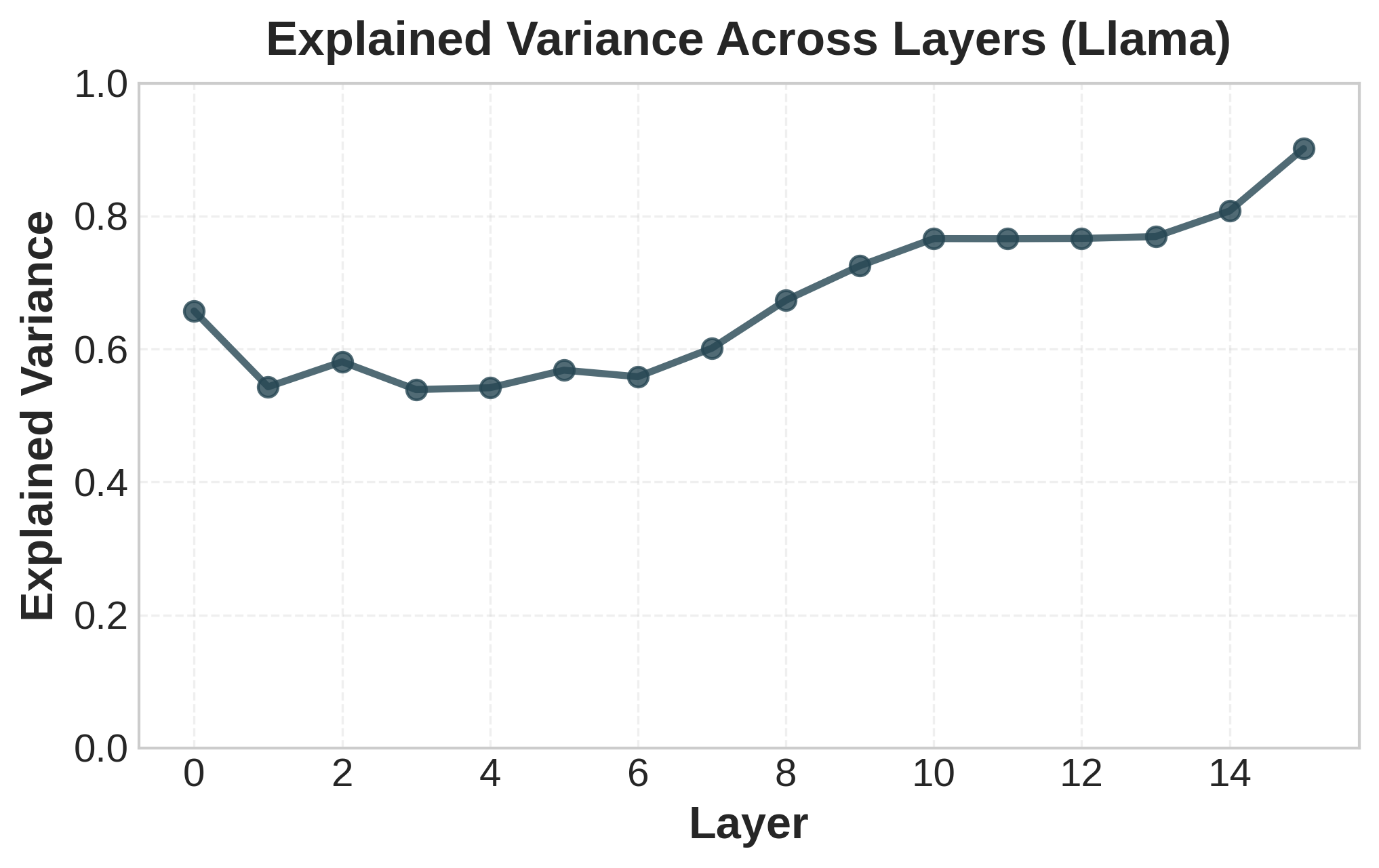}
    \caption{Explained variance over layers for the Llama CLT. }
    \label{fig:llama_mse_comparaison}
\end{figure}

\begin{figure*}[h!]
    \centering
    \includegraphics[width=1.\textwidth,trim=0 0 0 25, clip]{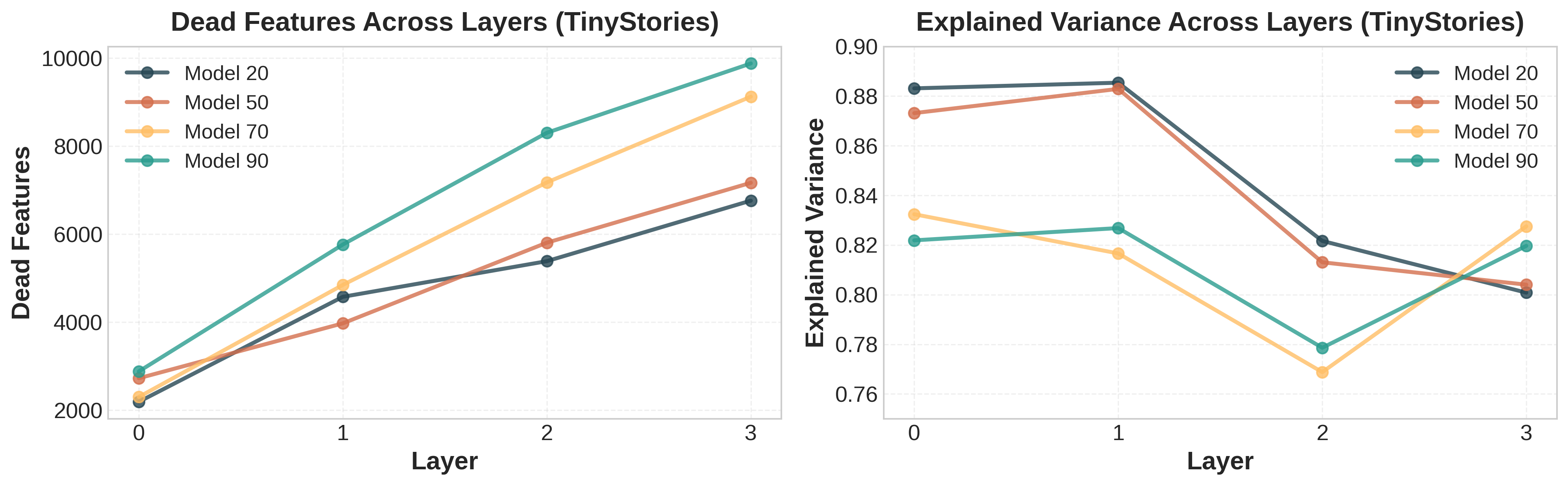}
    \caption{Dead feature count and explained variance across layers for the Tinystories CLTs.}
    \label{fig:model_metrics_comparison_tiny_stories}
\end{figure*}

We report the amount of dead features and the performance of the different models (TinyStories, GPT2, Llama-3.2-1B) CLTs in Figures ~\ref{fig:model_metrics_comparison}, ~\ref{fig:llama_mse_comparaison} and ~\ref{fig:model_metrics_comparison_tiny_stories}. 

\section{Additional Validation Loss Curves Under Data Imbalance}
\label{appendix:validation_losses}

\subsection{GPT2-Models}
\label{app:gpt2-loss}
In Section~\ref{sec:generalization_imbalance}, we reported that the validation loss curves for the four non-English languages (Chinese, Arabic, German, French) show competitive performance.  Here we show the curves.

\begin{figure}[H]
    \centering
    \includegraphics[width=0.5\linewidth,trim=0 0 0 18, clip]{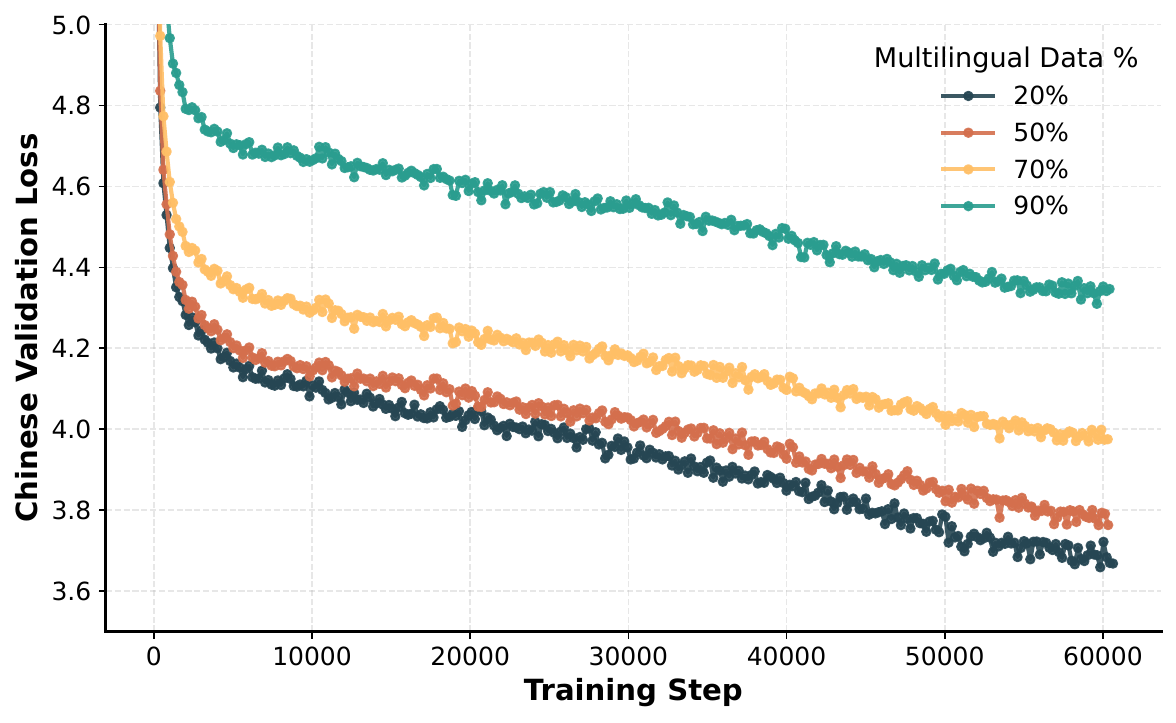}
    \caption{Validation cross-entropy loss curves for Chinese under varying proportions of English. The same robustness pattern holds as for other languages.}
    \label{fig:appendix_val_chinese}
\end{figure}

We also provide additional results for Chinese .
Figure~\ref{fig:appendix_val_chinese} shows the validation loss curves for Chinese.  
The results mirror the patterns observed for other languages: despite extreme imbalance in the training data, the model maintains competitive prediction performance in Chinese, indicating that the robustness of multilingual circuits extends to typologically distinct languages.

\subsection{TinyStories Models}
\label{app:tinystories-loss}
\begin{figure*}[h!]
\centering
\includegraphics[width=\linewidth]{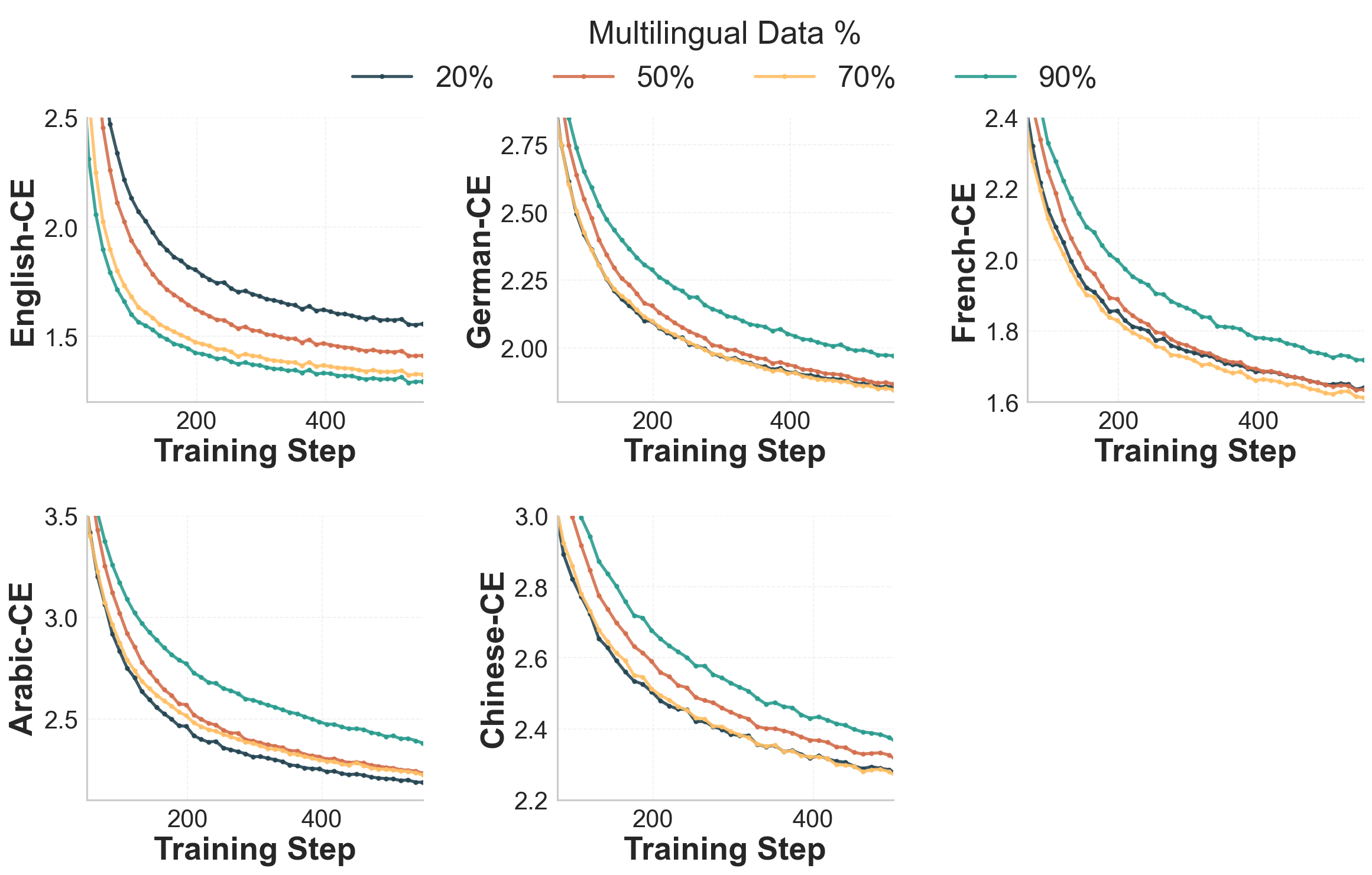}
\caption{Validation cross-entropy loss curves for the \textbf{TinyStories-trained models} across five languages (English, Arabic, German, French, Chinese). Despite small-scale training data, models retain strong multilingual generalization under imbalance.}
\label{fig:appendix_val_tinystories}
\end{figure*}

To assess whether the observed generalization dynamics also hold at small scale, we trained models on the \textbf{TinyStories} dataset.  
Figure~\ref{fig:appendix_val_tinystories} reports validation loss curves for five languages (English, Arabic, German, French, Chinese).  
Despite the reduced data scale and synthetic nature of TinyStories, the same trend emerges: performance remains robust across all languages, confirming that the organization of multilingual circuits is a stable property that persists across both large-scale and small-scale training regimes.

\section{Language Feature Analysis}
\label{appendix:language_features}


We analyze whether high-frequency features exhibit language specificity and quantify the extent to which language-specific features are shared across languages. This analysis uses precomputed activation statistics from CLT training with varying English proportions in the training data: 20\%, 50\%, 70\%, and 90\% English.

\subsection{Methodology and Definitions}
\label{app:method-def}
For each feature $f$ at layer $\ell$, we compute two statistics from the training data. First, the frequency $\text{freq}(f,\ell)$ represents the proportion of tokens for which the feature activates. Second, the language distribution $P(\text{lang} \mid f,\ell)$ captures the probability distribution over languages conditioned on feature activation. Specifically,
\[
P(\text{lang} \mid f,\ell)
=
\frac{N_{\text{seq}}(f,\ell,\text{lang})}
     {N_{\text{seq}}(f,\ell,\text{total})},
\]
where $N_{\text{seq}}(f,\ell,\text{lang})$ denotes the number of sequences in language $\text{lang}$ in which feature $f$ activates at least once.

To label a feature as language-specific, the feature must firstly be high-frequency, with $\text{freq}(f,\ell) \geq 0.03$. Second, the feature must be statistically associated with a given language, requiring $P(\text{lang} \mid f,\ell) \geq 0.5$, i.e., at least 50\% of the sequences on which it activates are from that language. For example, a feature $f$ at layer $\ell$ is labeled French-specific if both $\text{freq}(f,\ell) \geq 0.03$ and $P(\text{French} \mid f,\ell) \geq 0.5$.

For each language pair, we compute feature overlap via set intersection. Let
\[
F_{\text{lang\_a}}
=
\{(\ell,f) : \text{freq}(f,\ell) \geq 0.03 \land P(\text{lang\_a} \mid f,\ell) \geq 0.5\}
\]
denote the set of Language-A-specific features across all layers, with an analogous definition for $F_{\text{lang\_b}}$. We define overlap as
\[
\frac{|F_{\text{lang\_a}} \cap F_{\text{lang\_b}}|}
     {|F_{\text{lang\_a}}|}
\times 100\%,
\]

\subsection{Feature Overlap Results}
\label{app:lang-results}

\begin{figure}[H]
    \centering
    \includegraphics[width=0.7\linewidth]{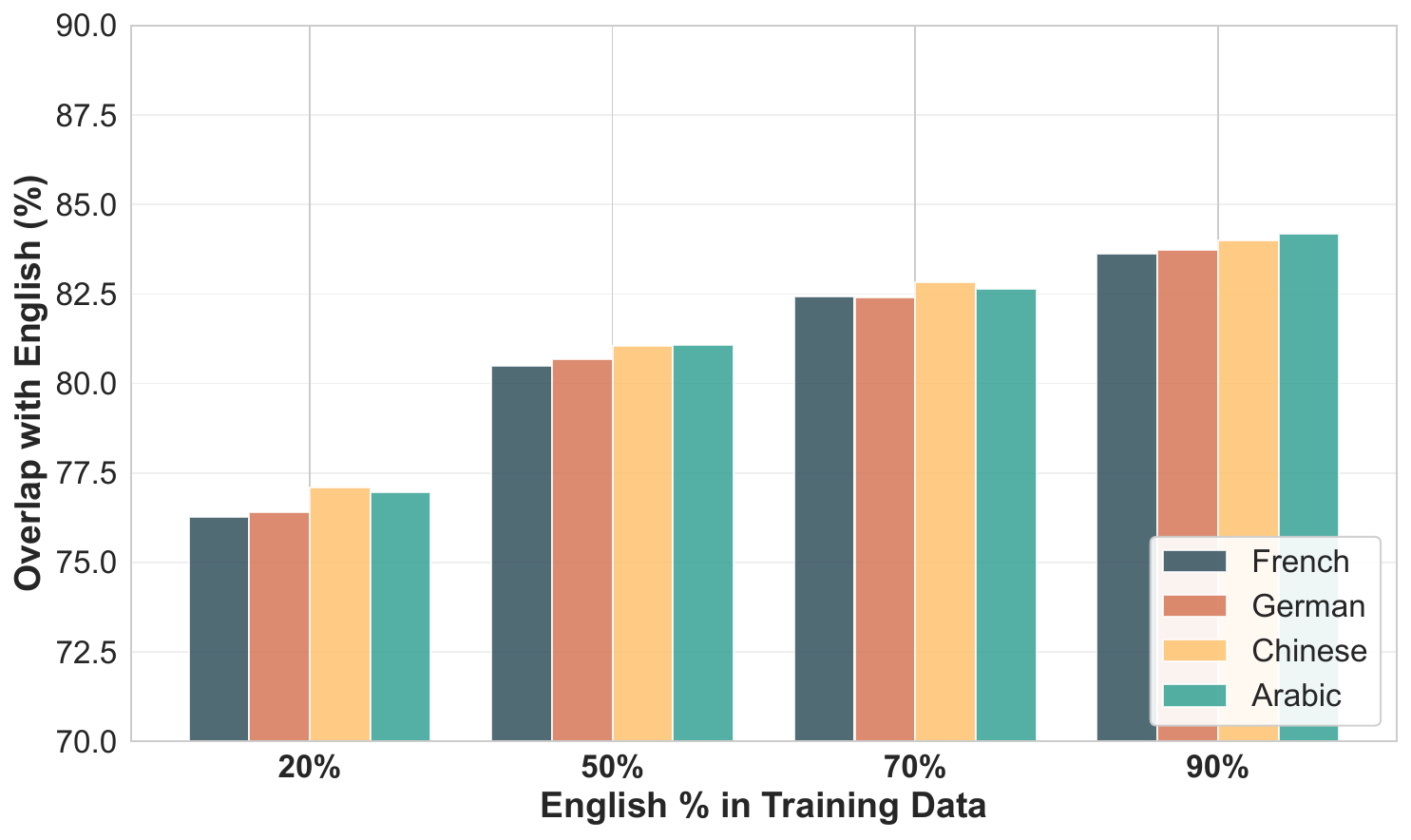}
    \caption{Overlap between non-English language–specific features and English-specific features across training checkpoints.}
    \label{fig:english_overlap}
\end{figure}

Figure~\ref{fig:english_overlap} shows the percentage of each non-English language’s language-specific features that overlap with English-specific features across the four training checkpoints. We observe a consistent increase in overlap as the proportion of English training data increases, rising from approximately 77\% at 20\% English to 84\% at 90\% English across all non-English languages. This trend indicates that increasing English exposure during training drives convergence toward an English-dominant feature space. 

\begin{figure}[H]
    \centering
    \includegraphics[width=0.7\linewidth]{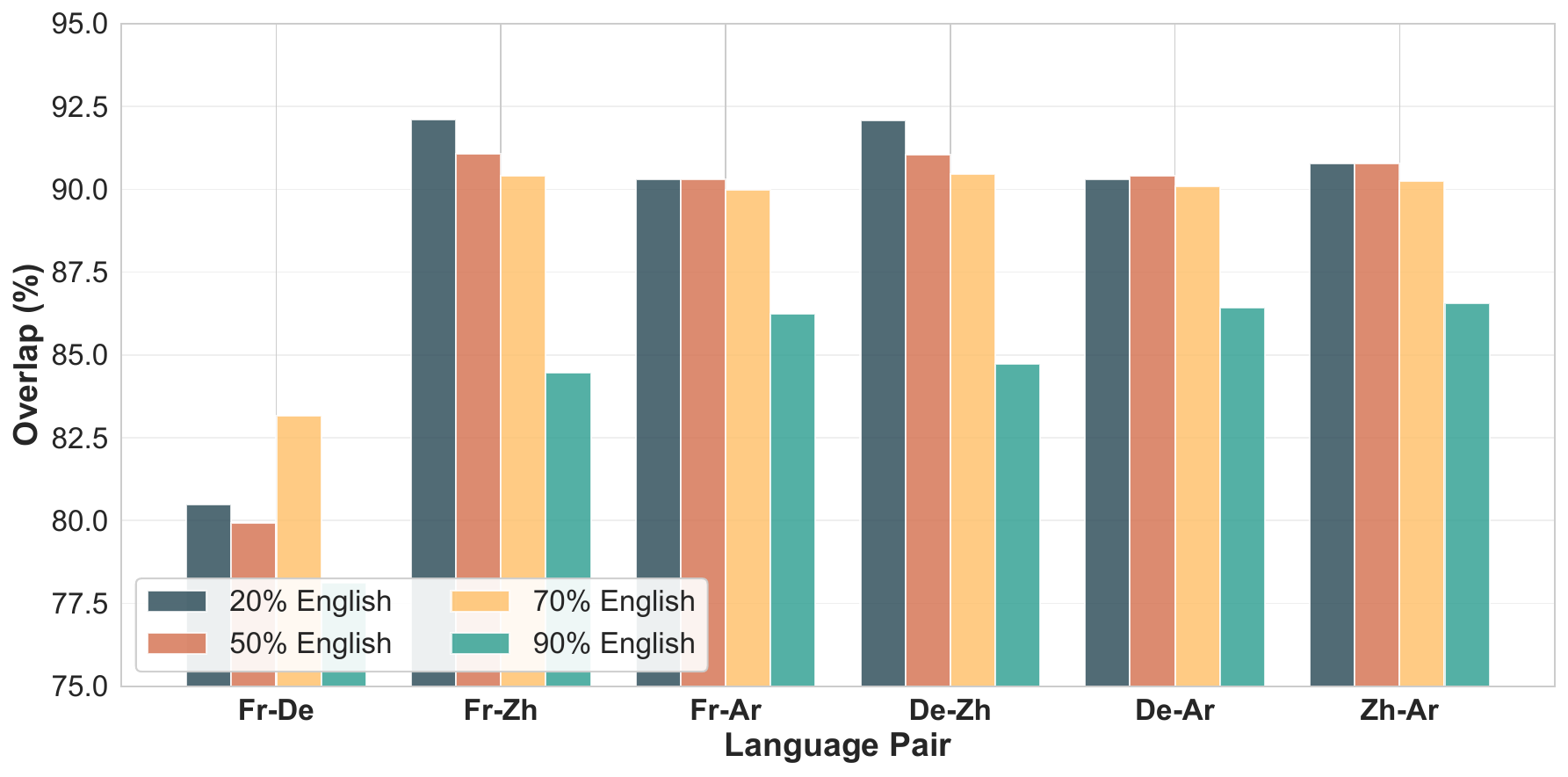}
    \caption{Pairwise overlap of language-specific features among non-English languages.}
    \label{fig:interdependencies}
\end{figure}

Figure~\ref{fig:interdependencies} presents pairwise feature overlap among all non-English language pairs. We find that non-English languages overlap more strongly with each other (averaging 89\% at 20\% English) than they do with English (averaging 77\% at 20\% English), yielding a gap of approximately 12 percentage points. However, this gap progressively closes as English training proportion increases, reaching near parity at 90\% English, indicating forced convergence. 

\begin{figure}[H]
    \centering
    \includegraphics[width=0.5\linewidth]{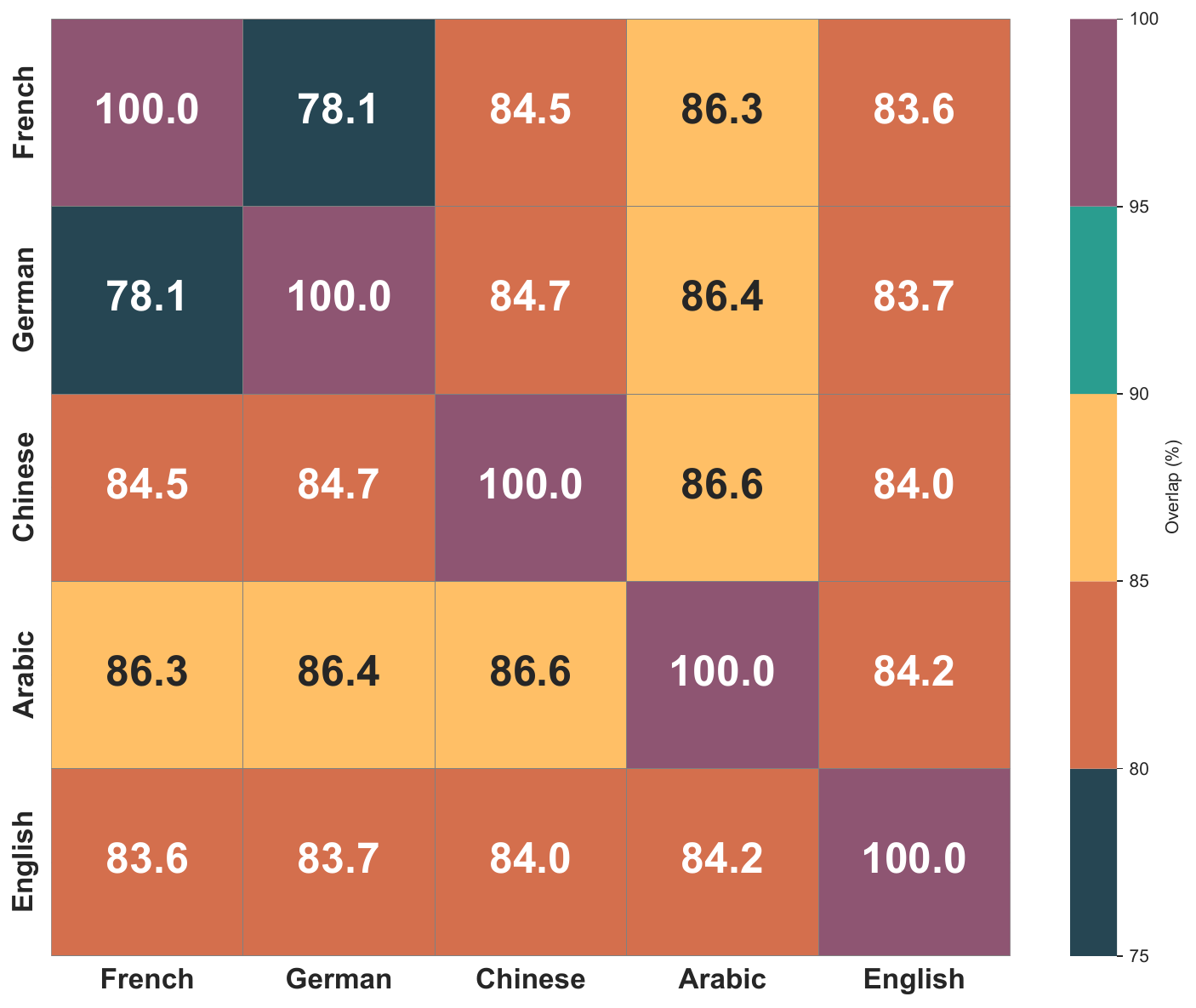}
    \caption{Complete pairwise feature overlap matrix at 90\% English training.}
    \label{fig:heatmap_90}
\end{figure}

Figure~\ref{fig:heatmap_90} shows the full pairwise overlap matrix for the 90\% English checkpoint as a heatmap. The visualization reveals notable asymmetries in pairwise relationships. For example, Chinese and Arabic maintain 86.6\% overlap, substantially higher than the 78.1\% overlap observed between French and German. All pairwise overlaps exceed 78\%, confirming the existence of a broadly shared feature space. English exhibits relatively symmetric overlap with all other languages (83--84\%), consistent with its role as a convergence point under high training proportions.

We add here two plots related to the language features. 
\begin{figure*}[h!]
    \centering
    \includegraphics[width=\linewidth,trim=0 0 0 25, clip]{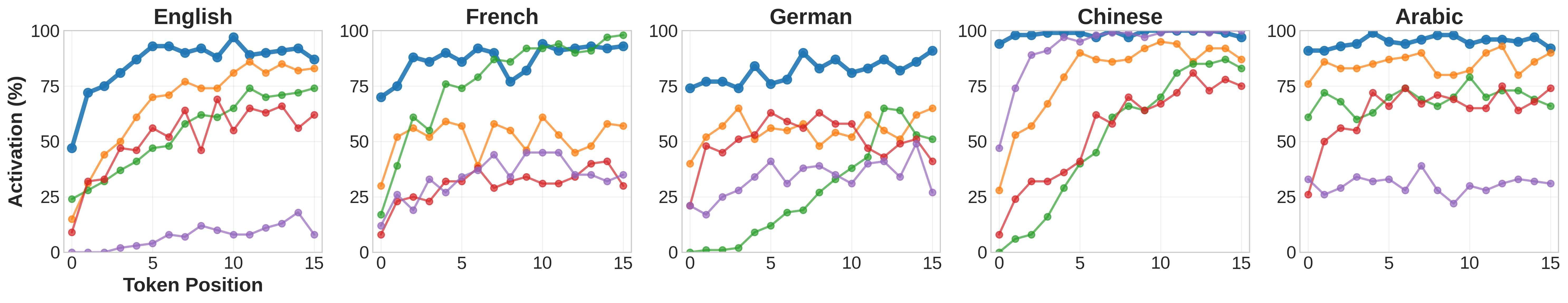}
     \caption{Activation rate over the top100 sequences of the language feature for the 20\% model. We see that some language features are more active towards the end of the sequence. Highlighted is the language feature with the highest frequency.}
     \label{fig:positional_activations_by_language}
\end{figure*}

\begin{figure}[H]
    \centering
    \includegraphics[width=0.6\linewidth,trim=0 0 0 25, clip]{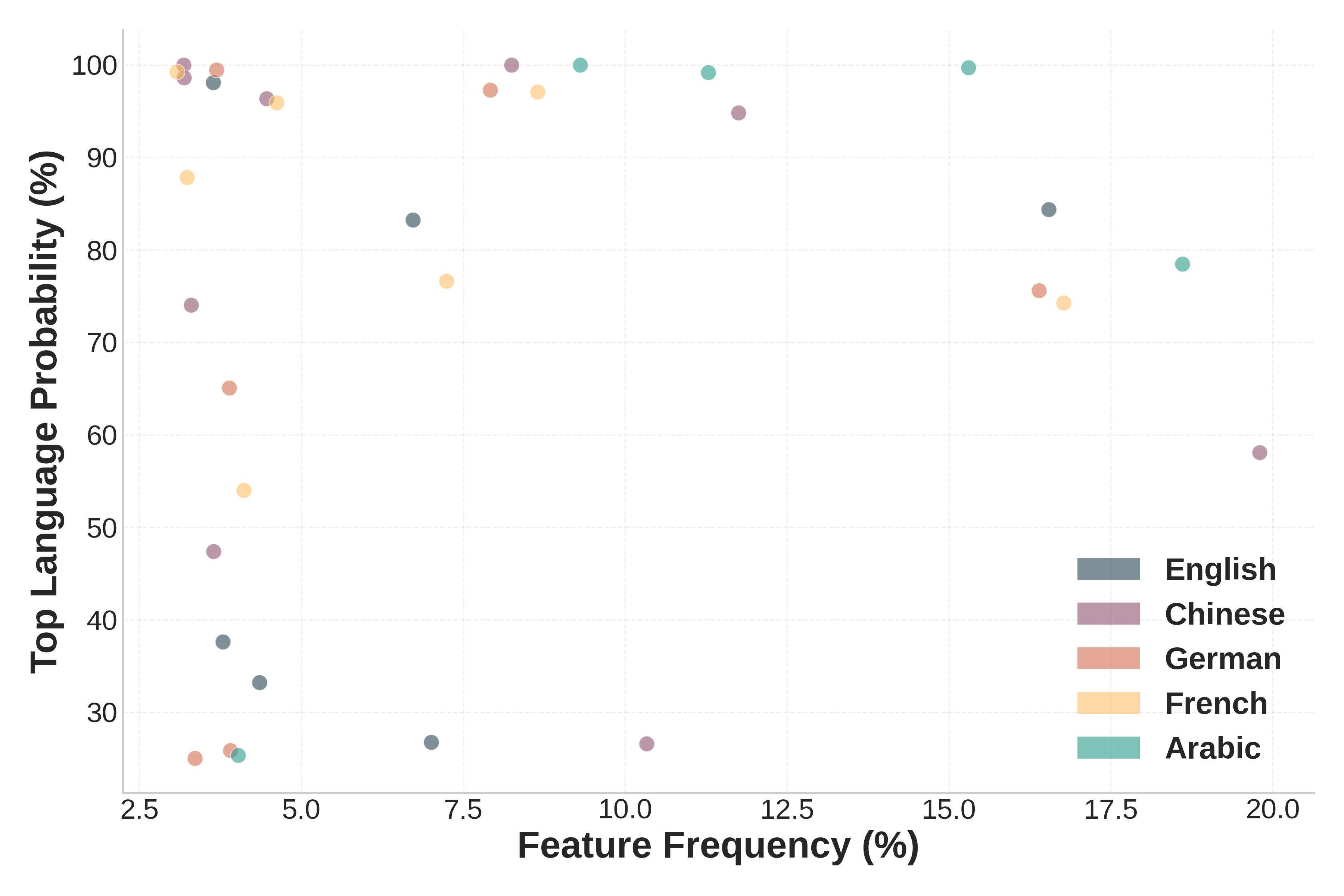}
     \caption{CLT features with the token activation frequency above 5\% for the 20\% model vs the probability in their top activating language. This shows that most high-frequency features are language features. }
    \label{fig:scatter}
\end{figure}

\section{Multilingual Score}
\label{appendix:comparaison_multilingual_score}

\begin{figure*}[h!]
    \centering
    \includegraphics[width=1.\textwidth,trim=0 0 0 25, clip]{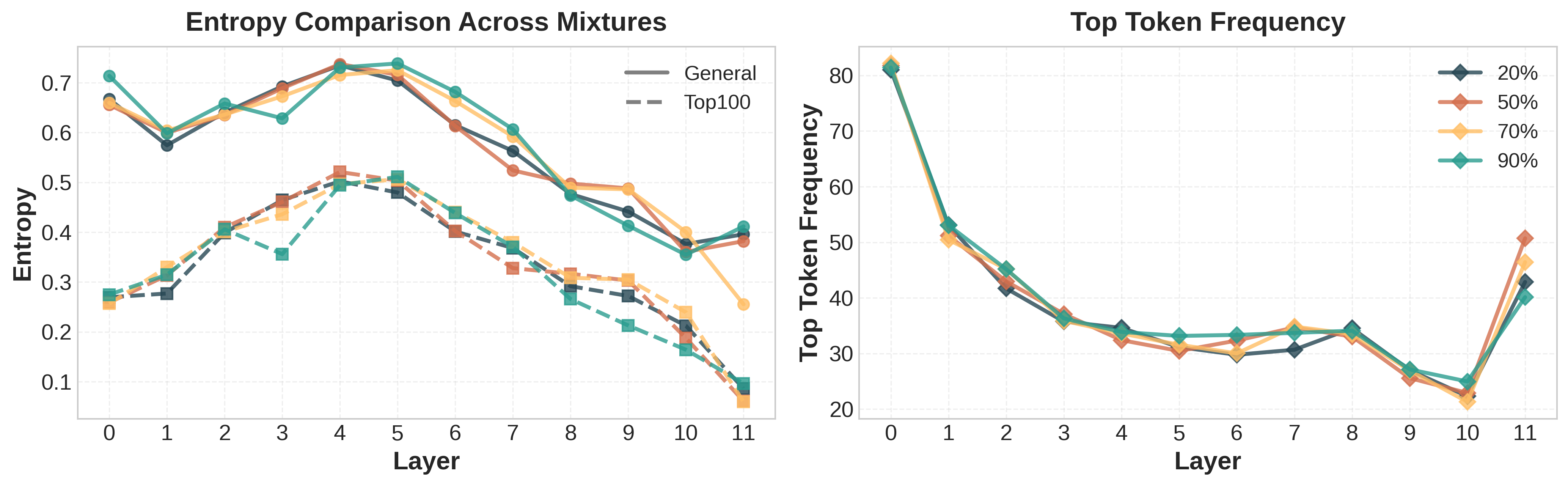}
    \caption{Comparaison between the two multilingual entropy measures. One computed on the top100 activated sequences for each feature and one computed over a large training subset. On the right the frequency of activation of the most frequent token over the top100 sequences. }
    \label{fig:comparaison_multilingual_score}
\end{figure*}

We compare the top100 multilingual score with the multilingual score computed over a large training subset (4M tokens)(Figure~\ref{fig:comparaison_multilingual_score}). Both entropy measures show the same overall trend across layers, with the top100 entropy consistently shifted toward lower values. The main difference appears in the earliest layers. This is likely because many features in these layers correspond to single-token activations, especially in layer~0. Given that the total number of features is limited to approximately 28K and multiple languages are involved, some features that mostly activate for a single token also tend to activate for other unrelated tokens.  

Figure~\ref{fig:comparaison_multilingual_score} also shows the frequency of the most activated token for each feature across the top100 sentences where that feature is strongest. We observe that in layer~0, and to a lesser extent in layers~1 and~2, most activations correspond to single tokens. For this reason, we decide to report the top100 score on the graphs. 

\begin{figure*}[h!]
    \centering
    \includegraphics[width=1.\textwidth,trim=0 0 0 25, clip]{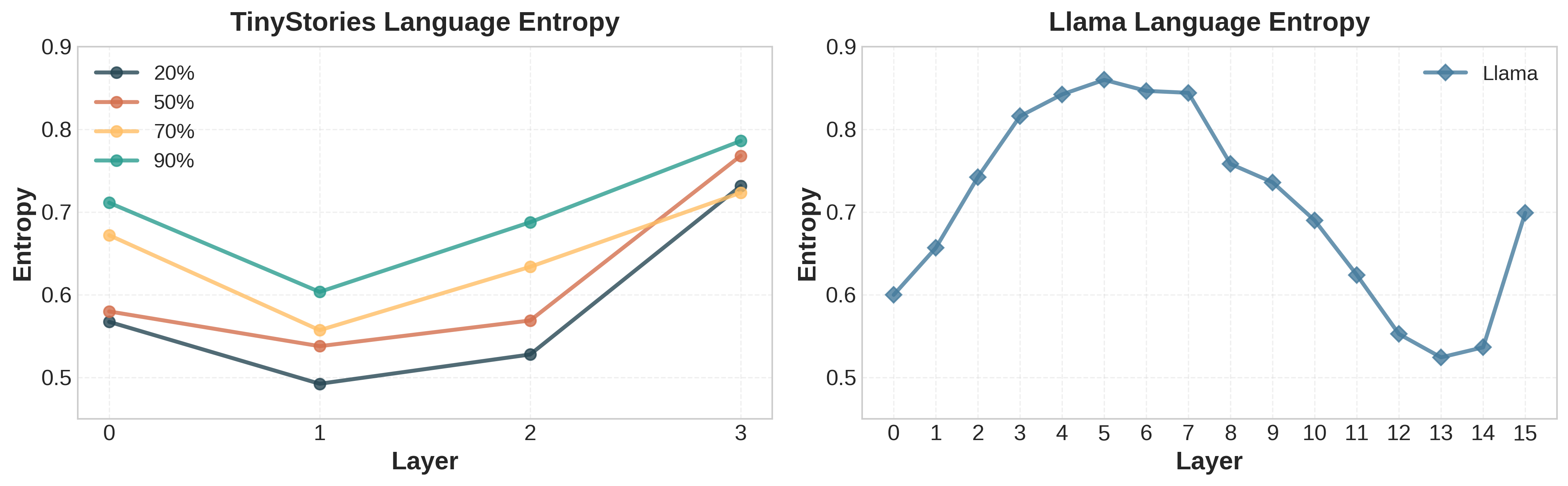}
    \caption{General multilingual score for the Tinystories model and Llama-3.2-1B. The final layer jump in the Llama CLT is potentially due to training dynamics. We find that the last layer has some features activated with small values that impact slightly the reconstruction performance but have a great impact on the multilingual score.}
    \label{fig:llama_multilingual_score}
\end{figure*}

We also report the general multilingual scores for Tinystories and Llama-3.2-1B in Figure~\ref{fig:llama_multilingual_score}.

\begin{figure*}[h!]
    \centering
    \includegraphics[width=1.\textwidth]{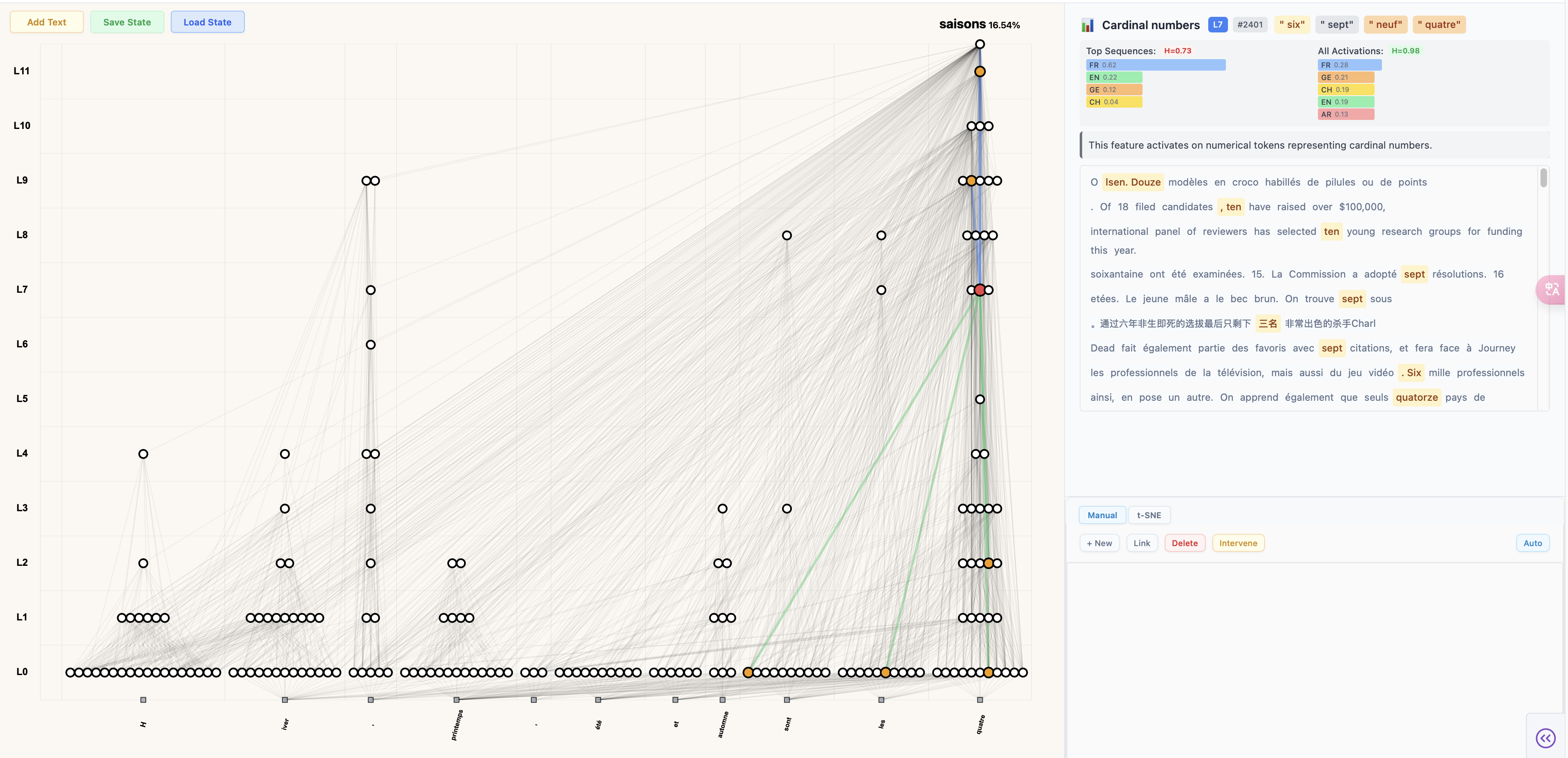}
    \caption{Graph visual interface with multilingual distributions and entropies on the top right.}
    \label{fig:visual_interface}
\end{figure*}

During graph analysis, we display the multilingual score and the language distributions on a graph visual interface as shown in Figure~\ref{fig:visual_interface}. 

\section{Shared Multilingual Space Independence from English Dominance}
\label{app:noenglish}

\begin{figure}[h]
\centering
\includegraphics[width=0.7\columnwidth]{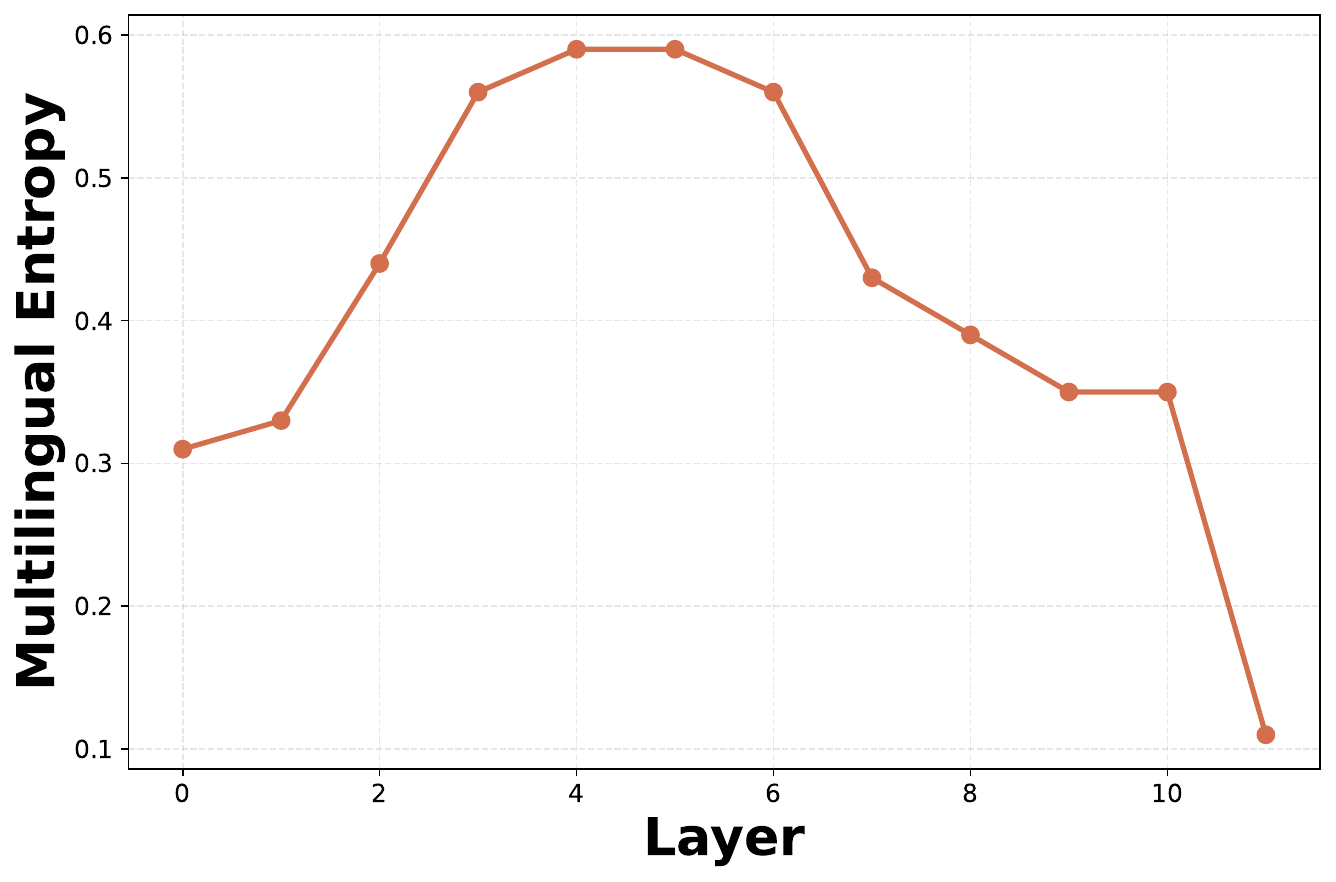}
\caption{The Shared Multilingual Space emergence is independent of English dominance. Model trained without English (FR/DE/AR/ZH, 25\% each) shows identical entropy curve to English-dominant model (20\% EN + others), with characteristic rise-and-fall pattern and peak in layers 5--6.}
\label{fig:pivot}
\end{figure}

To validate that the shared multilingual space structures do not require English dominance, we trained GPT-2 on balanced non-English data (25\% French, 25\% German, 25\% Arabic, 25\% Chinese) and computed weighted multilingual entropy across layers. The resulting curve exhibits the identical characteristic rise-and-fall pattern observed in English-dominant models, with peak entropy in layers 5--6 (Figure~\ref{fig:pivot}). This demonstrates that the shared multilngual space emergence is a fundamental architectural property, not an artifact of English dominance.

\section{Case Studies: Multilingual Circuits}
\label{app:multi-case-study}

Throughout this section, we present circuits to validate and visualize the findings discussed in the main text. Specifically, we organize our presentation as follows:

i) Multilingual validation: We show circuits across additional tasks that support the multilingual shared space finding described in Section~\ref{subsec:multiling_circuits}.

ii) Failure cases and interventions: We provide examples of model failures and demonstrate that interventions on circuits—primarily those that strengthen existing internal features—can successfully guide the model to produce correct outputs. This completes Section~\ref{subsec:circuit_examples}

iii) Translation and cultural prompts case study: We further validate our results with a case study on LLaMA-3.2-1B, using translation and culturally contextualized prompts, which confirms the robustness of our findings.

All examples produce clusters of circuits, which are manually constructed from features whose auto-interpretability descriptions and top-activating examples suggest similar functions. We further validate these clusters by intervening on them and confirming that the observed effects align with their interpreted descriptions.

\subsection{Multilingual Circuits: U-shaped Entropy}
\label{app:circuits}

In this appendix, we provide extended circuit visualizations complementing the analysis of Section~\ref{subsec:multiling_circuits}.  
For each example, we extract attribution graphs across layers and present the identified circuits.  
Depending on the setting, each figure contains either (i) four subfigures, corresponding to the four training data mixtures (90\%, 70\%, 50\% or 20\% English), or (ii) five subfigures, corresponding to the five languages (English, Arabic, German, French, Chinese).  

As observed in our main results, early and late layers predominantly encode language-specific circuits, while middle layers form clusters with high language entropy, reflecting multilingual representations. Exceptions arise for determinants and prepositions, whose circuits remain largely language-specific.

The top-right corner of each cluster shows its entropy score, defined as the average of the entropy values of the features it contains.

We now detail the examples analyzed.

\subsubsection{Preposition Sentences (English: \texttt{``It was a piece''})}

We begin by analyzing English preposition prediction across training mixtures. Prepositions, being primarily functional rather than semantic, offer insight into the model's handling of syntactic structures. In this case, we find that the model consistently relies on language-specific clusters.

\begin{figure*}[h!]
    \centering
    \begin{subfigure}[b]{0.49\textwidth}
        \centering
        \includegraphics[width=\textwidth,trim=0 1 0 0, clip]{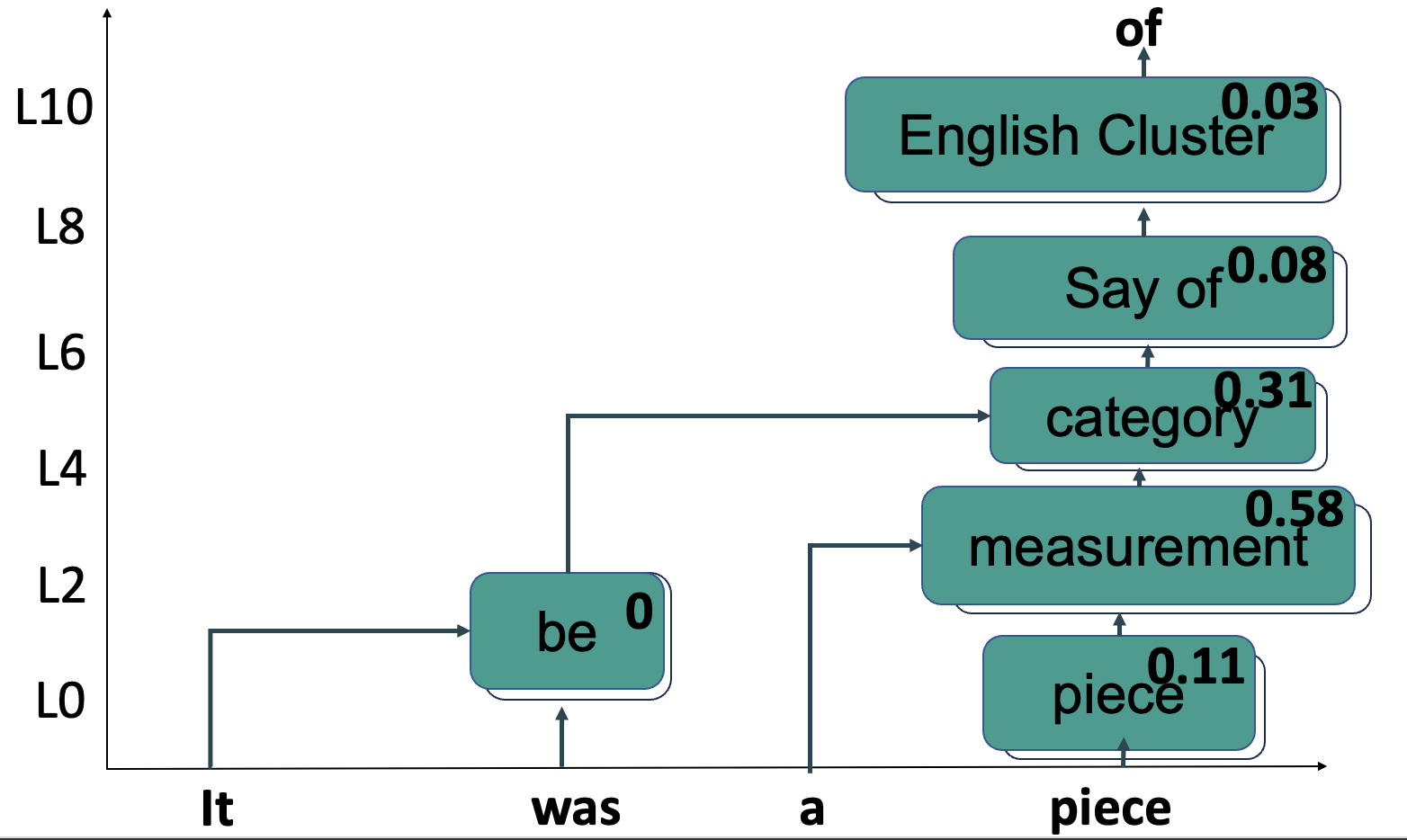}
        \caption{90\% mixture}
    \end{subfigure}
    \hfill
    \begin{subfigure}[b]{0.49\textwidth}
        \centering
        \includegraphics[width=\textwidth,trim=0 0 0 1, clip]{iclr2026/figures/of_70.png}
        \caption{70\% mixture}
    \end{subfigure}
    
    \vspace{0.3em}
    
    \begin{subfigure}[b]{0.49\textwidth}
        \centering
        \includegraphics[width=\textwidth]{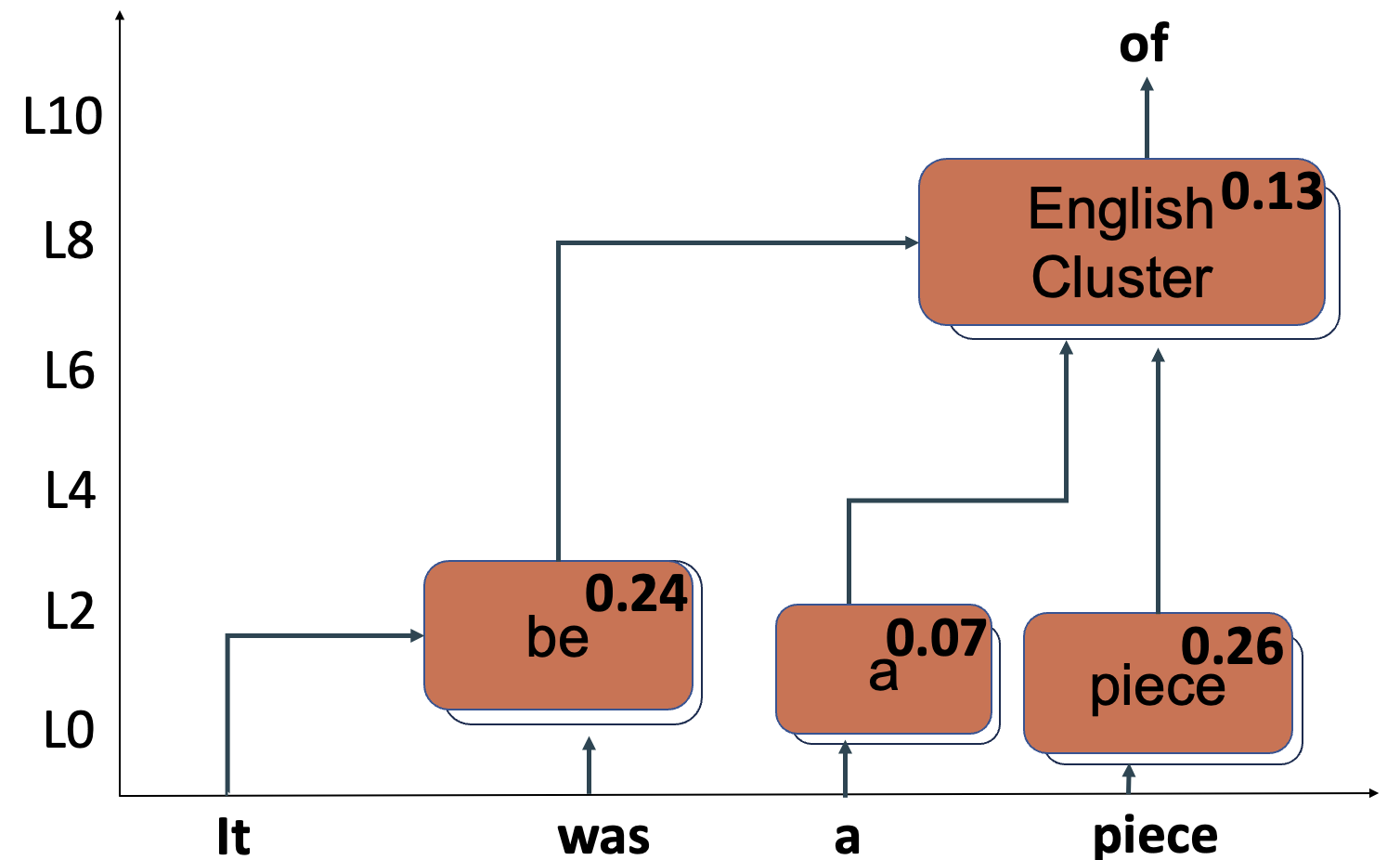}
        \caption{50\% English mixture}
    \end{subfigure}
    \hfill
    \begin{subfigure}[b]{0.49\textwidth}
        \centering
        \includegraphics[width=\textwidth,trim=0 0 0 1, clip]{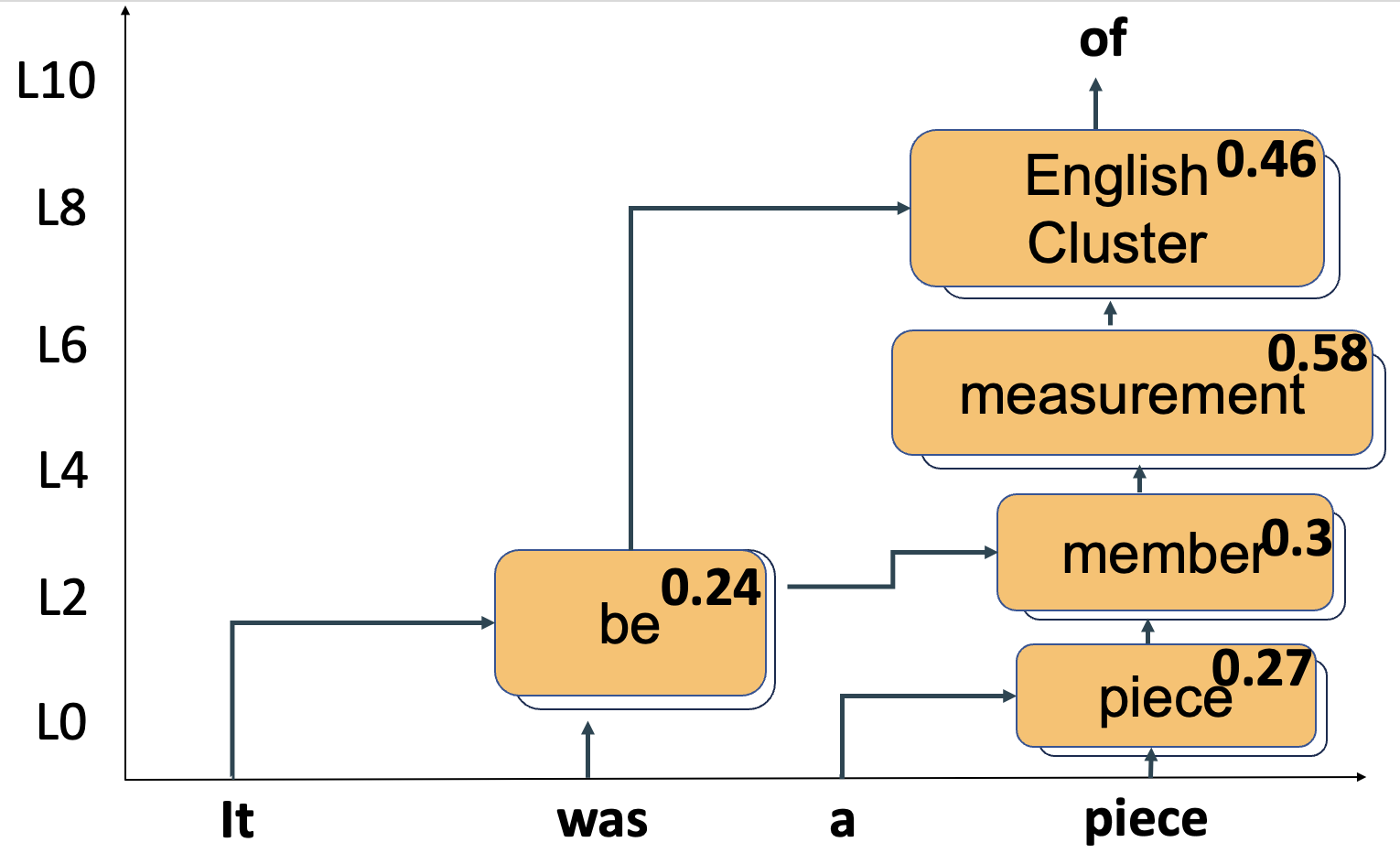}
        \caption{20\% multilingual mixture}
    \end{subfigure}
    
    \caption{Circuits for English preposition prediction (``It was a piece'') across training mixtures.}
    \label{fig:circuits_preposition_piece}
\end{figure*}

\subsubsection{Preposition Sentences (French: \texttt{``J'ai bu une tasse''})}

We next examine French preposition prediction across the training mixtures. This setting tests whether the patterns observed for English prepositions generalize to another language.

\begin{figure*}[h]
    \centering
    \begin{subfigure}[b]{0.49\textwidth}
        \centering
        \includegraphics[width=\textwidth,trim=0 1 0 0, clip]{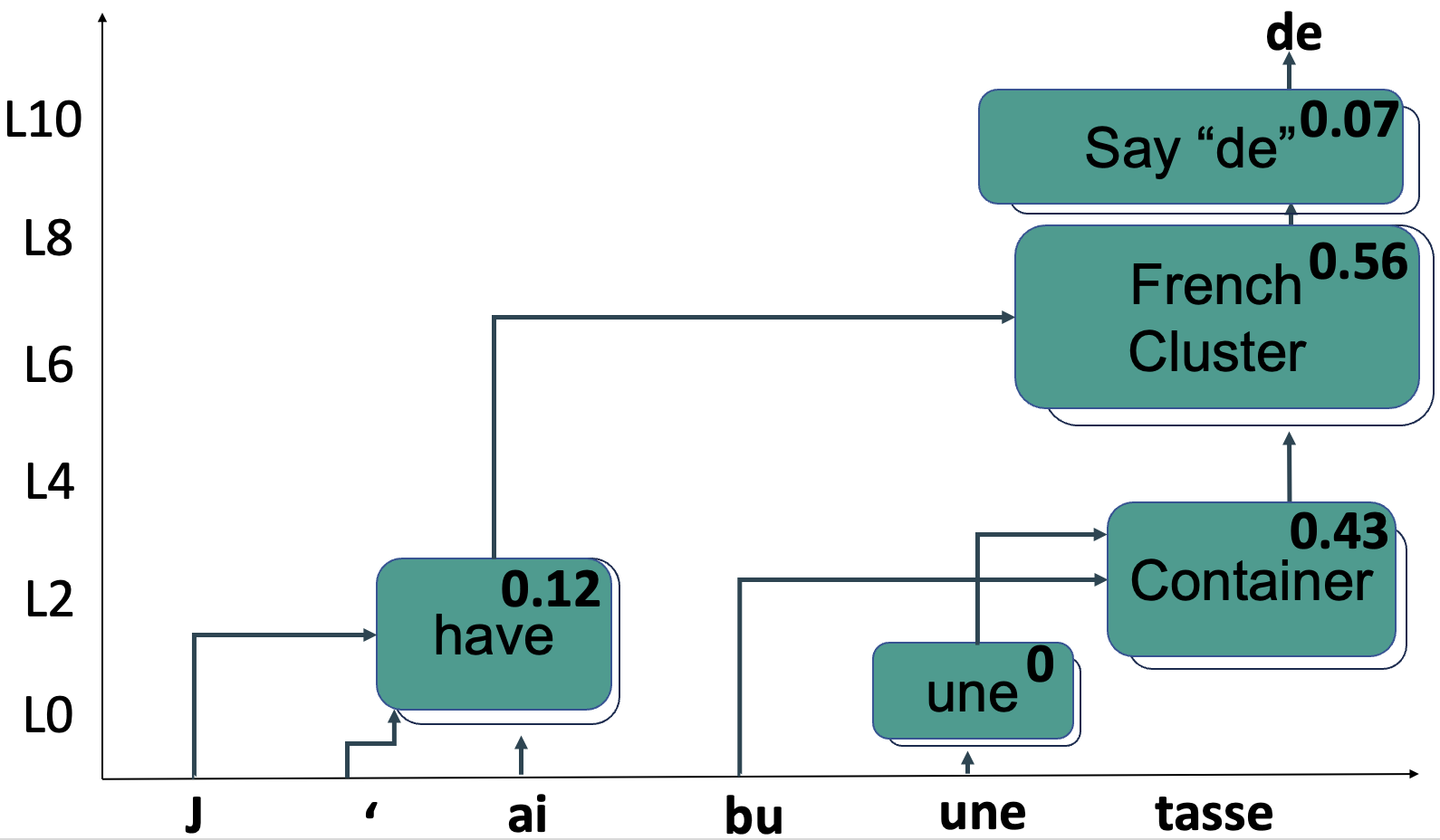}
        \caption{90\% mixture}
    \end{subfigure}
    \hfill
    \begin{subfigure}[b]{0.49\textwidth}
        \centering
        \includegraphics[width=\textwidth]{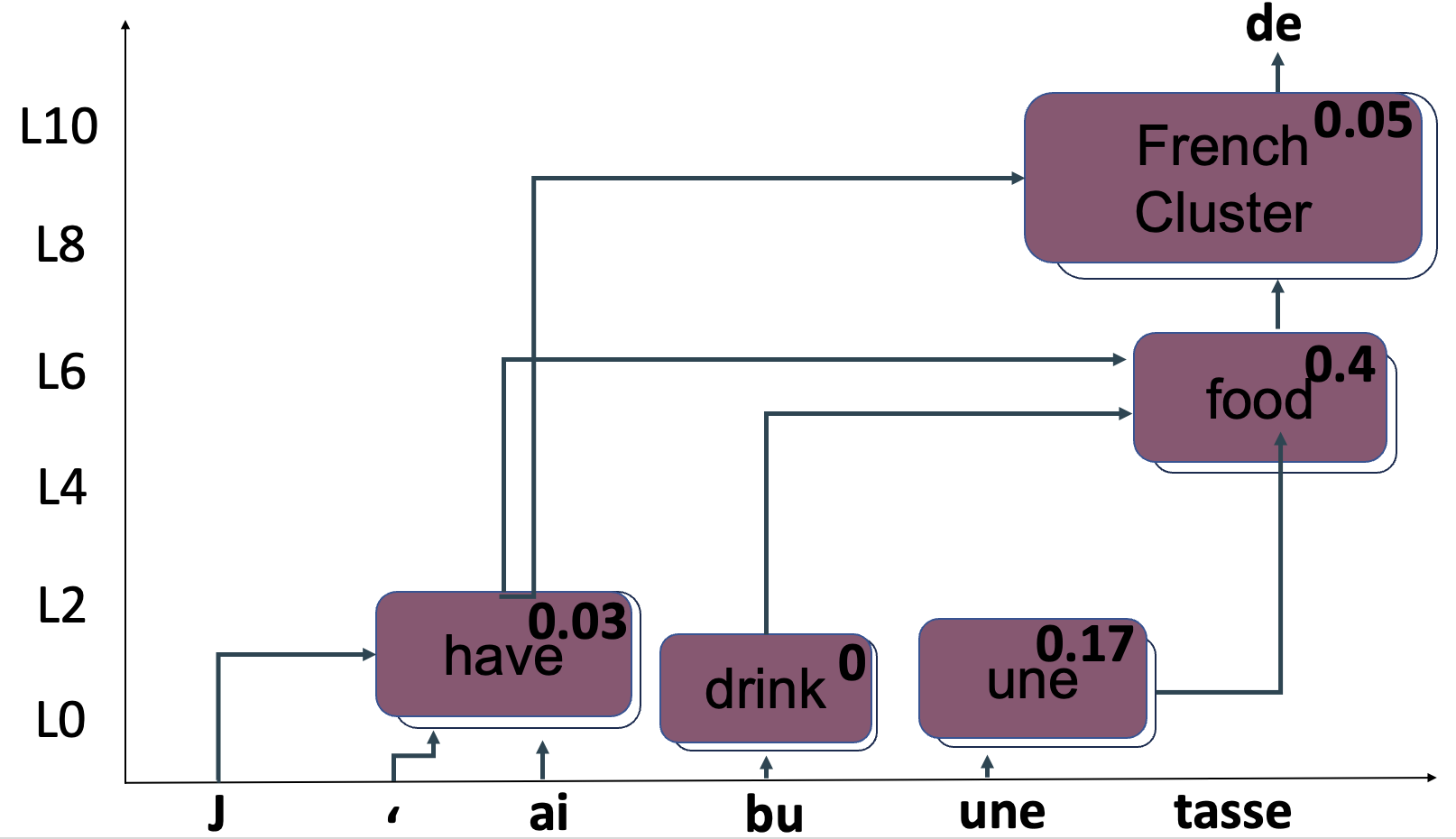}
        \caption{70\% mixture}
    \end{subfigure}
    
    \vspace{0.3em}
    
    \begin{subfigure}[b]{0.49\textwidth}
        \centering
        \includegraphics[width=\textwidth]{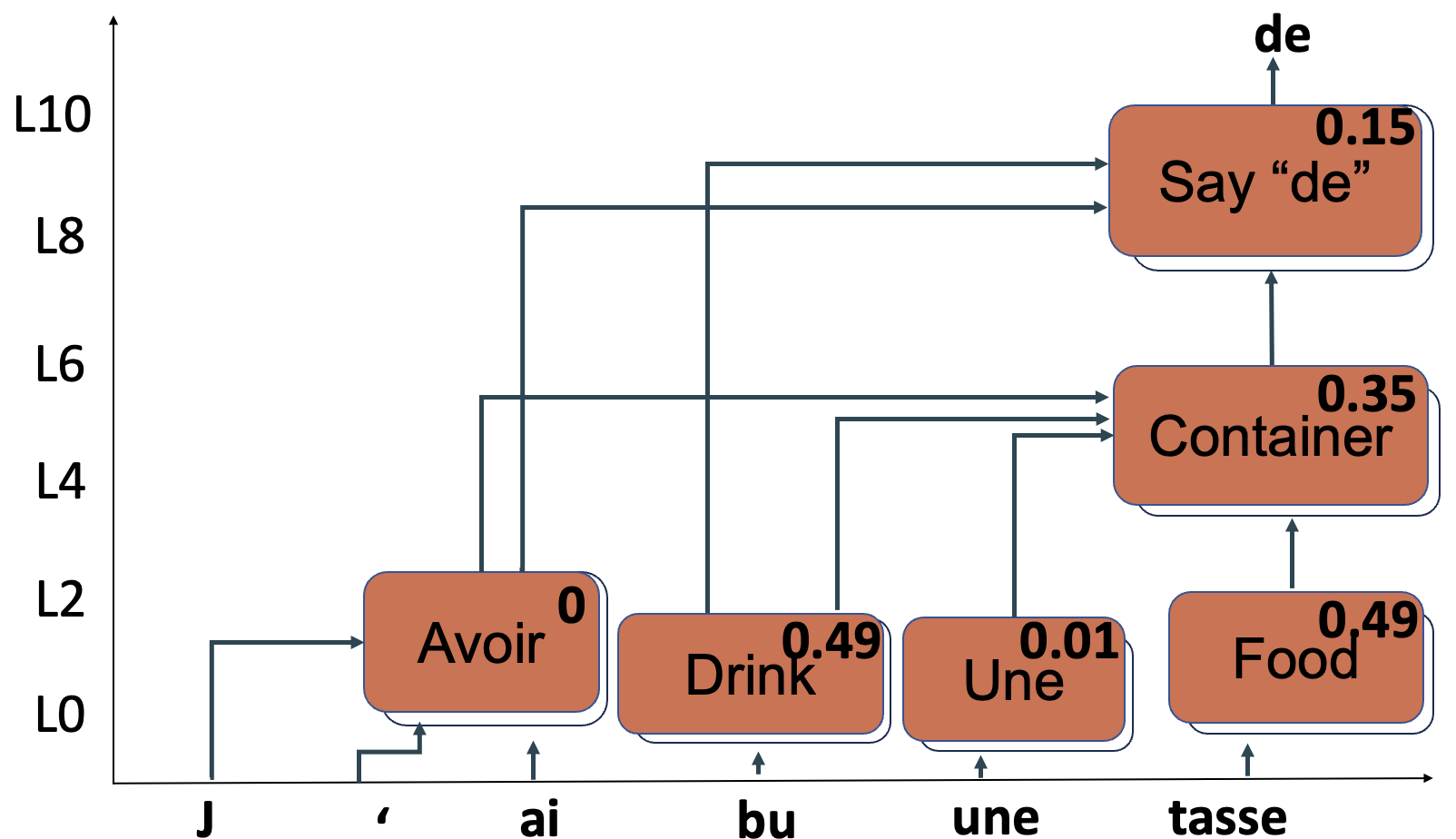}
        \caption{20\% mixture}
    \end{subfigure}
    \hfill
    \begin{subfigure}[b]{0.49\textwidth}
        \centering
        \includegraphics[width=\textwidth]{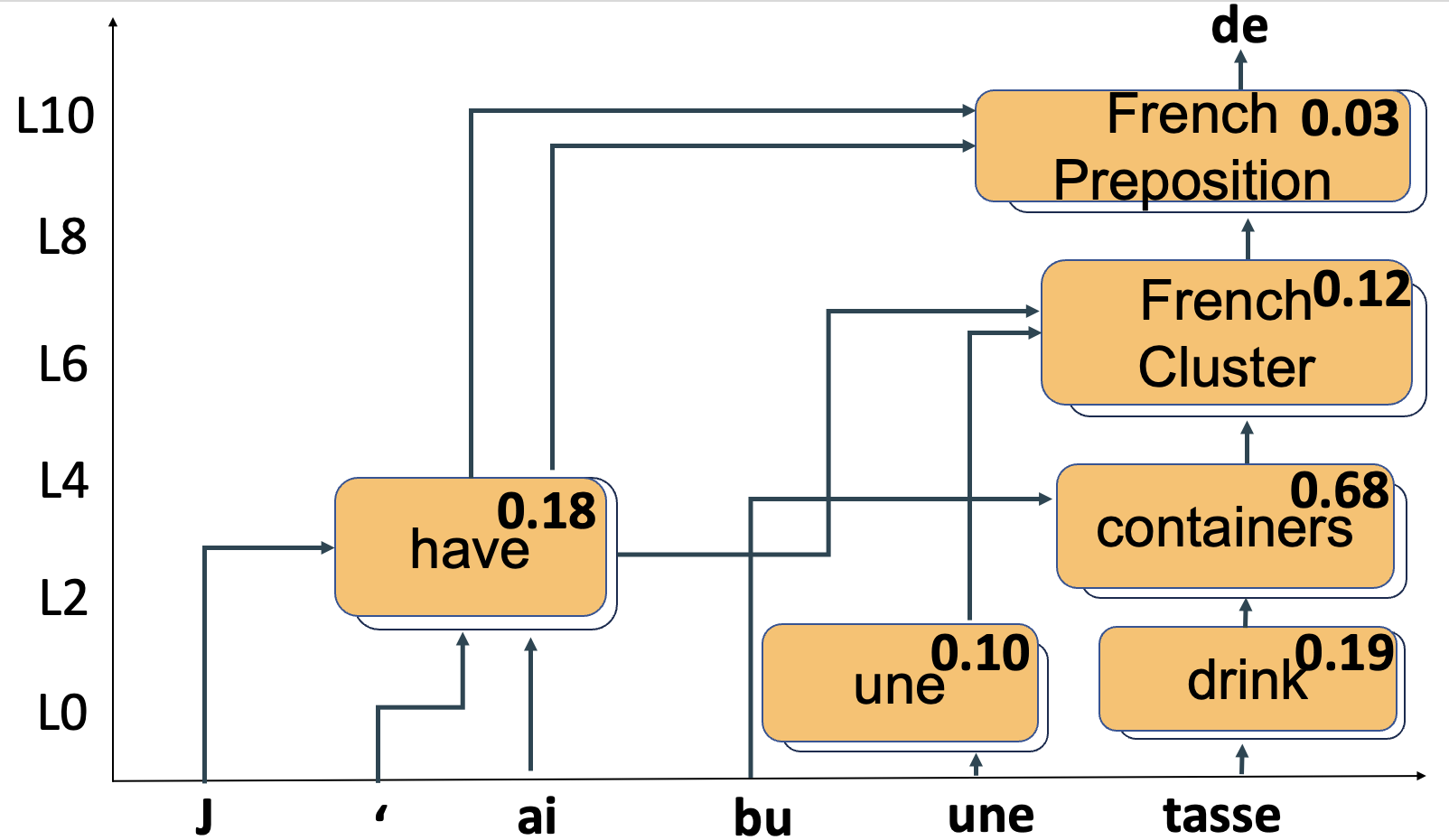}
        \caption{70\% mixture}
    \end{subfigure}    
    \caption{Circuits for French preposition prediction (``J'ai bu une tasse'') across training mixtures.}
    \label{fig:circuits_preposition_fr}
\end{figure*}

\subsubsection{Content Word Sentences (English: \texttt{``I prefer drinking tea to drinking''})}
Next, we analyze content word prediction across languages in the 90\% English mixture.  
Content words engage semantic and contextual representations, making them a strong testbed for cross-lingual generalization.  
We find robust multilingual clusters in middle layers across all five languages, even when trained under extreme imbalance.  

\begin{figure*}[h]
    \centering
    \begin{subfigure}[b]{0.49\textwidth}
        \centering
        \includegraphics[width=\textwidth,trim=0 1 0 2, clip]{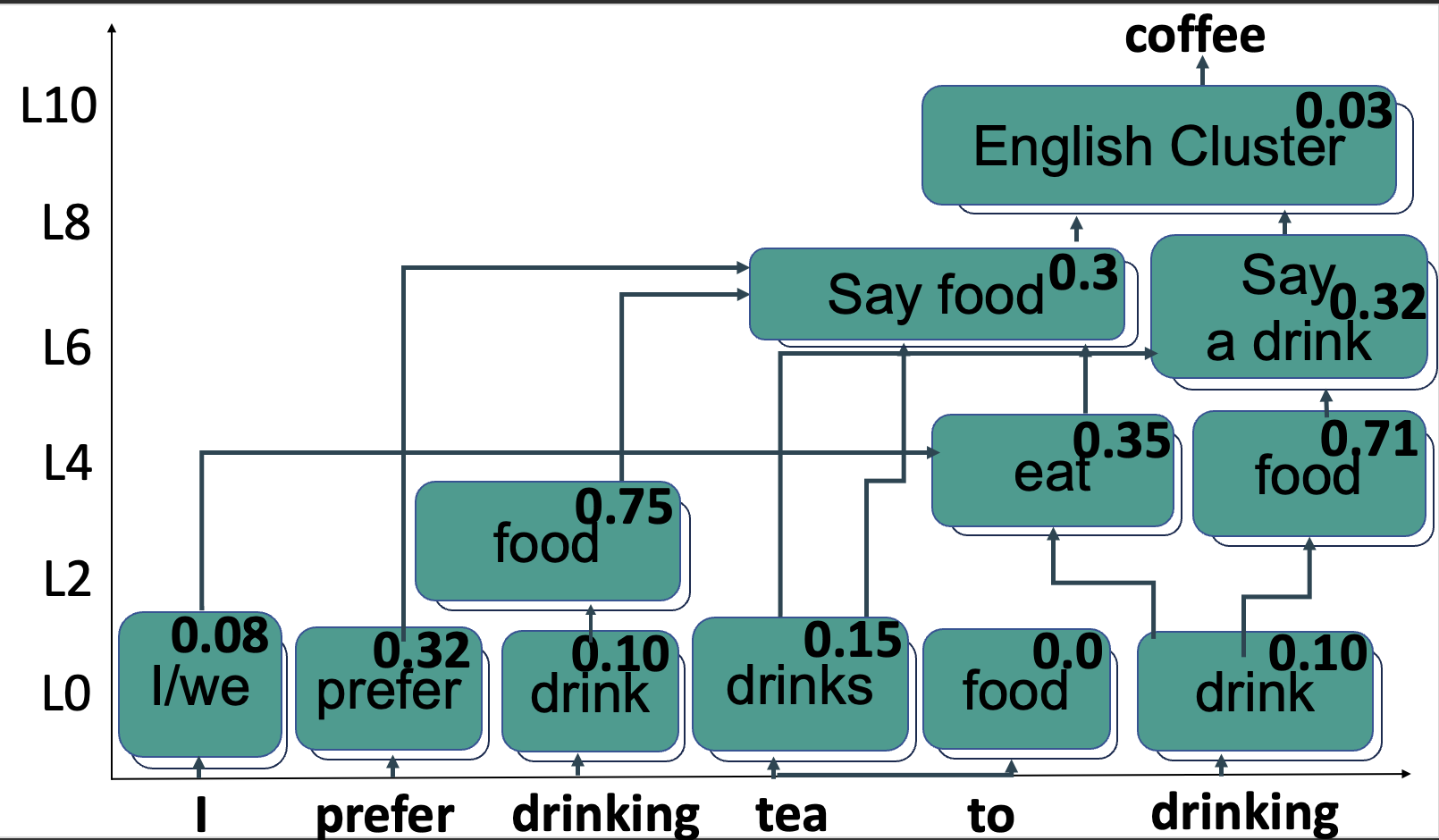}
        \caption{English}
    \end{subfigure}
    \hfill
    \begin{subfigure}[b]{0.49\textwidth}
        \centering
        \includegraphics[width=\textwidth]{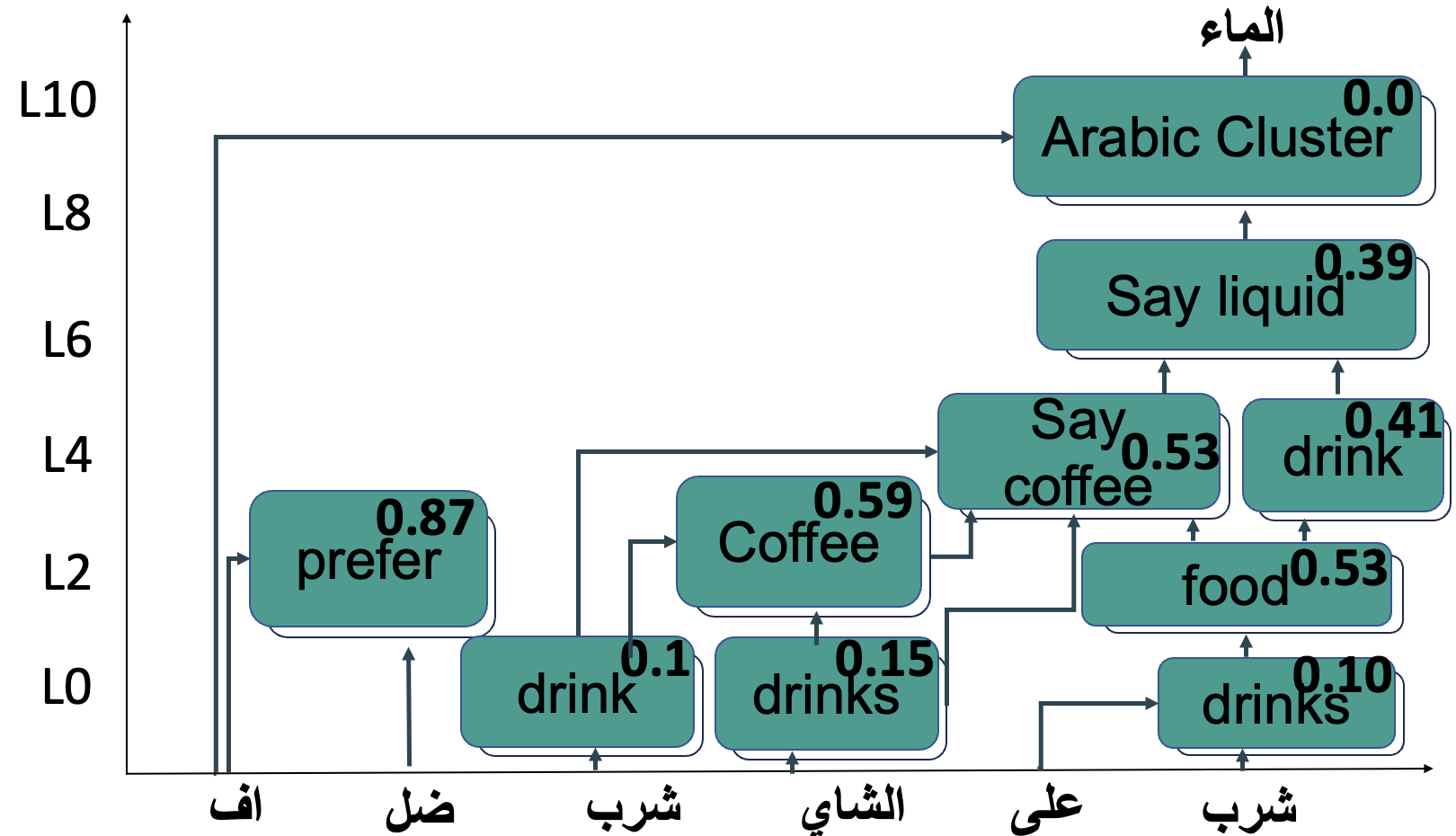}
        \caption{Arabic}
        \label{fig:arabic_fail_1}
    \end{subfigure}
    
    \vspace{0.3em}
    
    \begin{subfigure}[b]{0.49\textwidth}
        \centering
        \includegraphics[width=\textwidth,trim=0 1 0 1, clip]{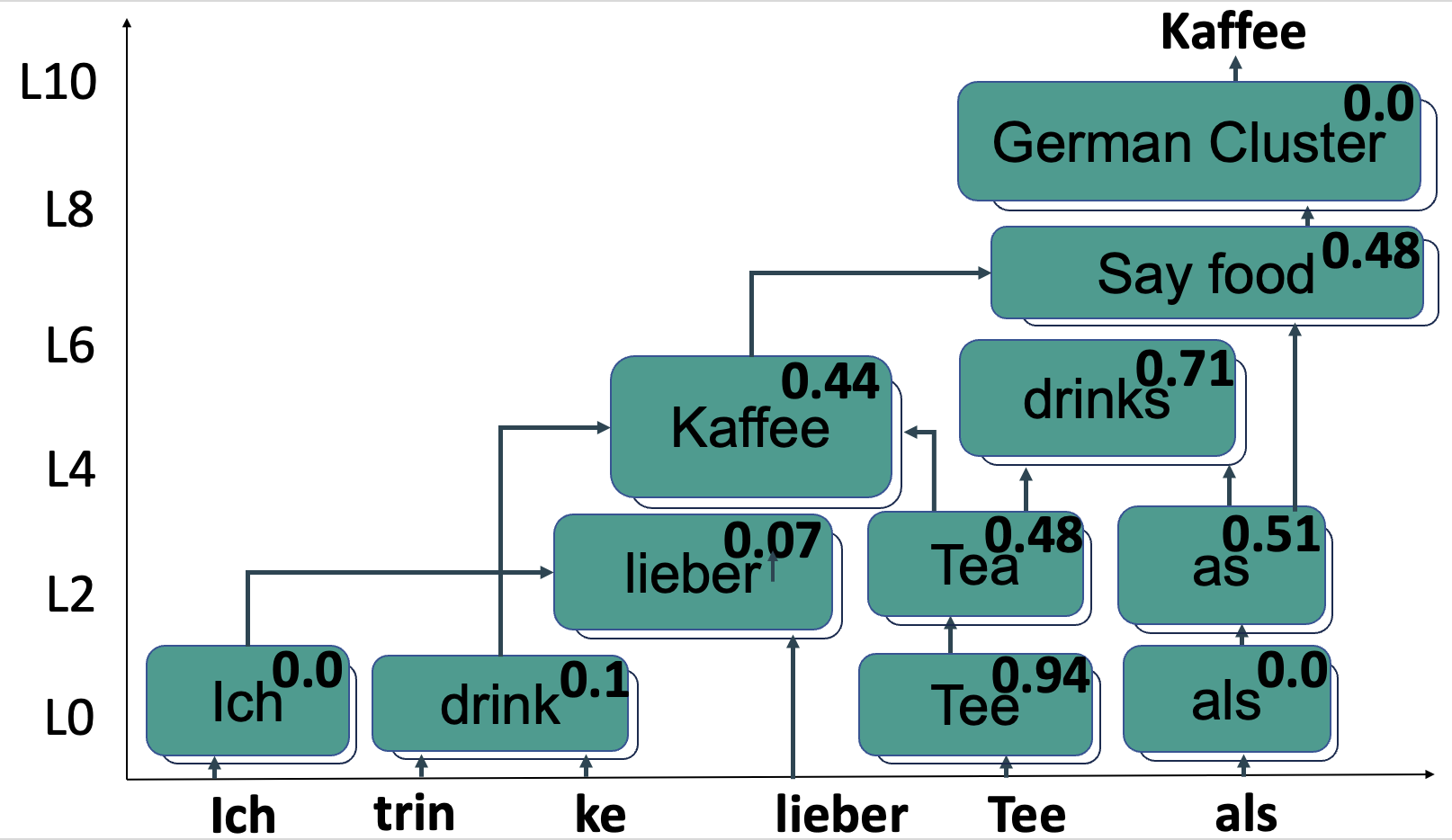}
        \caption{German}
    \end{subfigure}
    \hfill
    \begin{subfigure}[b]{0.49\textwidth}
        \centering
        \includegraphics[width=\textwidth,trim=0 1 0 0, clip]{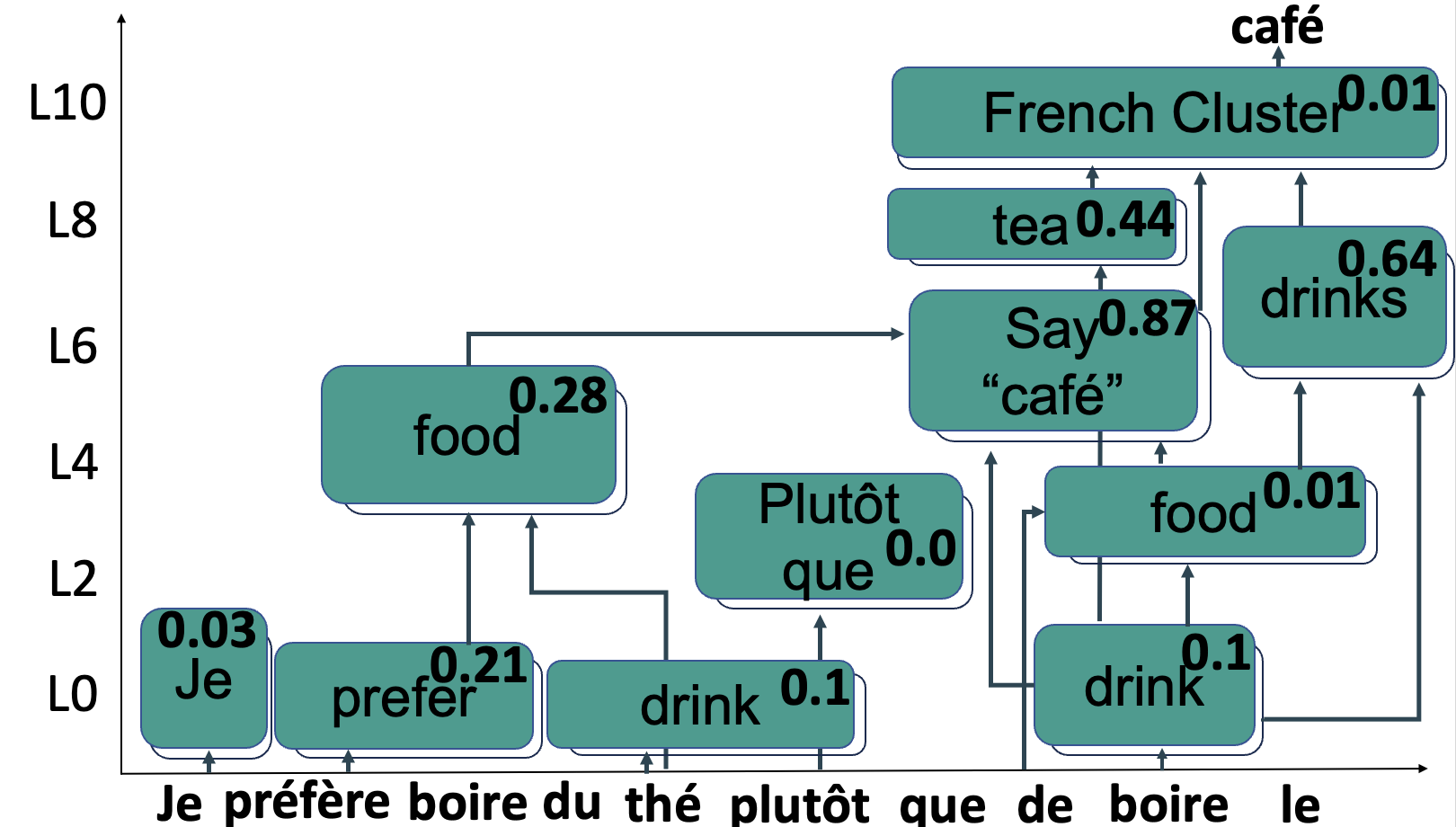}
        \caption{French}
    \end{subfigure}
    
    \vspace{0.3em}
    
    \begin{subfigure}[b]{0.49\textwidth}
        \centering
        \includegraphics[width=\textwidth,trim=0 0 0 1.5, clip]{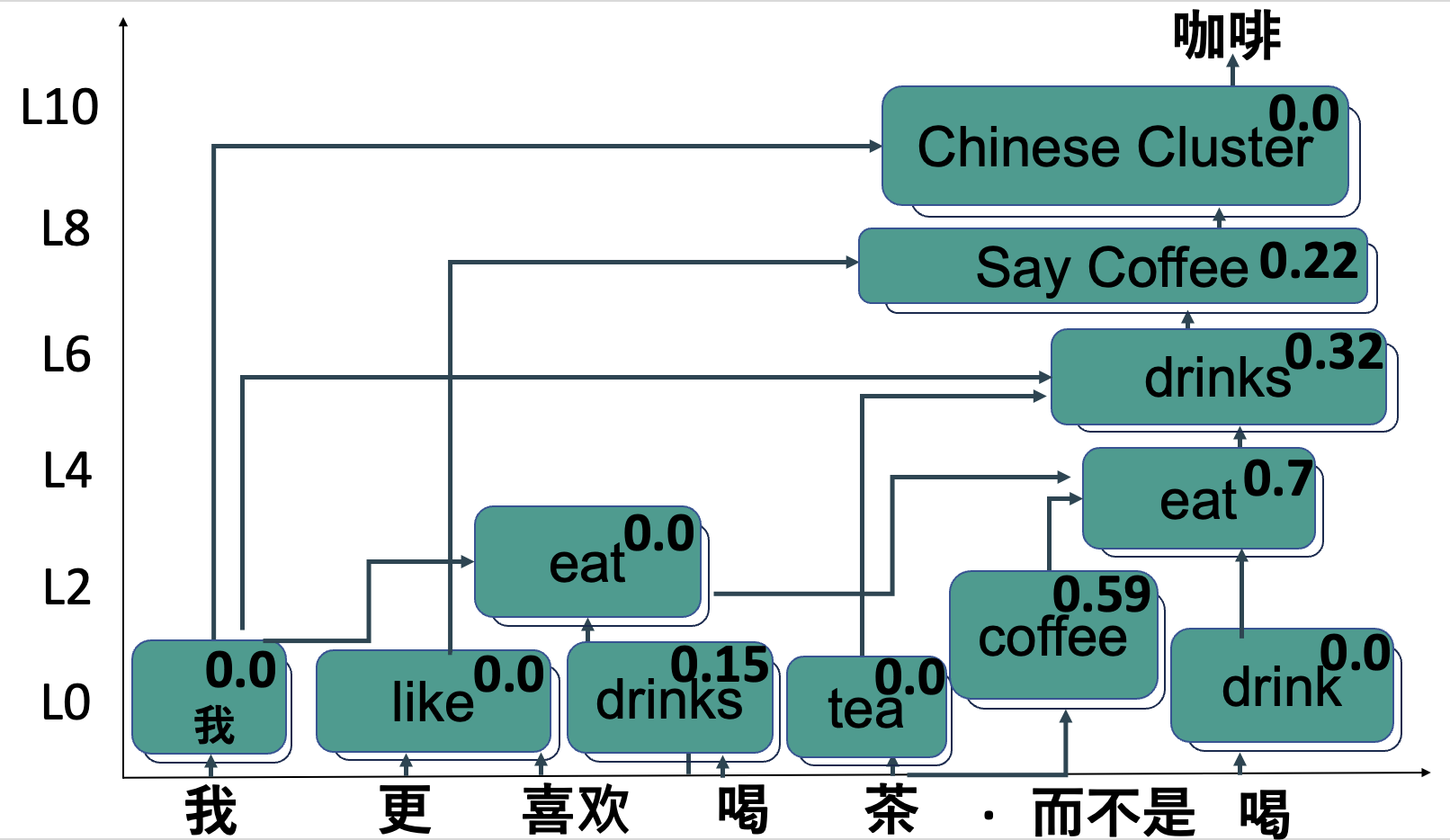}
        \caption{Chinese}
    \end{subfigure}
    
    \caption{Circuits for content word prediction (``I prefer drinking tea to drinking'') across languages in the 90\% English mixture.}
    \label{fig:circuits_content_tea}
\end{figure*}

\subsubsection{Content Word Sentences: \texttt{``Winter, spring, summer, and autumn are the four''}}
We analyze the prediction of content words in the sentence ``Winter, spring, summer, and autumn are the four'' across languages in the 20\% English mixture.  
This example probes semantic and sequential reasoning across multiple languages, highlighting whether the model reuses internal circuits for similar content words.  
We find that middle layers consistently form multilingual clusters across all five languages, while early and late layers remain largely language-specific.  

\begin{figure*}[h]
    \centering
    \begin{subfigure}[b]{0.49\textwidth}
        \centering
        \includegraphics[width=\textwidth]{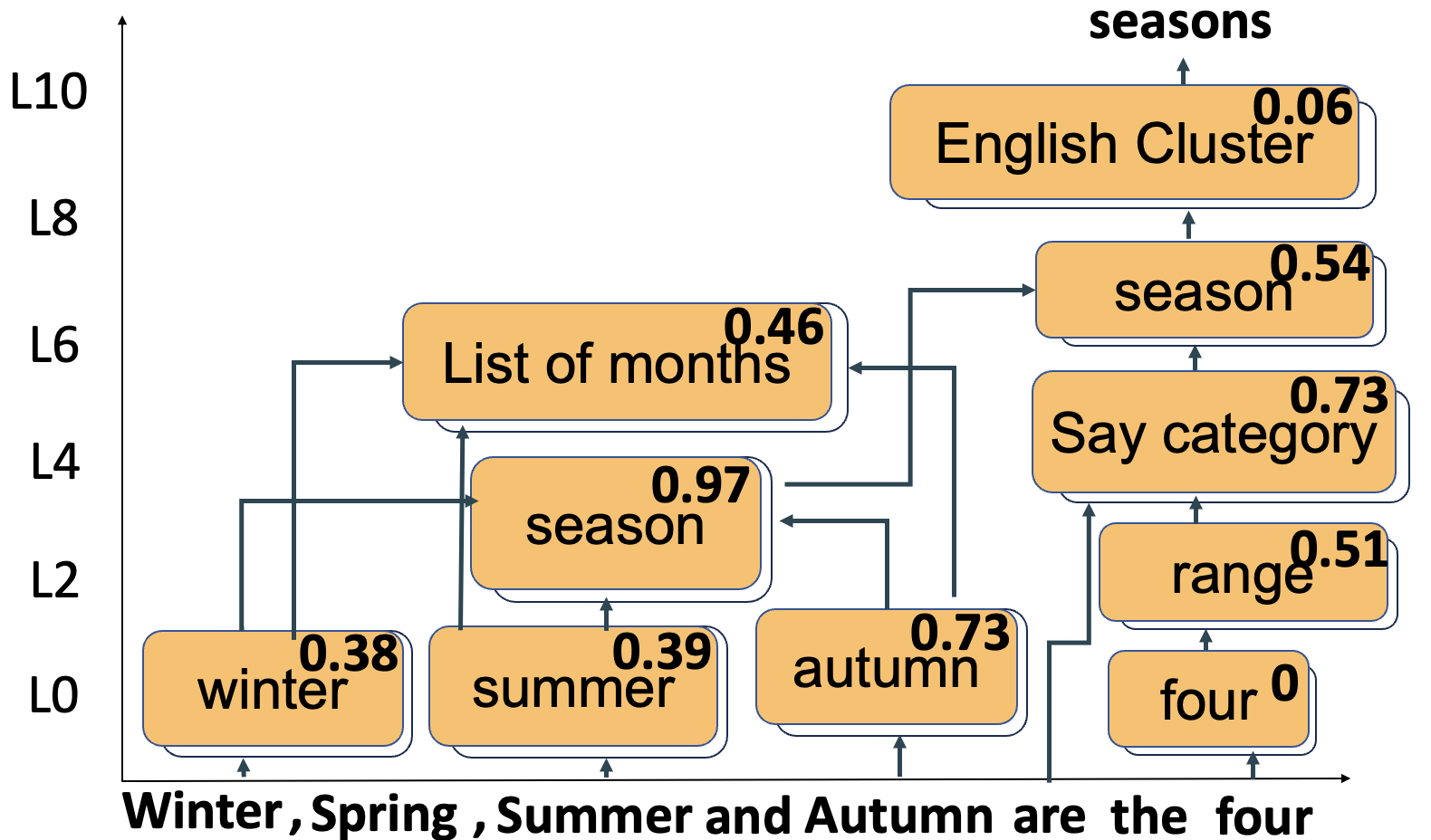}
        \caption{English}
    \end{subfigure}
    \hfill
    \begin{subfigure}[b]{0.49\textwidth}
        \centering
        \includegraphics[width=\textwidth]{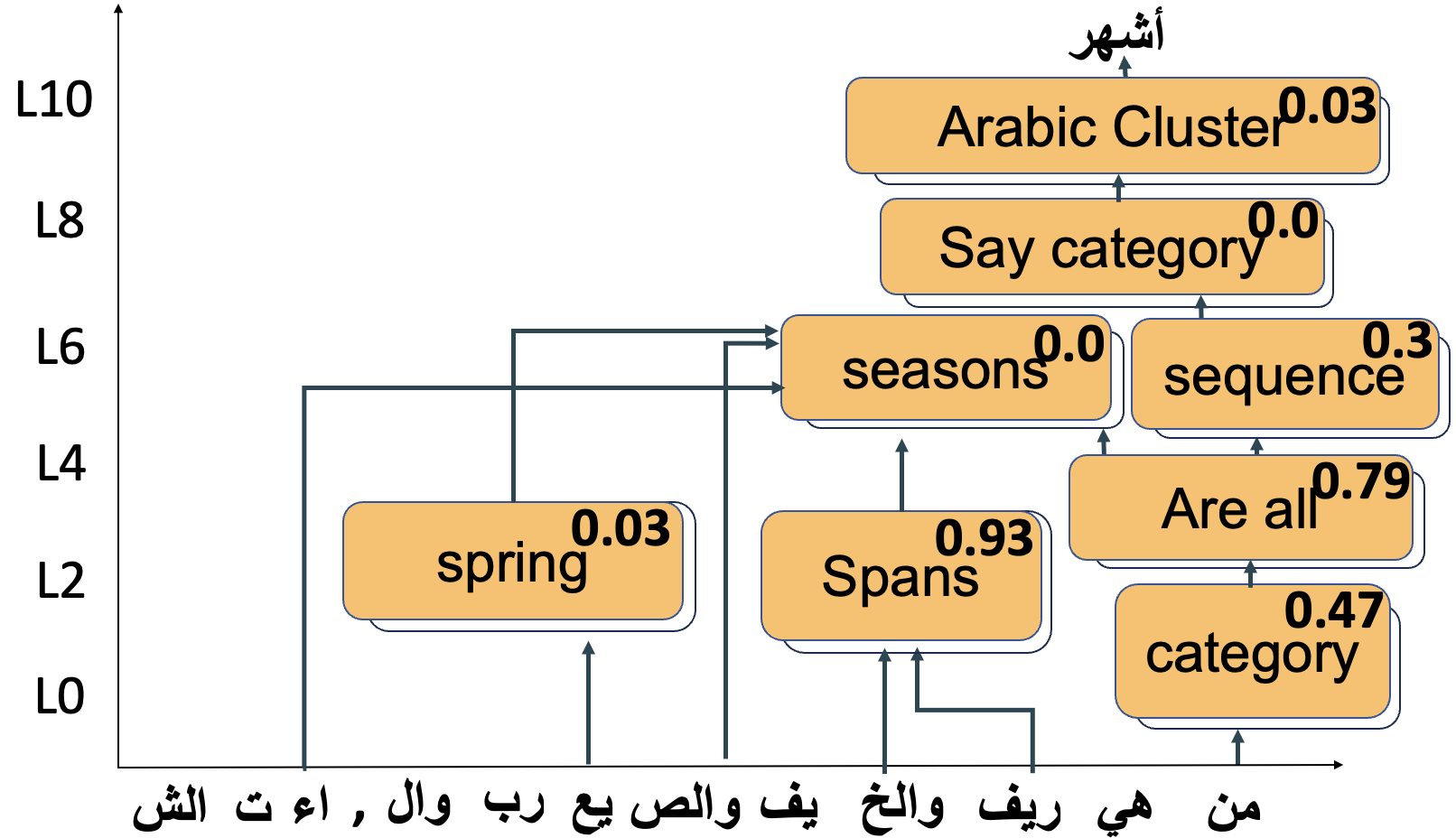}
        \caption{Arabic}
        \label{fig:arabic_fail_2}
    \end{subfigure}
    
    \vspace{0.3em}
    
    \begin{subfigure}[b]{0.49\textwidth}
        \centering
        \includegraphics[width=\textwidth,trim=0 1 0 1, clip]{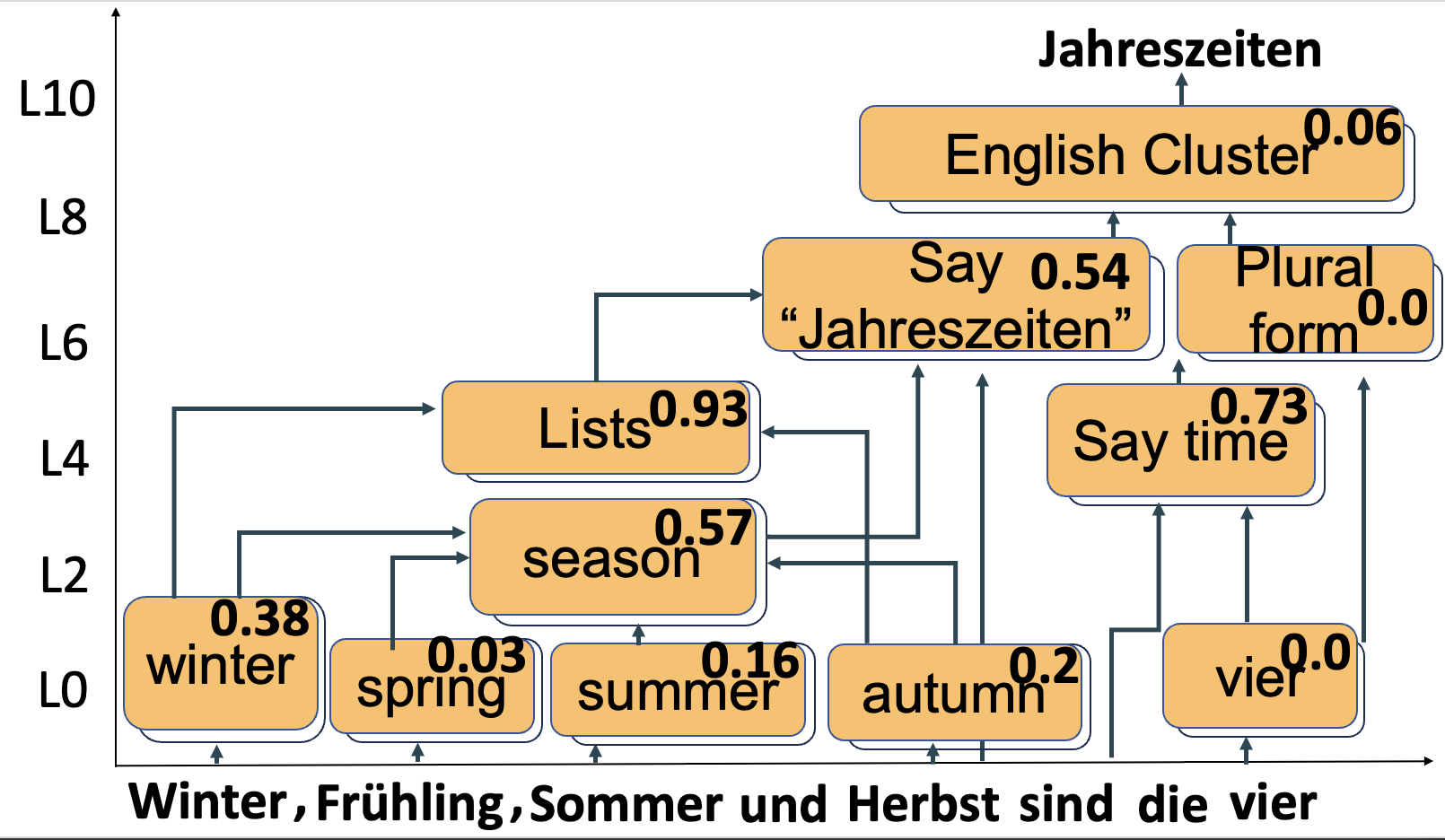}
        \caption{German}
    \end{subfigure}
    \hfill
    \begin{subfigure}[b]{0.49\textwidth}
        \centering
        \includegraphics[width=\textwidth,trim=0 1 0 1, clip]{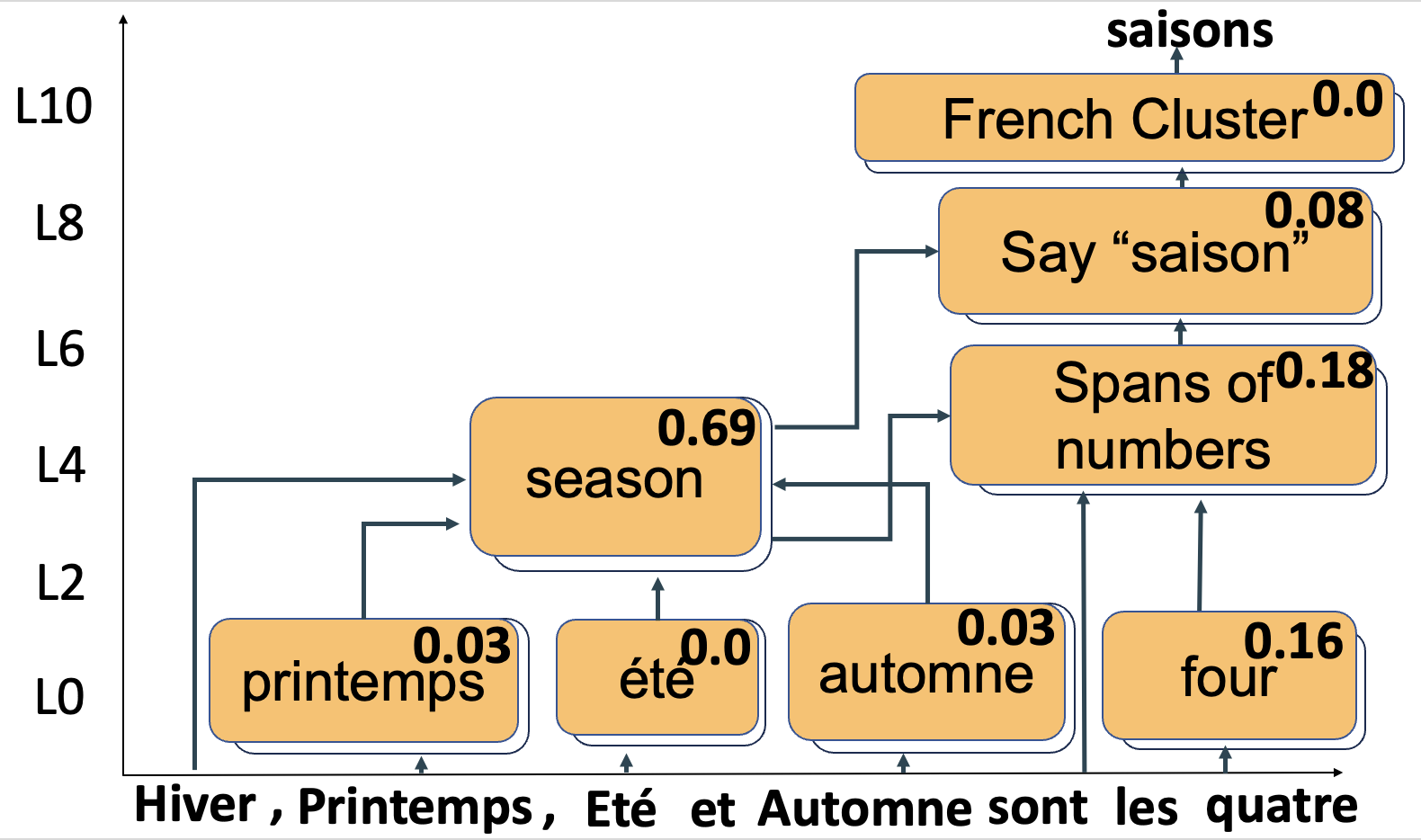}
        \caption{French}
    \end{subfigure}
    
    \vspace{0.3em}
    
    \begin{subfigure}[b]{0.49\textwidth}
        \centering
        \includegraphics[width=\textwidth,trim=0 1 0 1, clip]{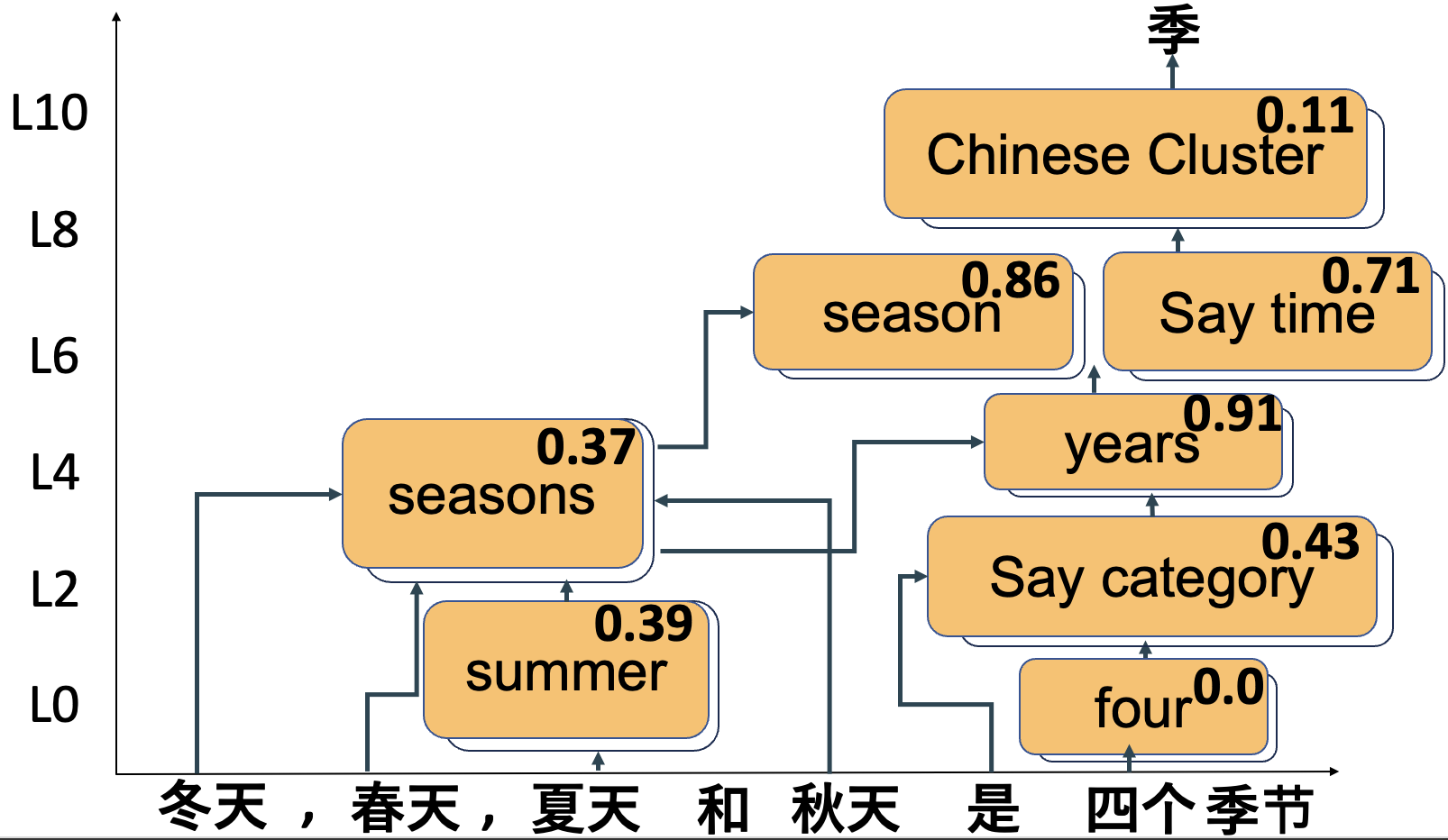}
        \caption{Chinese}
    \end{subfigure}
    
    \caption{Circuits for content word prediction (``Winter, spring, summer, and autumn are the four'') across five languages in the 20\% English mixture.}
    \label{fig:circuits_seasons}
\end{figure*}

\subsubsection{Calendar Term Prediction (\texttt{``Monday, Tuesday, Wednesday, Thursday''})}
We study calendar term prediction across mixtures. Calendar terms primarily involve lexical memorization, providing a probe for multilingual circuit reuse. Across all mixtures, middle layers consistently form multilingual clusters, while early and late layers remain more language-specific.  

\begin{figure*}[h]
    \centering
    \begin{subfigure}[b]{0.49\textwidth}
        \centering
        \includegraphics[width=\textwidth,trim=0 0.5 0 1, clip]{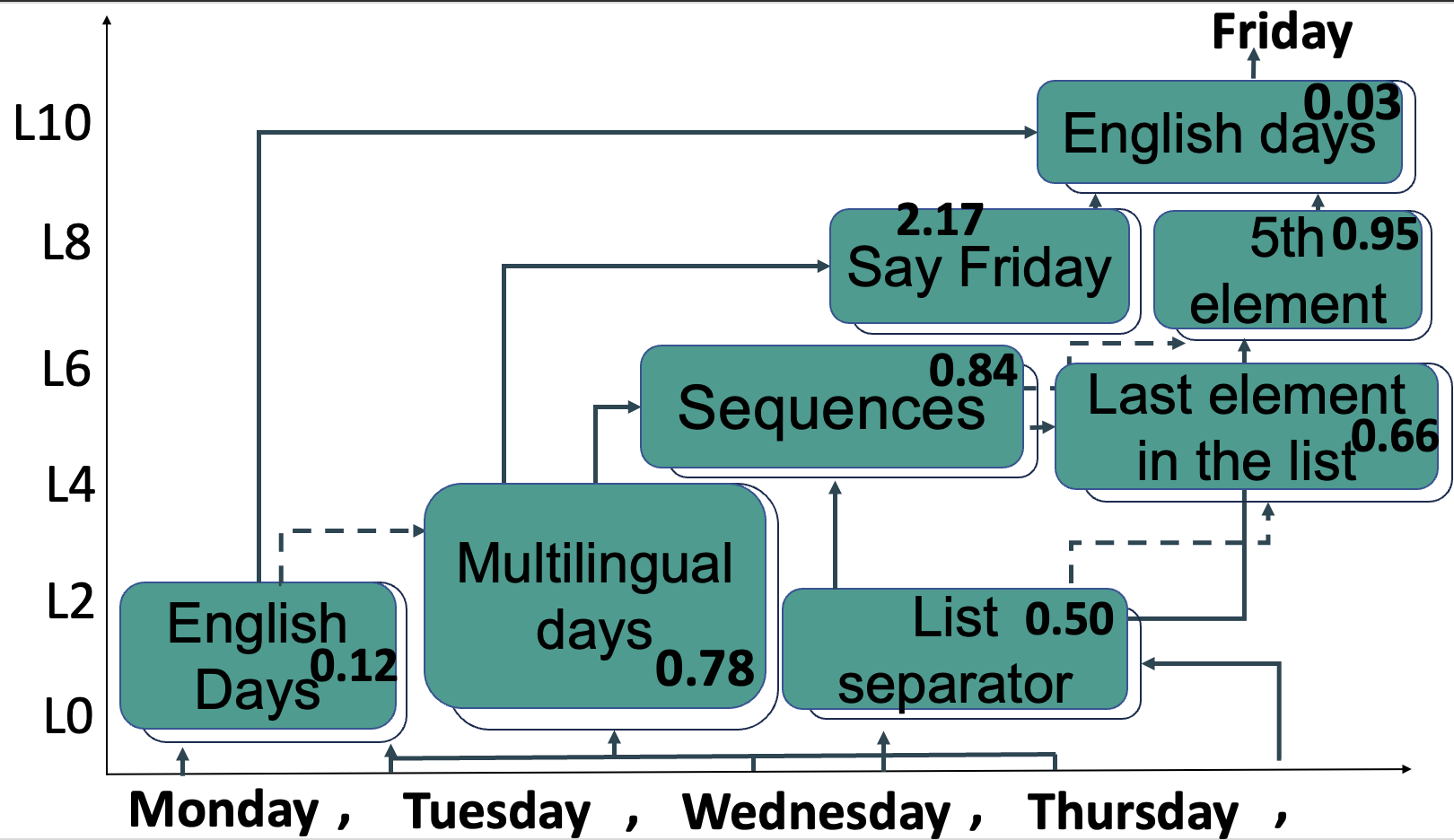}
        \caption{90\% mixture}
    \end{subfigure}
    \hfill
    \begin{subfigure}[b]{0.49\textwidth}
        \centering
        \includegraphics[width=\textwidth]{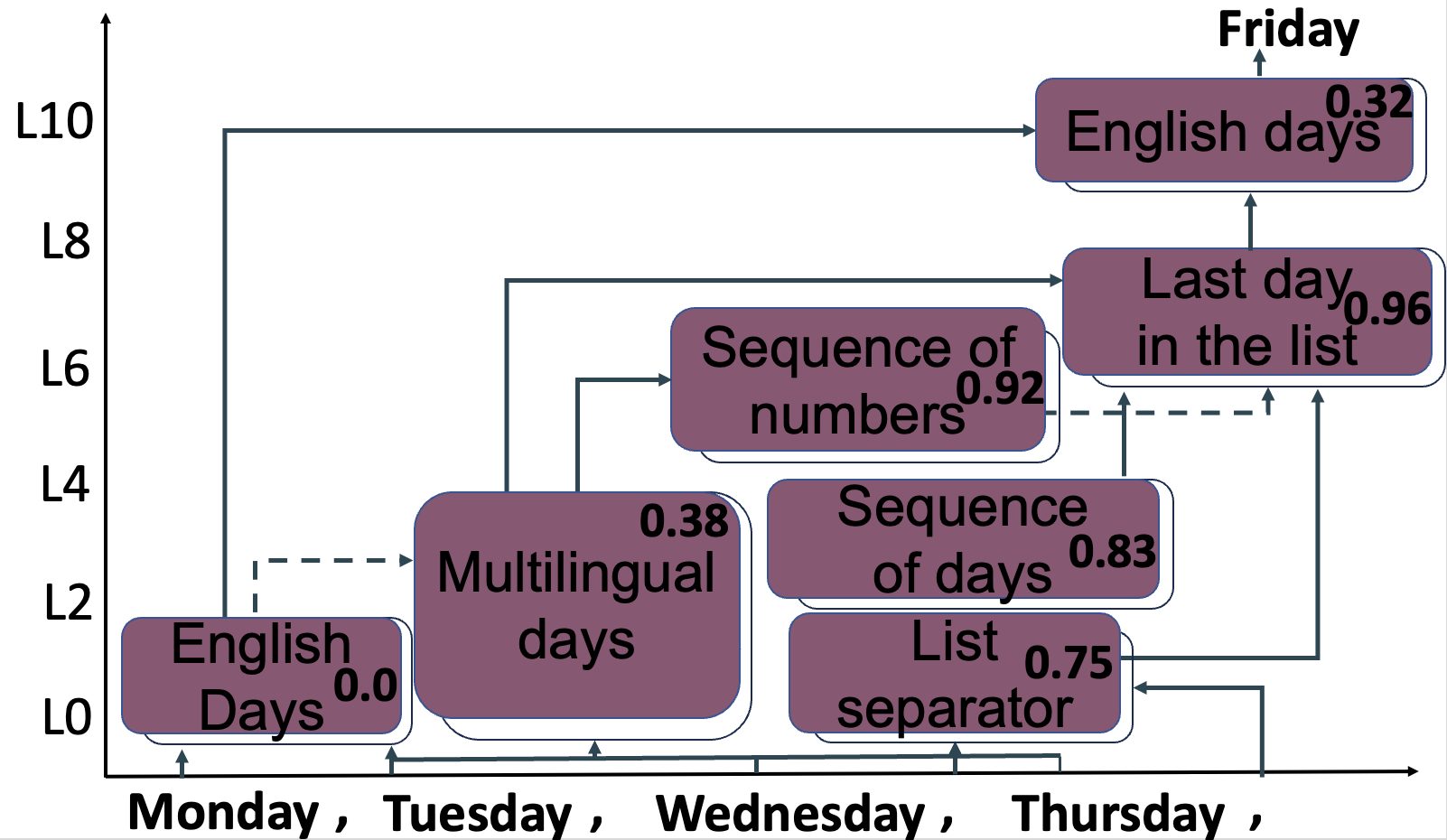}
        \caption{70\% mixture}
    \end{subfigure}
    
    \vspace{0.3em}
    
    \begin{subfigure}[b]{0.49\textwidth}
        \centering
        \includegraphics[width=\textwidth,trim=0 0.5 0 0.5, clip]{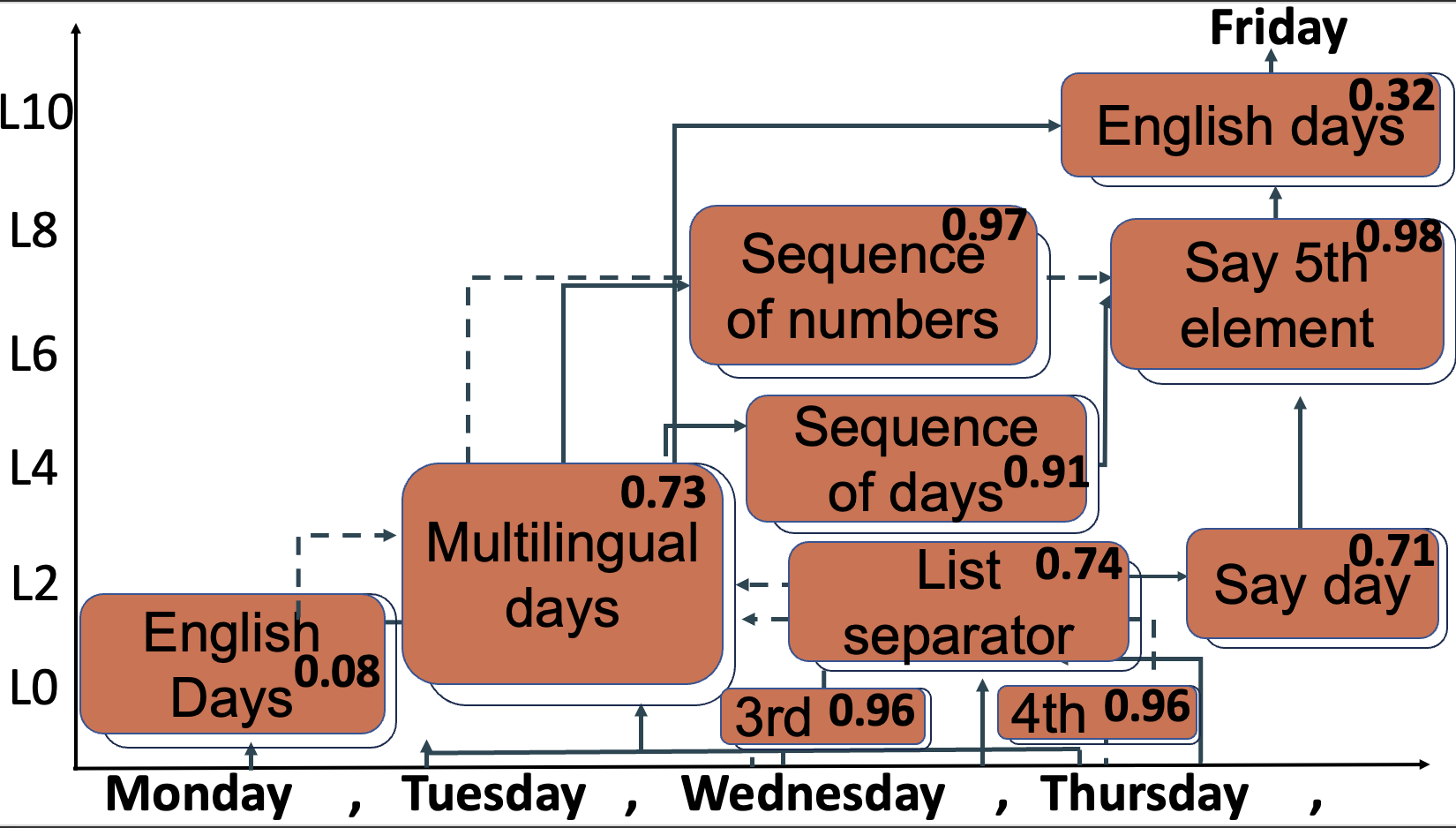}
        \caption{50\% mixture}
    \end{subfigure}
    \hfill
    \begin{subfigure}[b]{0.49\textwidth}
        \centering
        \includegraphics[width=\textwidth]{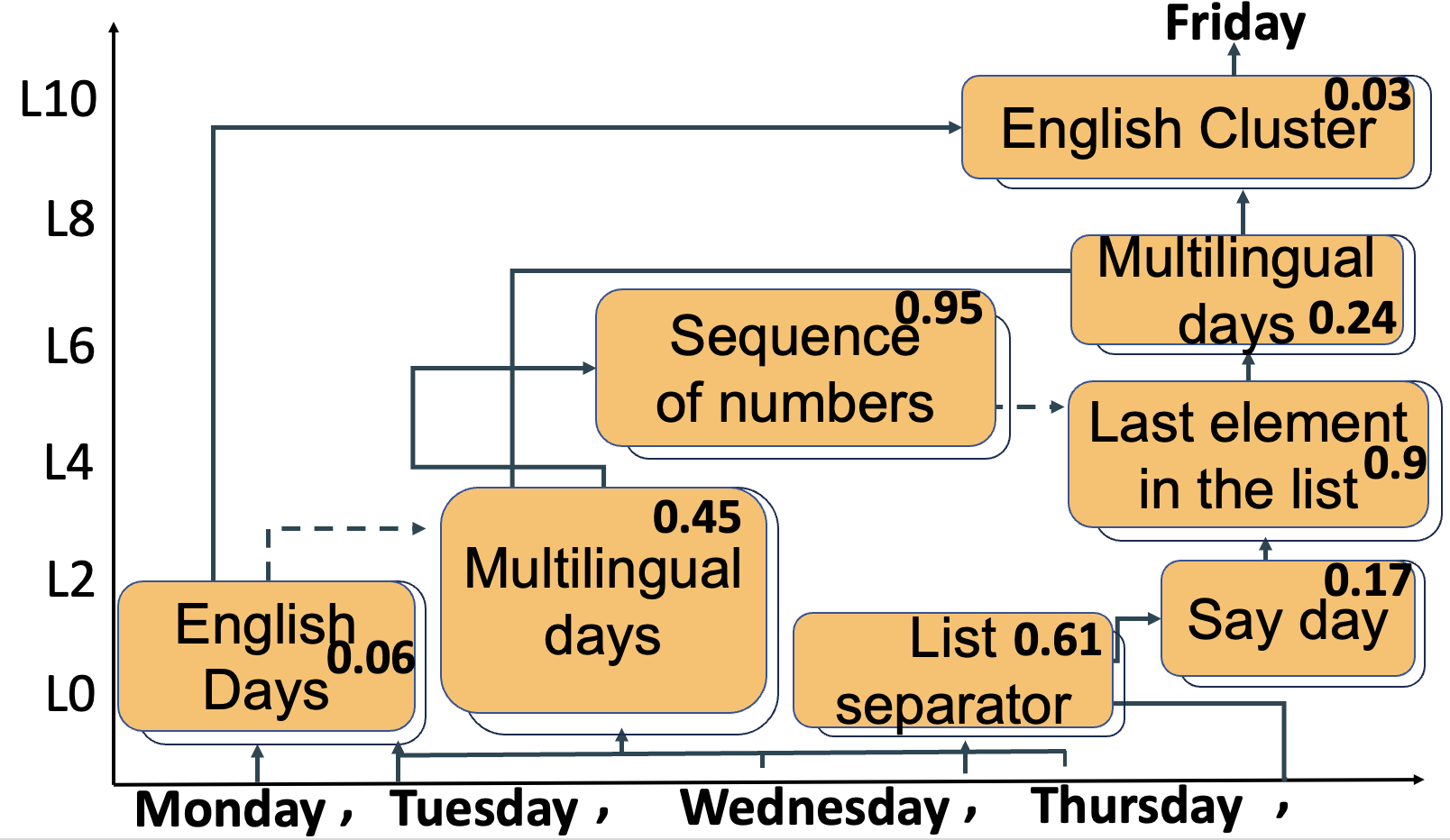}
        \caption{20\% mixture}
    \end{subfigure}
    
    \caption{Circuits for calendar term prediction (``Monday, Tuesday, Wednesday, Thursday'') across mixtures.}
    \label{fig:circuits_calendar}
\end{figure*}

\subsubsection{Analogy Sentences (\texttt{``the opposite of 'men' is ...''})}
Finally, we analyze analogy-based prediction across mixtures.  
Unlike lexical recall, analogies engage relational reasoning.  
We find that the same multilingual clusters are reused across mixtures, especially in middle layers, underscoring their role in cross-lingual generalization.  

\begin{figure*}[h]
    \centering
    \begin{subfigure}[b]{0.49\textwidth}
        \centering
        \includegraphics[width=\textwidth,trim=0 0 1 0, clip]{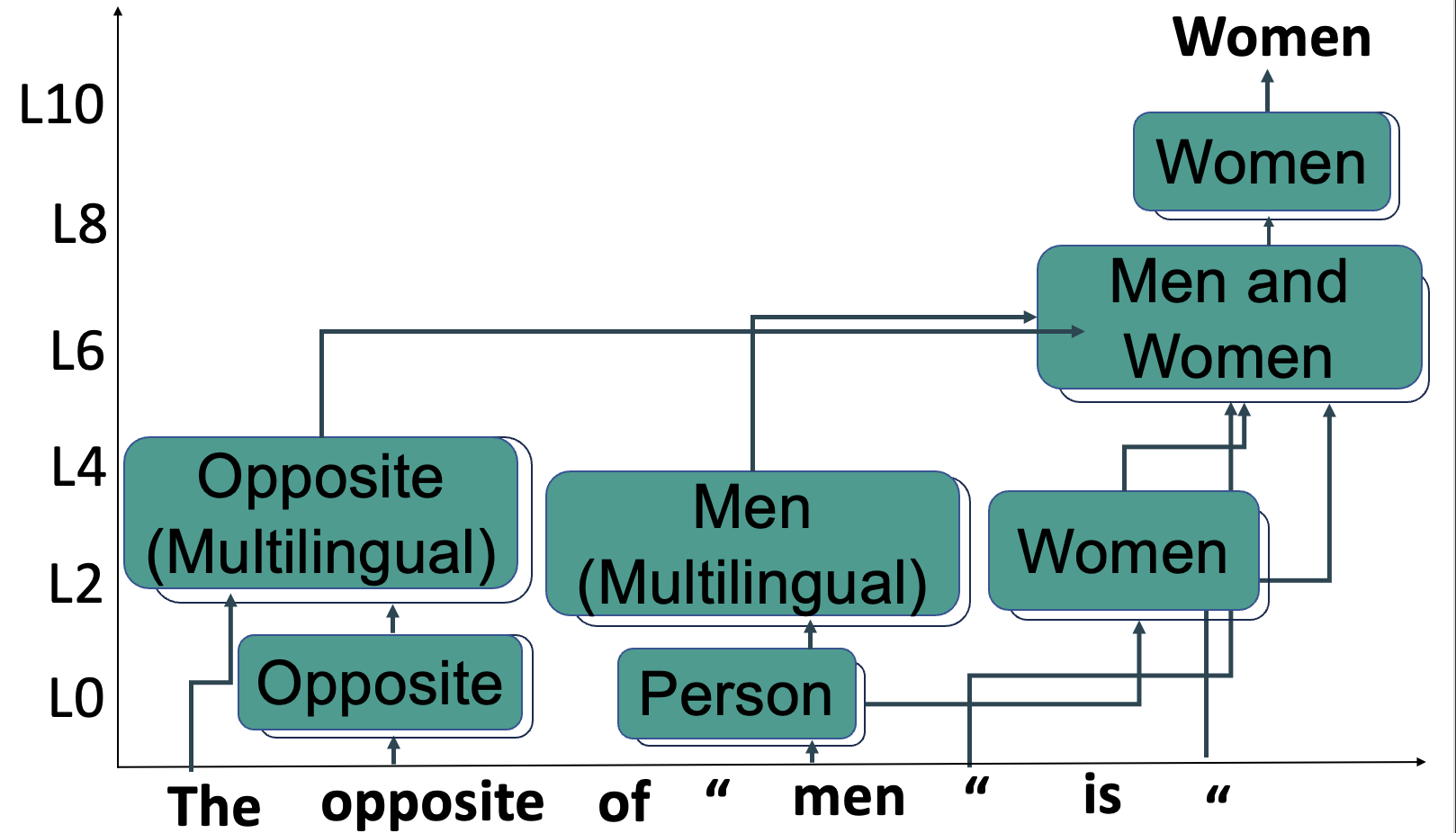}
        \caption{English}
    \end{subfigure}
    \hfill
    \begin{subfigure}[b]{0.49\textwidth}
        \centering
        \includegraphics[width=\textwidth,trim=0 0.5 0 0, clip]{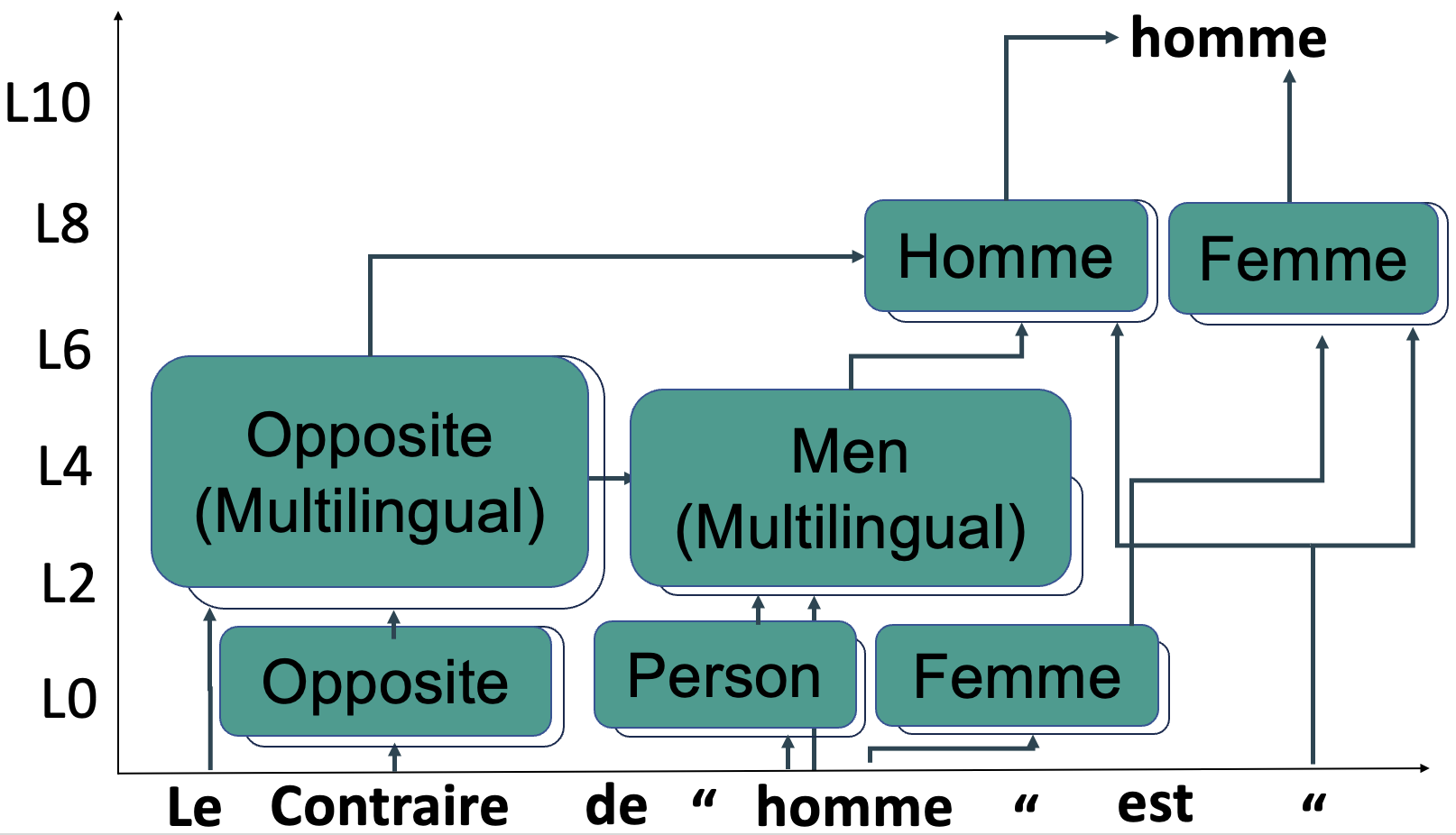}
        \caption{French}
    \end{subfigure}
    
    \vspace{0.3em}
    
    \begin{subfigure}[b]{0.49\textwidth}
        \centering
        \includegraphics[width=\textwidth,trim=0 0 0 0.5, clip]{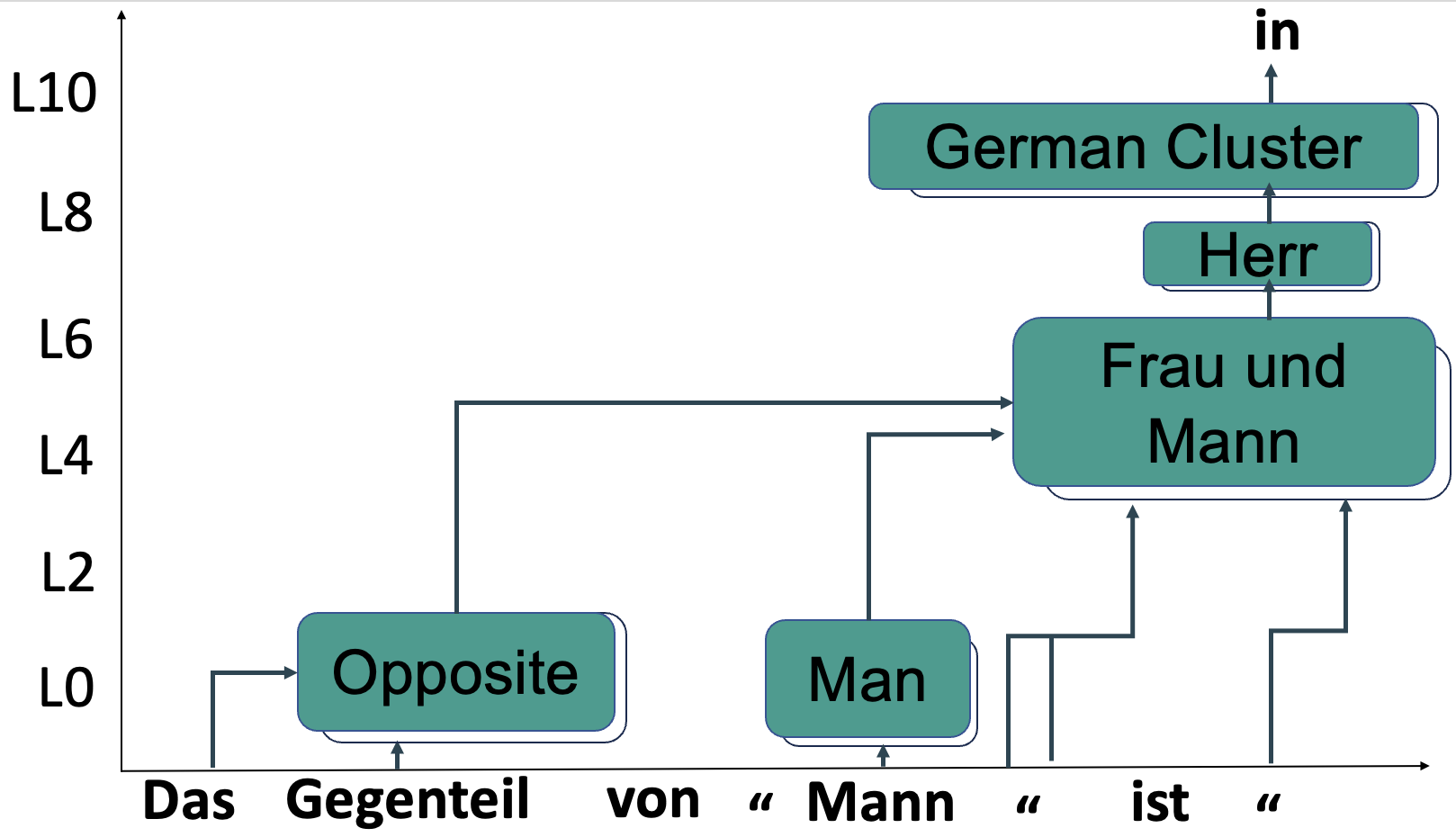}
        \caption{German}
    \end{subfigure}
    \hfill
    \begin{subfigure}[b]{0.49\textwidth}
        \centering
        \includegraphics[width=\textwidth,trim=0 1 0 1, clip]{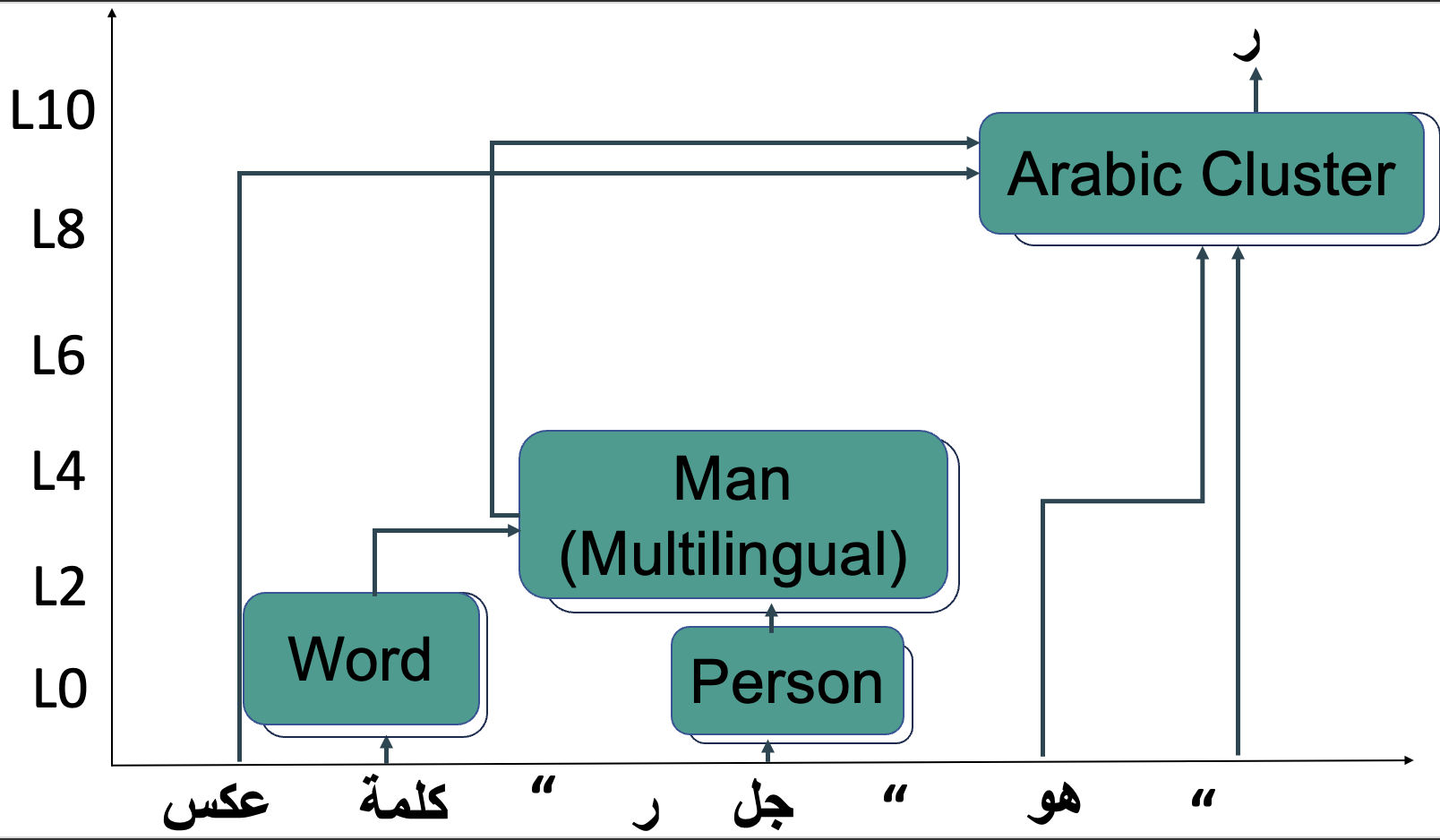}
        \caption{Arabic}
    \end{subfigure}
    
    \vspace{0.3em}
    
    \begin{subfigure}[b]{0.49\textwidth}
        \centering
        \includegraphics[width=\textwidth]{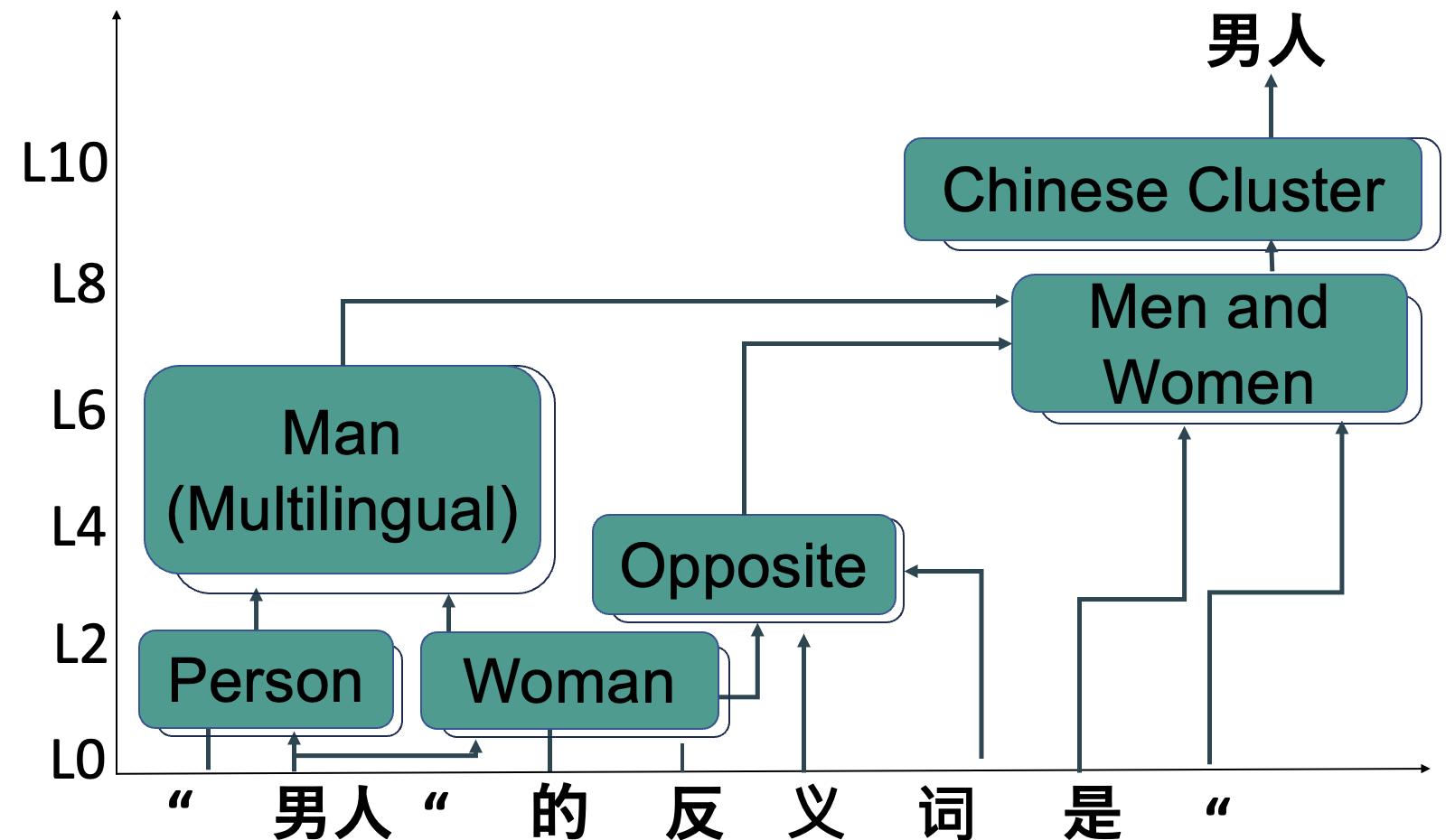}
        \caption{Chinese}
    \end{subfigure}
    
    \caption{Circuits for analogy-based prediction (``the opposite of 'men' is ...'') across mixtures.}
    \label{fig:circuits_analogy}
\end{figure*}

\subsection{Multilingual Circuits: Failure Modes}
\label{appendix:circuits_eng_better}

As discussed in Section~\ref{sec:why_models_fail}, models trained on balanced data across languages still exhibit lower performance in non-English languages compared to English. To further support the findings from Section~\ref{sec:why_models_fail}, we examine two tasks that probe distinct types of reasoning:

\begin{enumerate}
\item \textbf{Antonym Task}: Prompts such as \texttt{The opposite of ``men'' is ``}'' test semantic relations. \item \textbf{Category Task}: Prompts like ``\texttt{Football, cycling, baseball are all}'' probe conceptual grouping.
\end{enumerate}

We analyze five languages (English, German, French, Arabic, Chinese) across four training mixtures (20\%, 50\%, 70\%, 90\% English dominance) to uncover the mechanisms behind non-English failures.

\begin{figure*}[h!]
\centering
\begin{subfigure}[b]{0.32\textwidth}
\centering
\includegraphics[width=\textwidth]{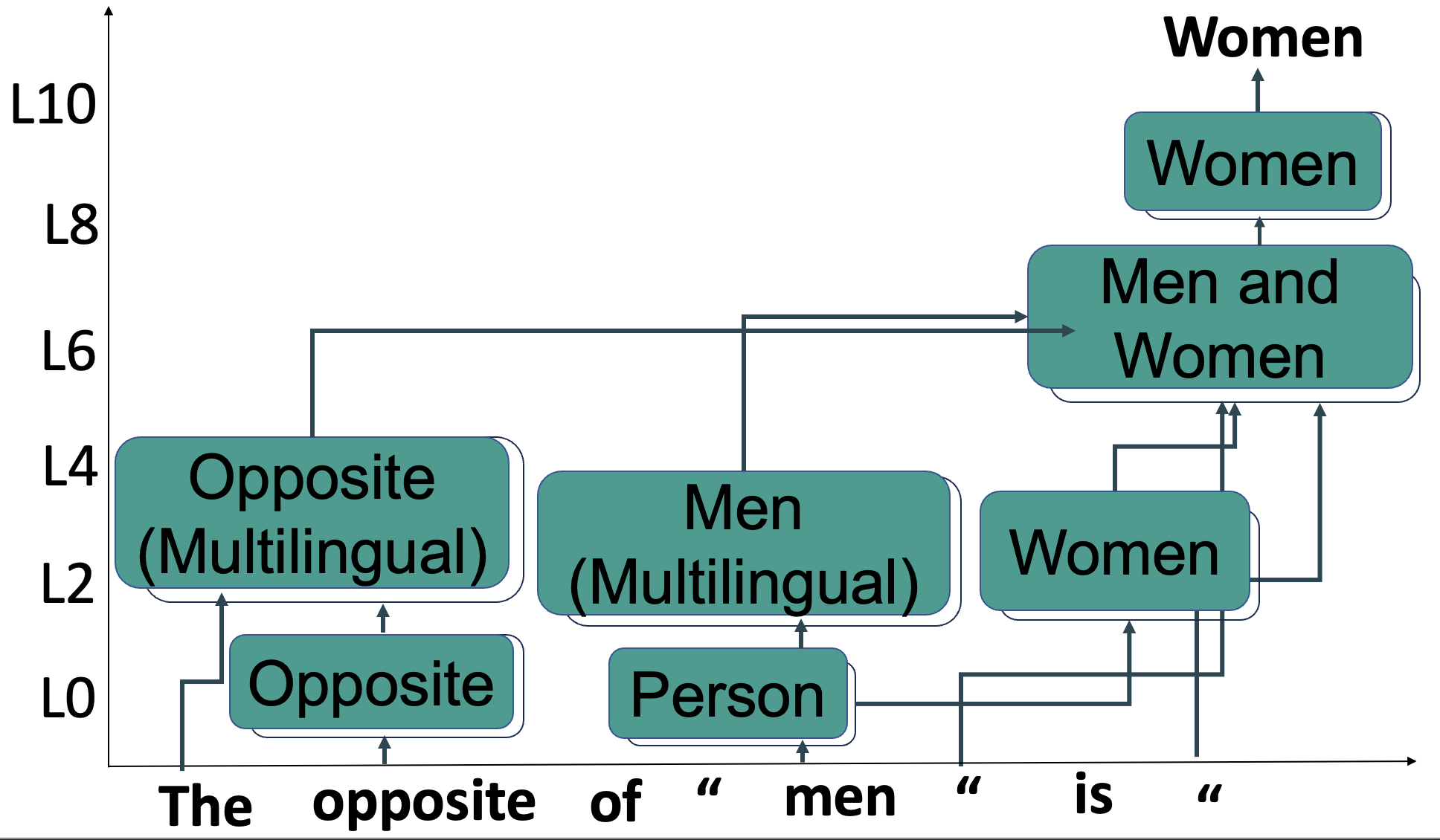}
\label{fig:subfig1}
\end{subfigure}
\hfill
\begin{subfigure}[b]{0.32\textwidth}
\centering
\includegraphics[width=\textwidth,trim=0 0 0 0.5, clip]{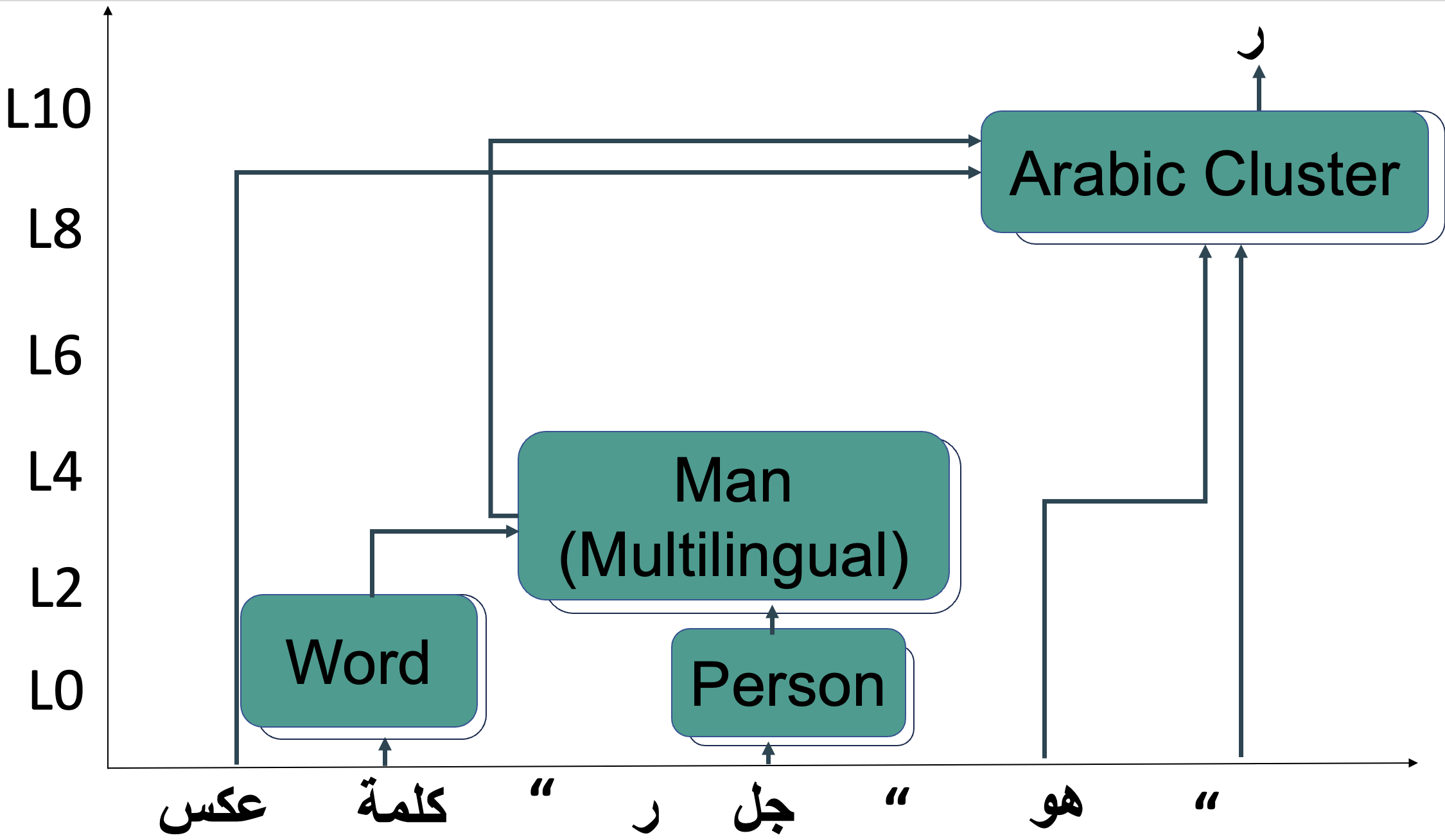}
\label{fig:subfig2}
\end{subfigure}
\hfill
\begin{subfigure}[b]{0.32\textwidth}
\centering
\includegraphics[width=\textwidth,trim=0 0 0 0.5, clip]{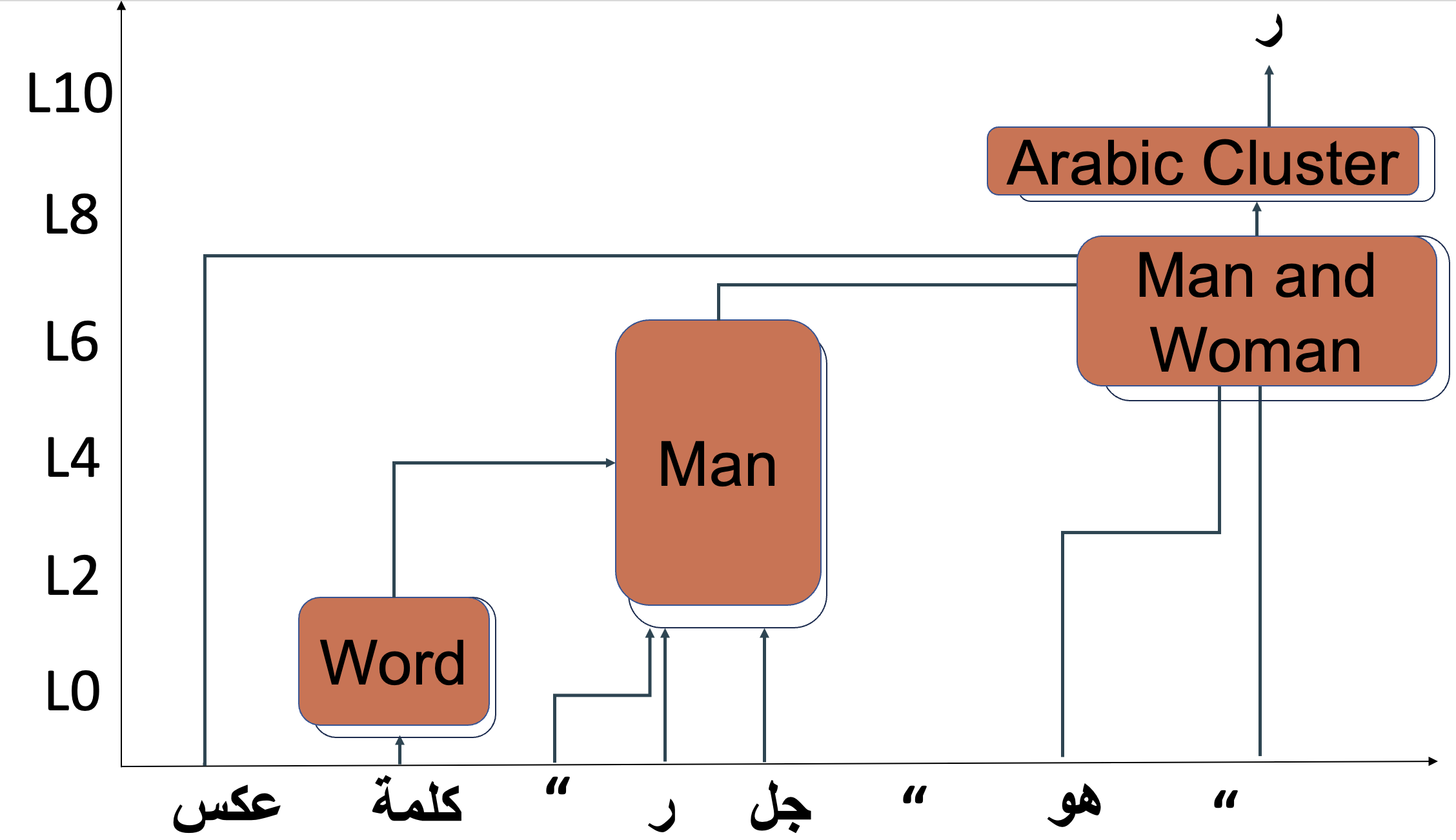}
\label{fig:subfig3}
\end{subfigure}
\caption{Circuit structures across language mixtures. Green: 90\% English mixture, orange: 50\% mixture. Key clusters missing in Arabic at 90\% emerge by 50\%, indicating mixture-dependent circuit formation. Additional examples appear in Appendix~\ref{appendix:circuits_eng_better}.}
\label{fig:eng_better_circuit}
\end{figure*}



\subsubsection{Antonym Task}

\begin{figure}[h!]
\centering
\includegraphics[width=0.5\linewidth, trim=0 1 0 2, clip]{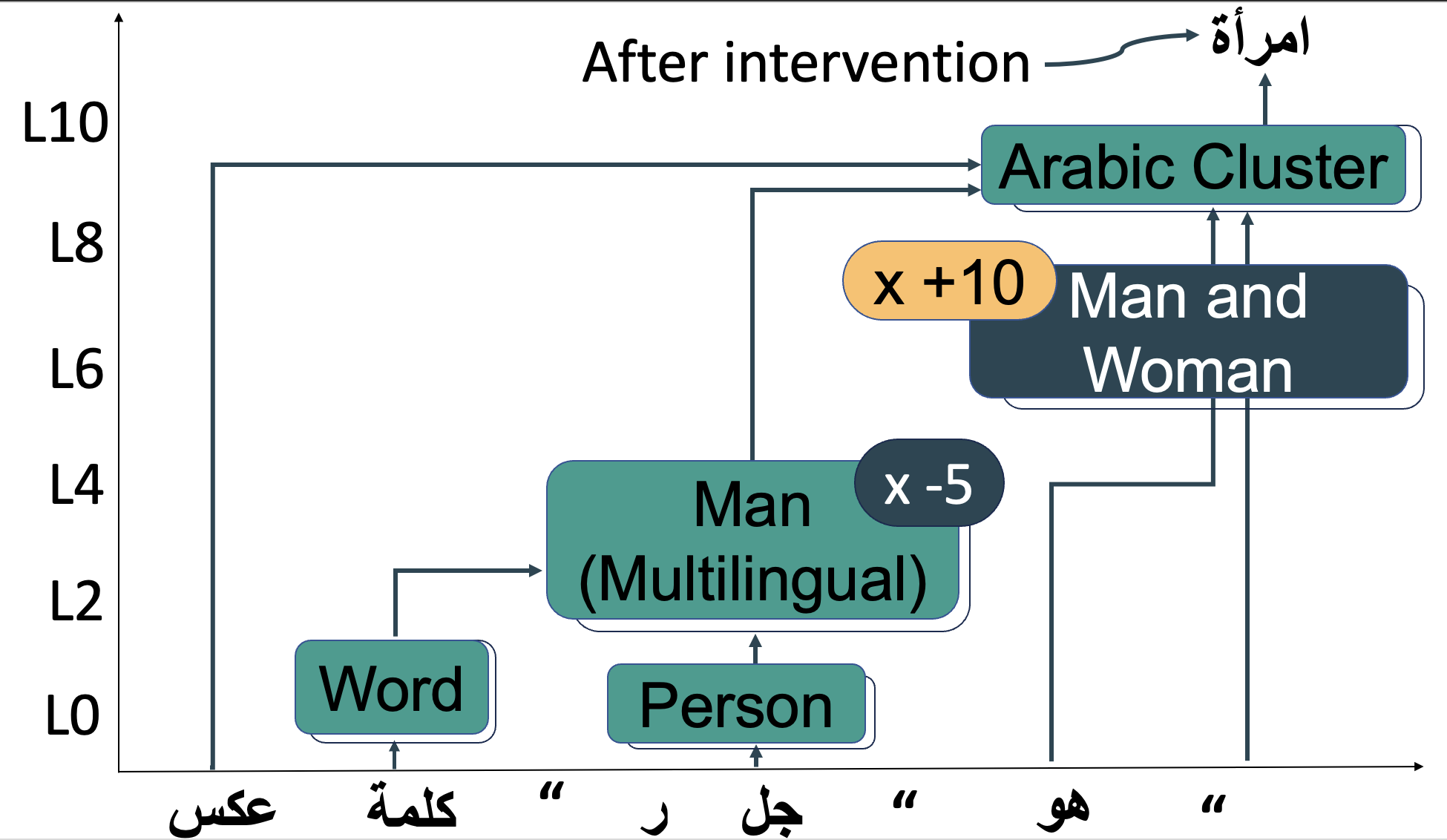}
\caption{Adding the missing \texttt{Men\&Women} cluster to the Arabic circuit restores performance.}
\label{fig:antonym_task}
\end{figure}

Circuits driving antonym prediction are consistent across languages and mixtures, but their completeness depends on mixture balance.
In Arabic under the 90\% English mixture, the \texttt{Men\&Women} cluster, present in most other languages, is missing but appears from the 50\% mixture onward (Figure~\ref{fig:eng_better_circuit}).
Intervention experiments (Figure~\ref{fig:antonym_task}) confirm its causal role: adding this missing cluster restores correct Arabic predictions (Appendix~\ref{appendix:interventions}).

\paragraph{90\% English Mixture}

Figure~\ref{fig:multiling_evidence_90} shows circuit graphs for the antonym prompt ``\texttt{The opposite of ``men'' is ``}'' across all five languages.

\begin{figure*}[h!]
    \centering
    \begin{subfigure}[b]{0.48\textwidth}
        \centering
        \includegraphics[width=\textwidth,trim=0 1 1 0, clip]{iclr2026/figures/woman_eng_90.png}
        \caption{English}
        \label{fig:antonym_eng}
    \end{subfigure}
    \hfill
    \begin{subfigure}[b]{0.48\textwidth}
        \centering
        \includegraphics[width=\textwidth,trim=0 0.5 0 0, clip]{iclr2026/figures/woman_fr_90.png}
        \caption{French}
        \label{fig:antonym_fr}
    \end{subfigure}
    
    \begin{subfigure}[b]{0.48\textwidth}
        \centering
        \includegraphics[width=\textwidth,trim=0 0 0 0.5, clip]{iclr2026/figures/woman_de_90.png}
        \caption{German}
        \label{fig:antonym_de}
    \end{subfigure}
    \hfill
    \begin{subfigure}[b]{0.48\textwidth}
        \centering
        \includegraphics[width=\textwidth,trim=0 1 0 1, clip]{iclr2026/figures/woman_ar_90.png}
        \caption{Arabic}
        \label{fig:antonym_ar}
    \end{subfigure}
    
    \begin{subfigure}[b]{0.48\textwidth}
        \centering
        \includegraphics[width=\textwidth]{iclr2026/figures/woman_zh_90.png}
        \caption{Chinese}
        \label{fig:antonym_zh}
    \end{subfigure}
    
    \caption{Circuits identified across five languages at 90\% mixture.}
    \label{fig:multiling_evidence_90}
\end{figure*}

The circuits reveal why English achieves superior performance: languages that fail to predict the correct antonym consistently lack the \texttt{Men\&Woman} cluster. Arabic shows particularly sparse connectivity, missing most multilingual clusters found in other languages due to limited circuit sharing with English.

\paragraph{70\% English Mixture}

\begin{figure*}[h!]
\centering
\resizebox{0.8\textwidth}{!}{%
\begin{minipage}{\textwidth}

    \begin{subfigure}[b]{0.49\textwidth}
        \centering
        \includegraphics[width=\textwidth,trim=0 1 0 0, clip]{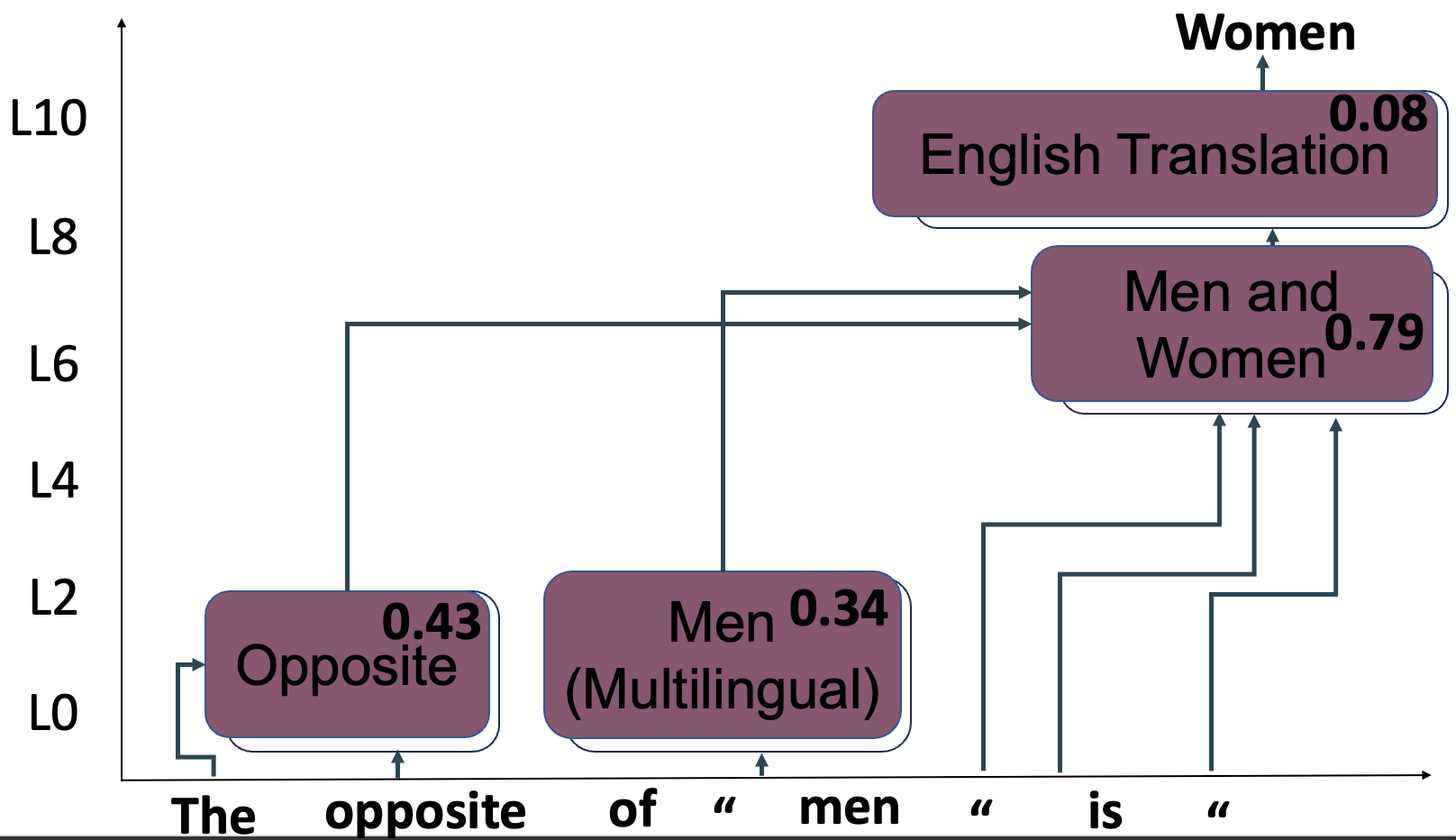}
        \caption{English}
    \end{subfigure}
    \hfill
    \begin{subfigure}[b]{0.49\textwidth}
        \centering
        \includegraphics[width=\textwidth,trim=0 0.5 0 0, clip]{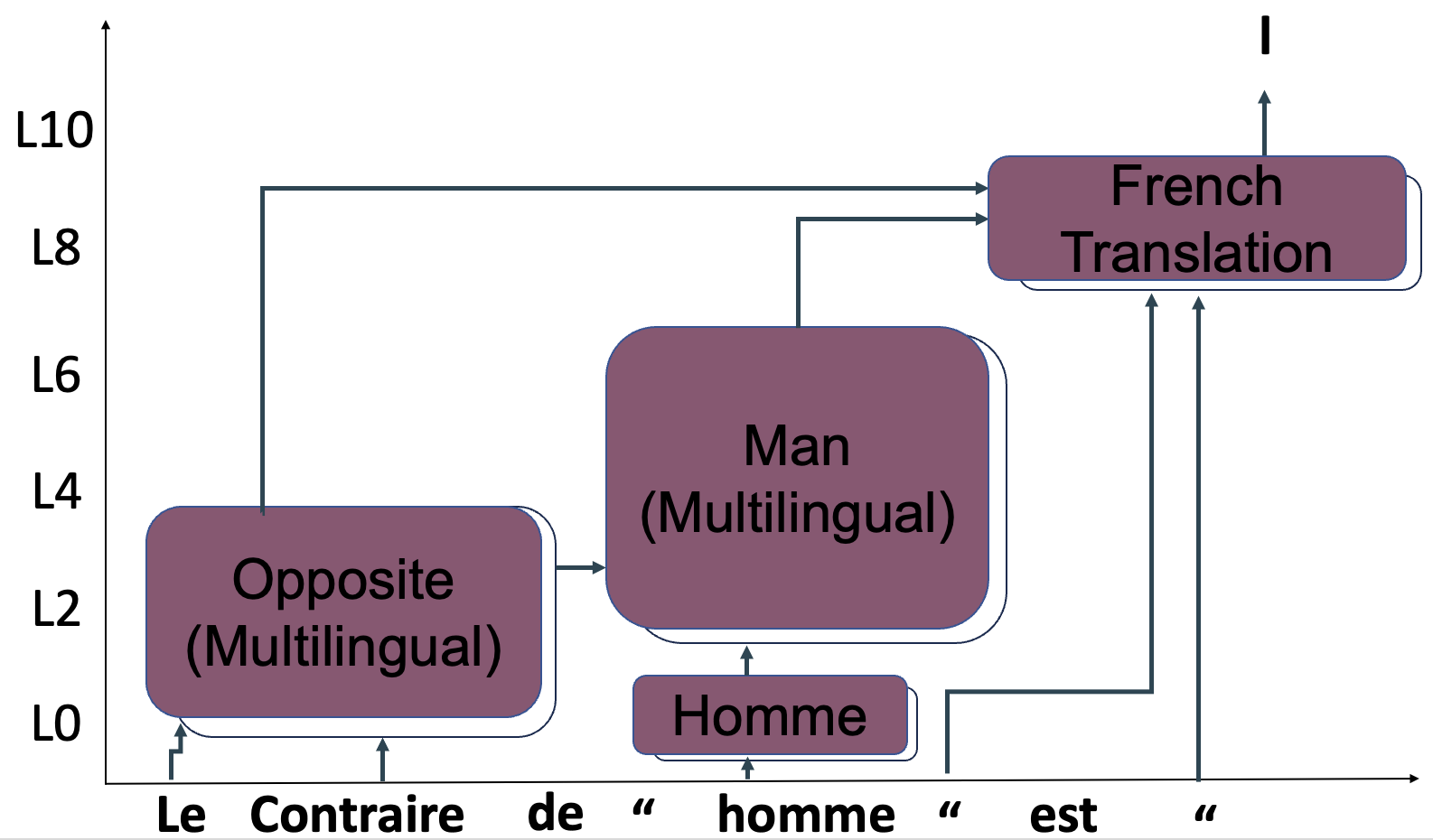}
        \caption{French}
    \end{subfigure}

    \vskip\baselineskip

    \begin{subfigure}[b]{0.49\textwidth}
        \centering
        \includegraphics[width=\textwidth]{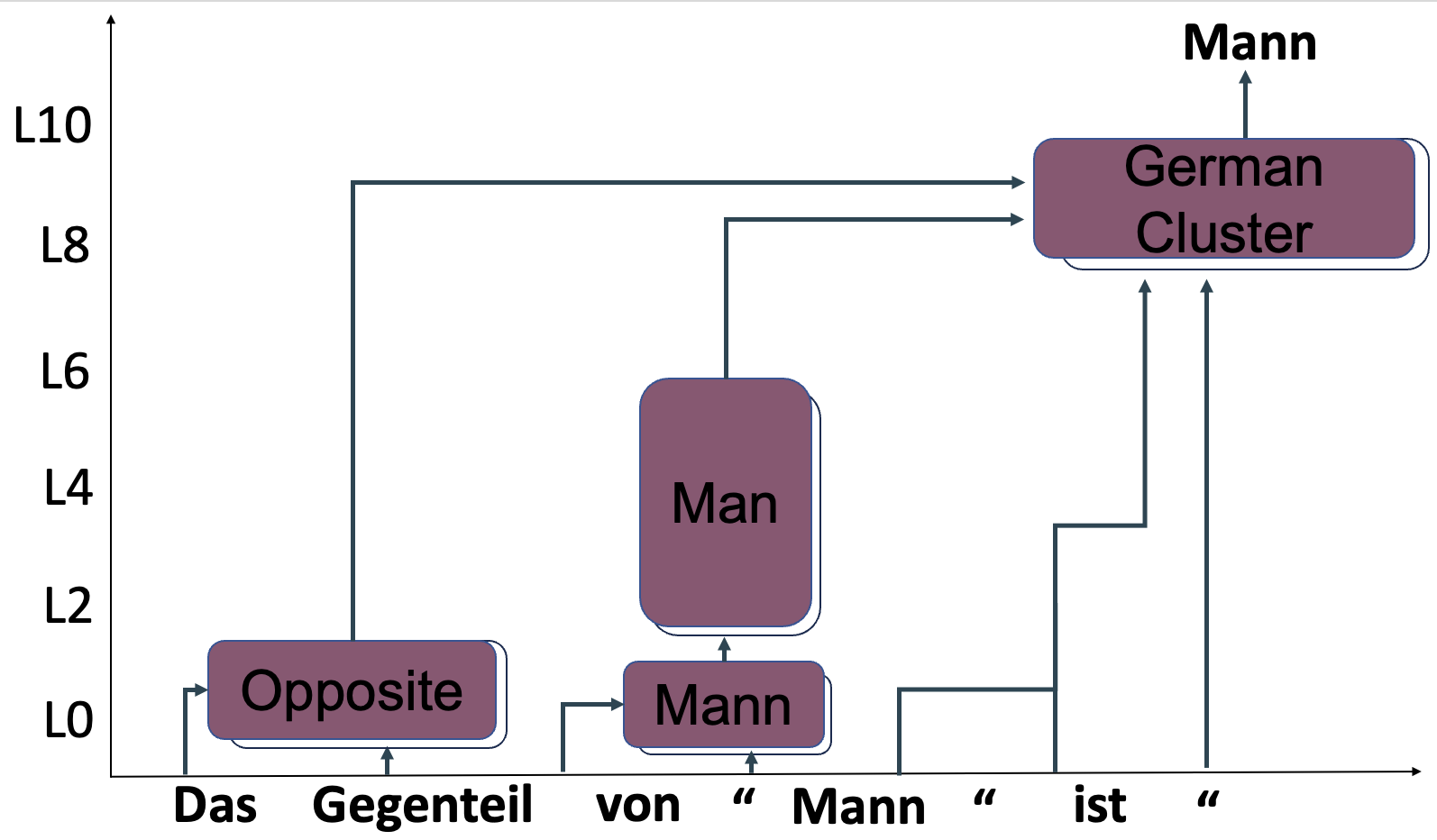}
        \caption{German}
    \end{subfigure}
    \hfill
    \begin{subfigure}[b]{0.49\textwidth}
        \centering
        \includegraphics[width=\textwidth]{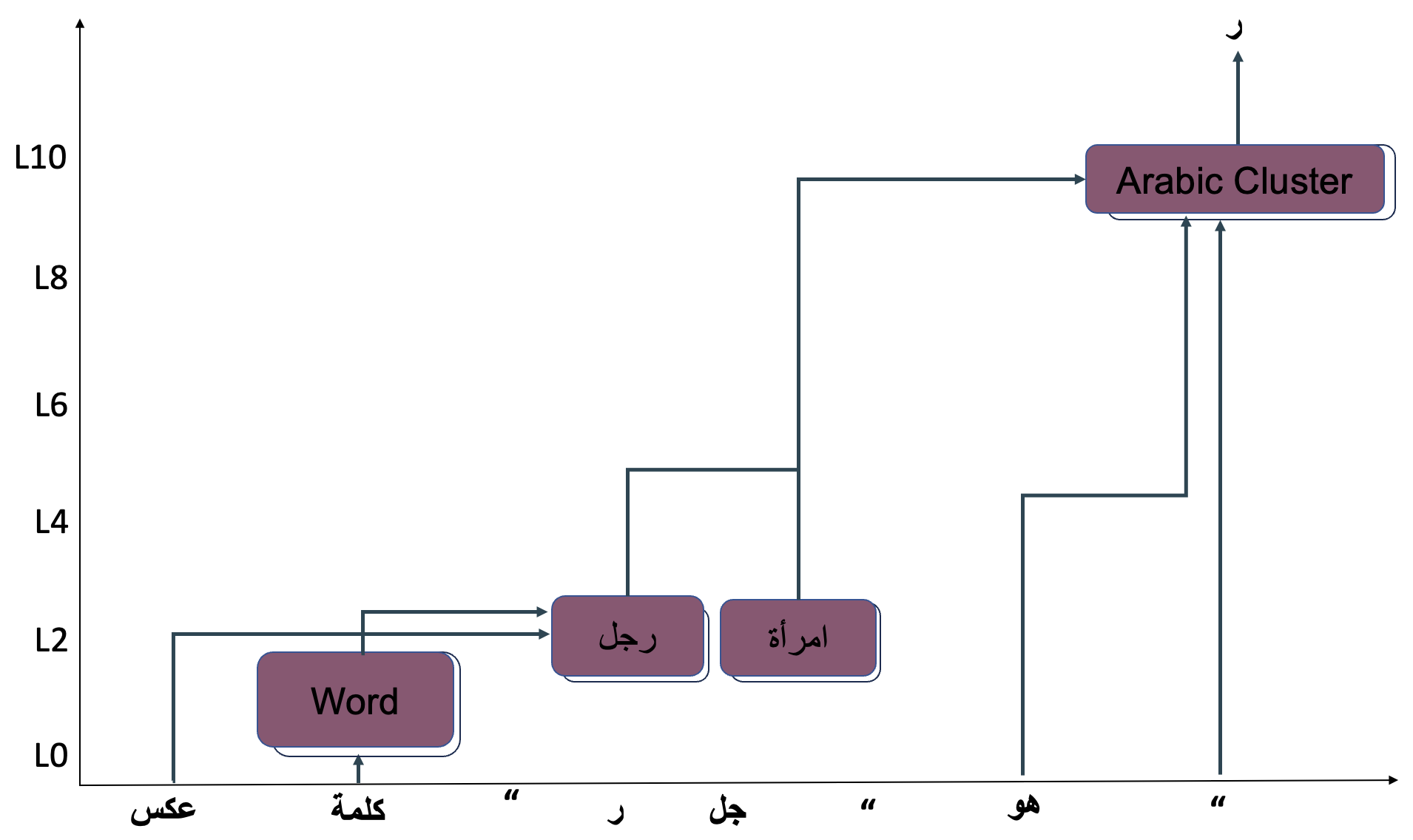}
        \caption{Arabic}
    \end{subfigure}

\end{minipage}
}
\caption{Circuits identified across five languages at 70\% mixture.}
\label{fig:multiling_evidence_70}
\end{figure*}

Figure~\ref{fig:multiling_evidence_70} shows identical patterns: the \texttt{Men\&Women} cluster remains absent where models fail. Only English sufficiently activates the reasoning circuit at this mixture level.

\paragraph{50\% English Mixture}

\begin{figure*}[h!]
    \centering
    \resizebox{0.8\textwidth}{!}{%
\begin{minipage}{\textwidth}
    \begin{subfigure}[b]{0.49\textwidth}
        \centering
        \includegraphics[width=\textwidth,trim=0 0.5 0 0, clip]{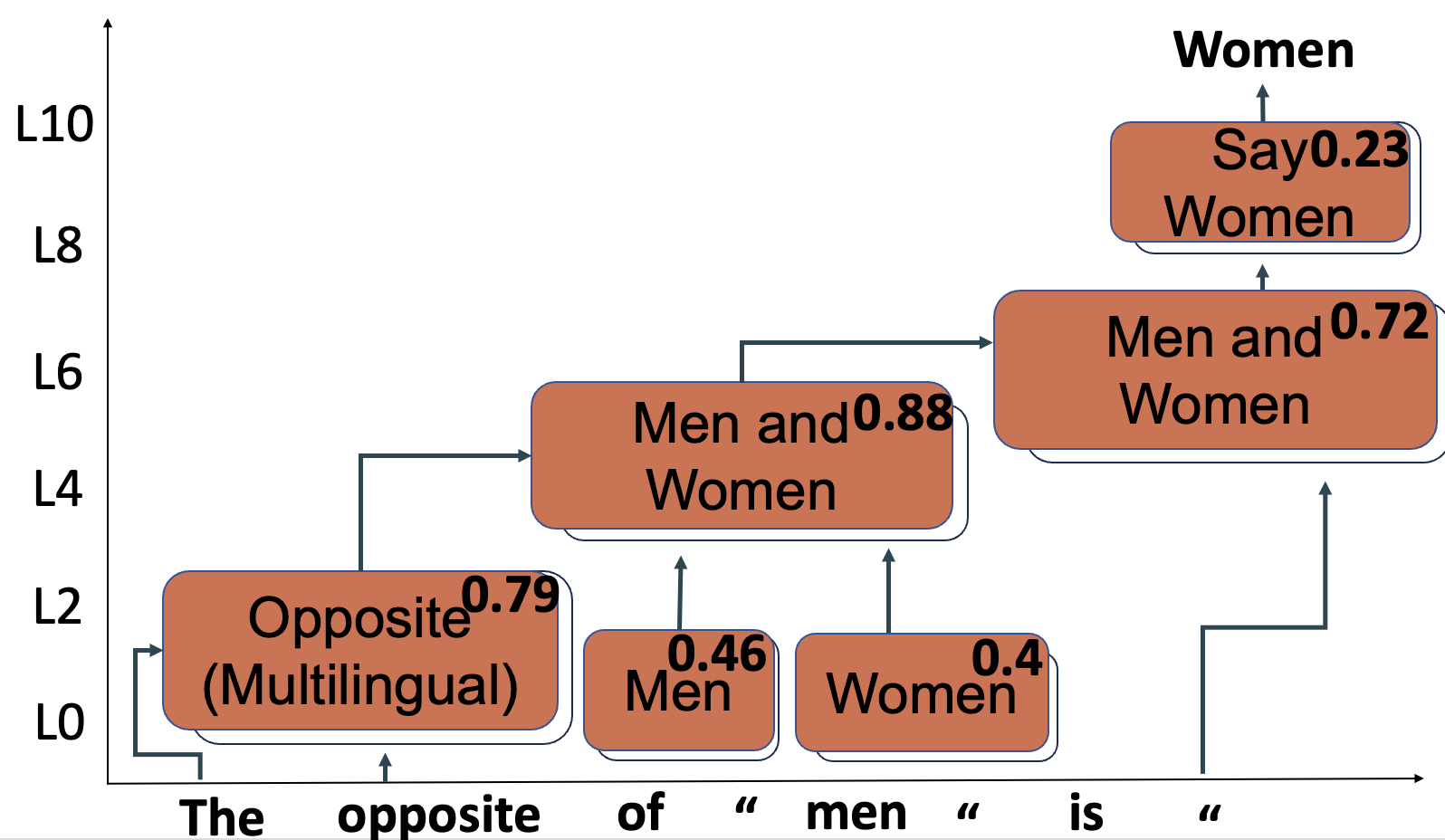}
        \caption{English}
        \label{fig:multi_en_50}
    \end{subfigure}
    \hfill
    \begin{subfigure}[b]{0.49\textwidth}
        \centering
        \includegraphics[width=\textwidth,trim=0 0.5 0 0, clip]{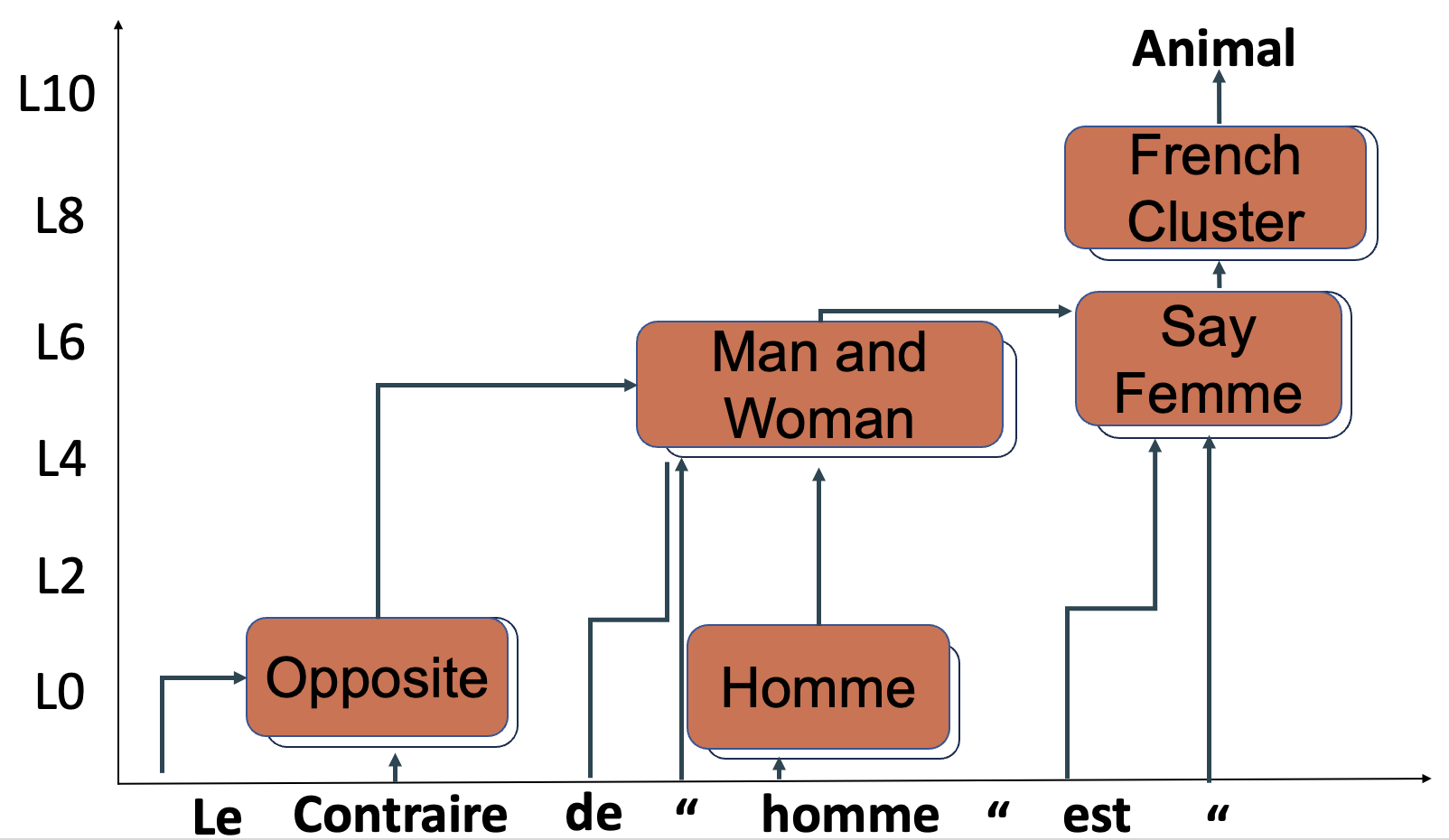}
        \caption{French}
        \label{fig:multi_fr_50}
    \end{subfigure}

    \vskip\baselineskip
    \begin{subfigure}[b]{0.49\textwidth}
        \centering
        \includegraphics[width=\textwidth,trim=0 0 0 1, clip]{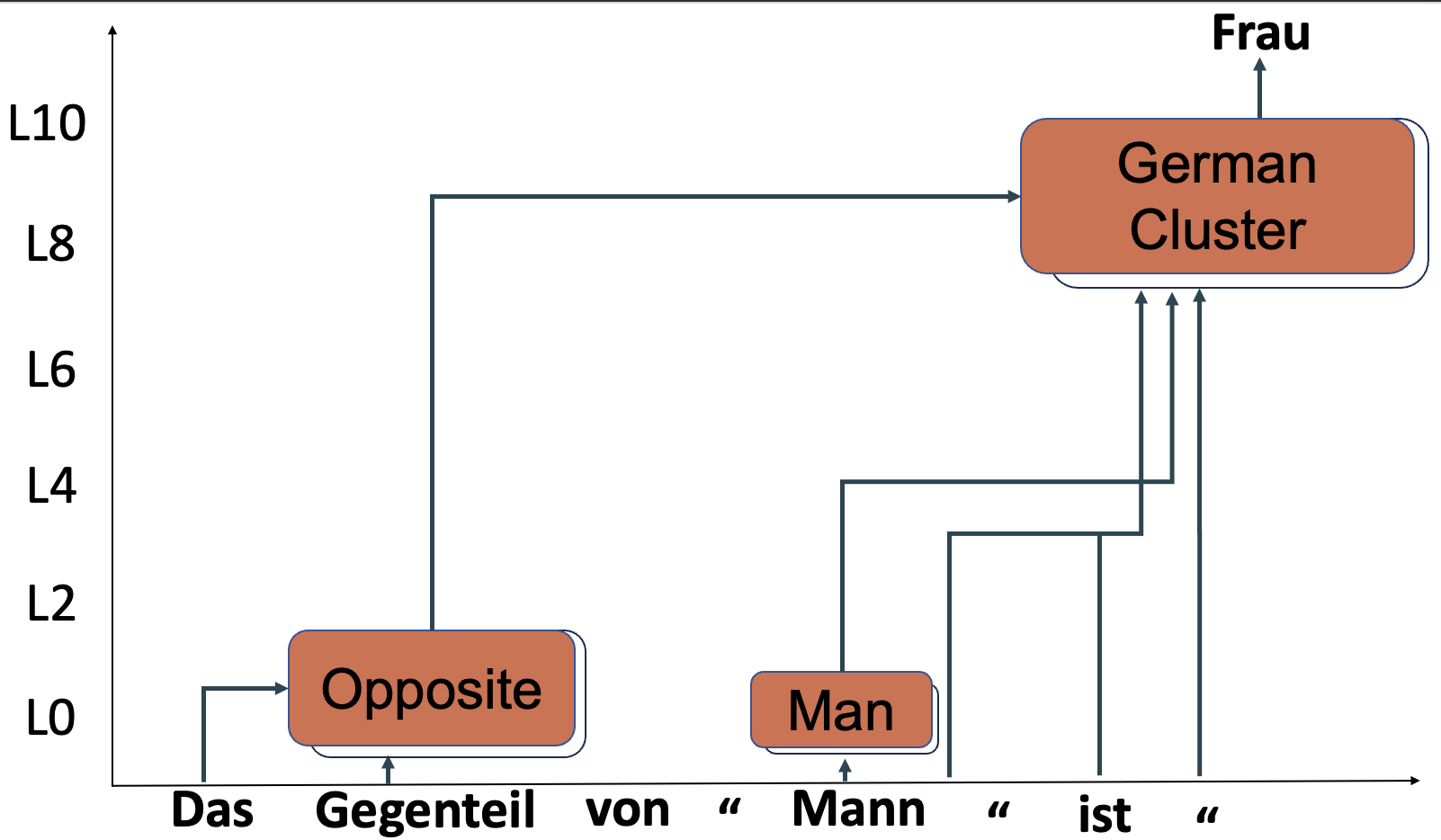}
        \caption{German}
        \label{fig:multi_de_50}
    \end{subfigure}
    \hfill
    \begin{subfigure}[b]{0.49\textwidth}
        \centering
        \includegraphics[width=\textwidth,trim=0 0.5 0 0, clip]{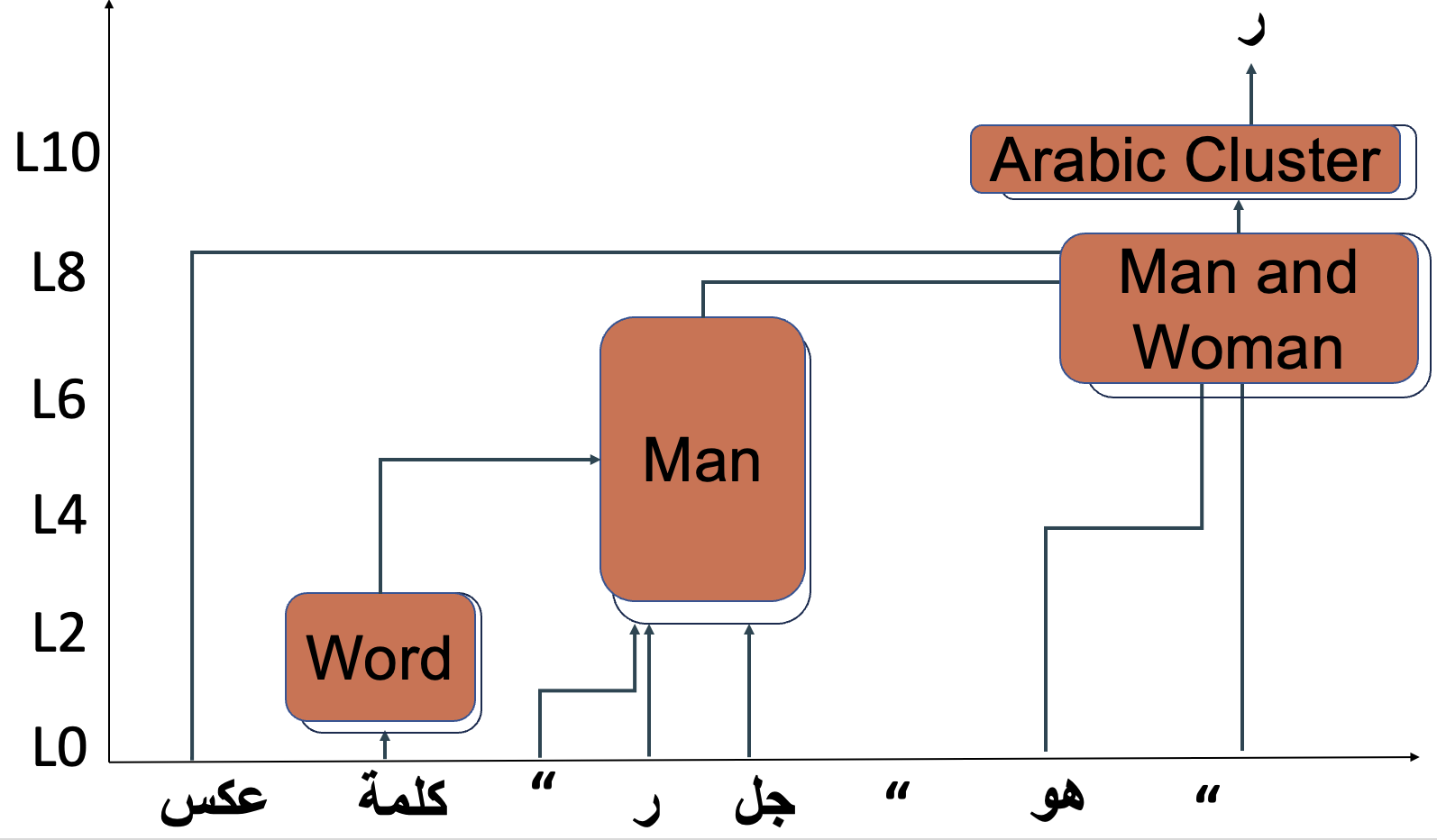}
        
        \caption{Chinese}
        \label{fig:multi_zh_50}
    \end{subfigure}
\end{minipage}
}
    \caption{Circuits identified across five languages at 50\% mixture.}
    \label{fig:multiling_evidence_50}
\end{figure*}

Figure~\ref{fig:multiling_evidence_50} validates previous patterns while showing increased presence of \texttt{Men\&Women} clusters in non-English languages, reflecting the shift toward more uniform language distribution.

\paragraph{20\% English Mixture}

\begin{figure*}[h!]
    \centering
        \resizebox{0.8\textwidth}{!}{%
\begin{minipage}{\textwidth}
    \begin{subfigure}[b]{0.49\textwidth}
        \centering
        \includegraphics[width=\textwidth,trim=0 0.5 0 0, clip]{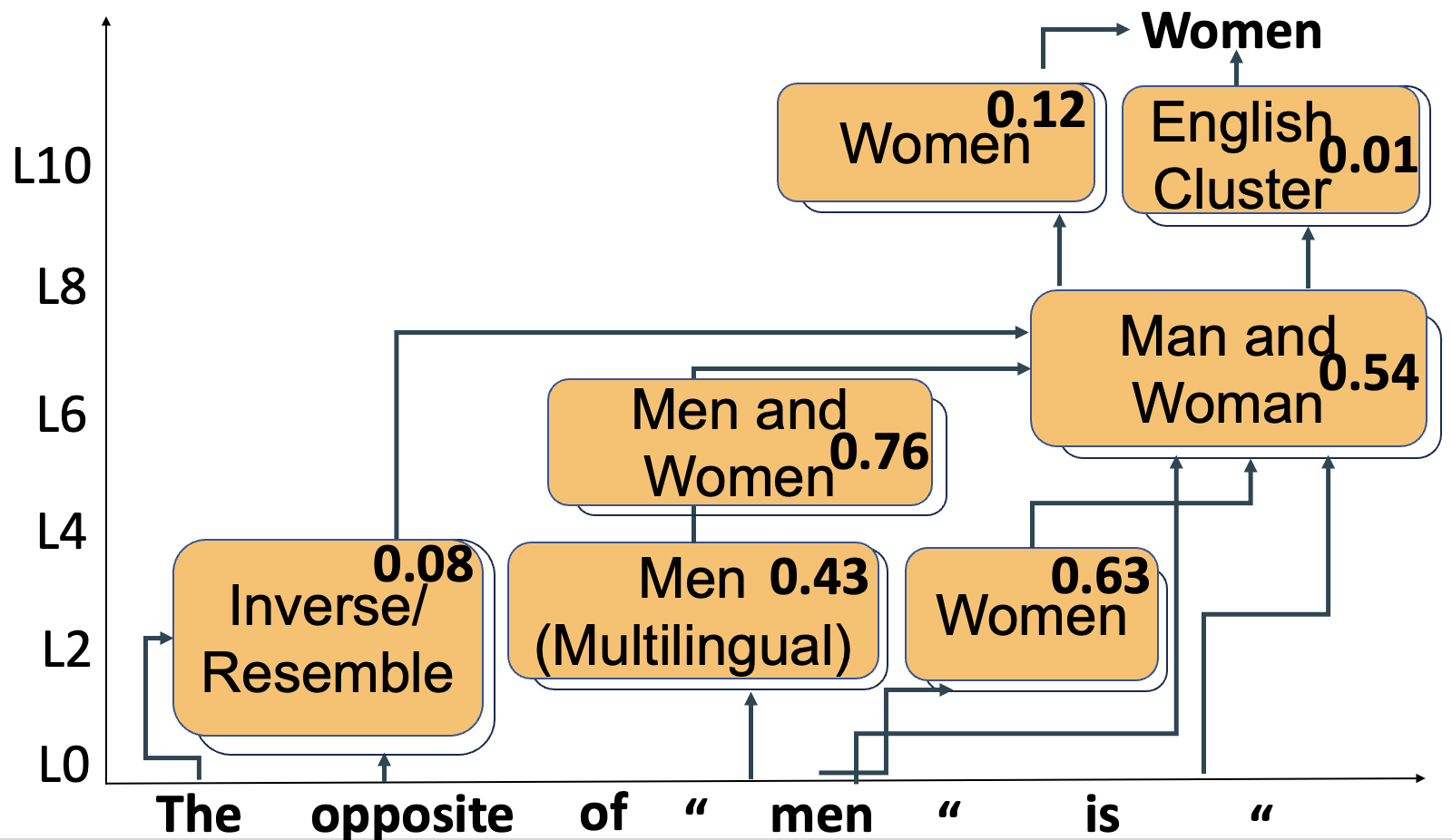}
        \caption{English}
        \label{fig:multi_eng_20}
    \end{subfigure}
    \hfill
    \begin{subfigure}[b]{0.49\textwidth}
        \centering
        \includegraphics[width=\textwidth,trim=0 0.5 0 0, clip]{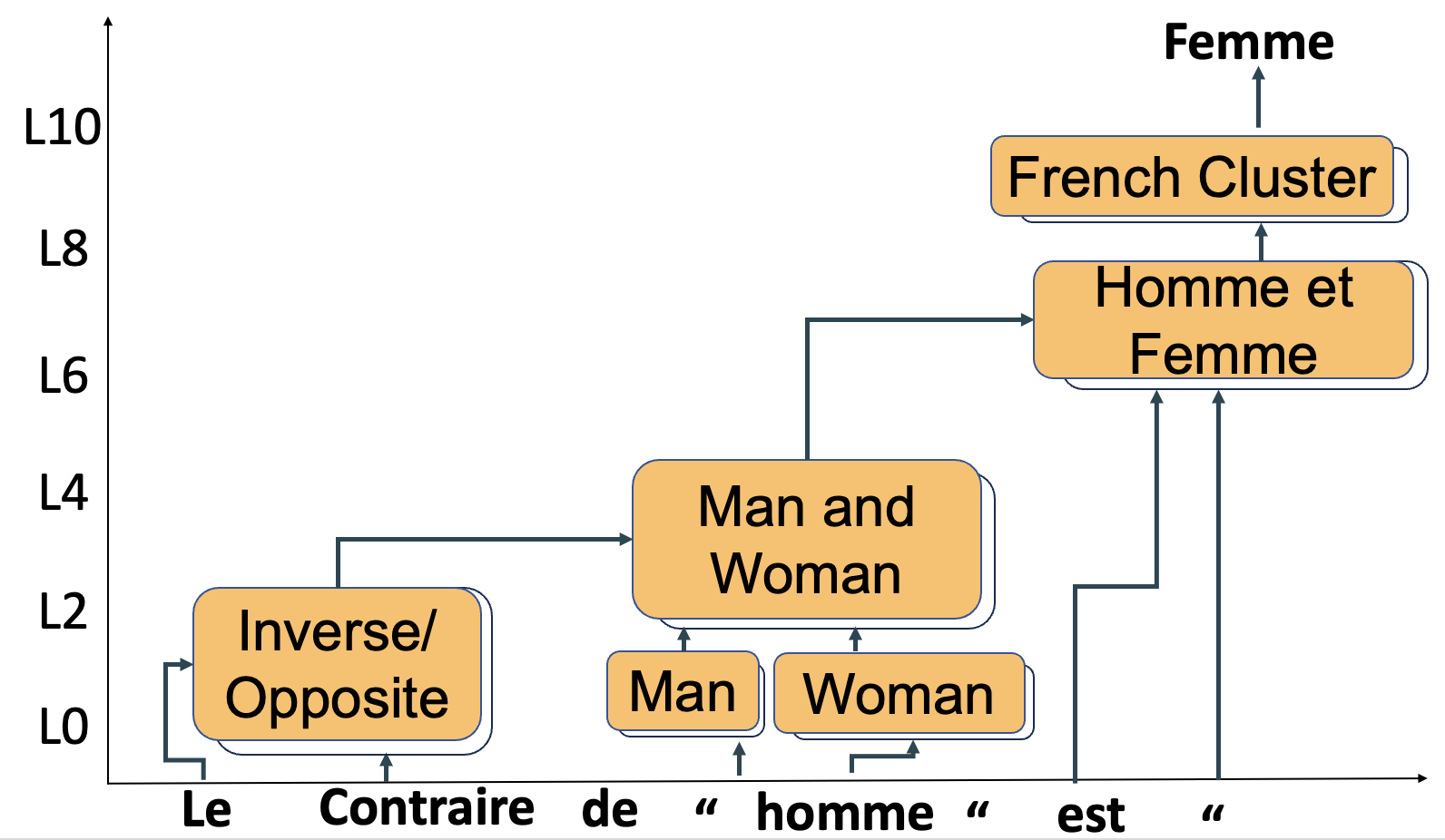}
        \caption{French}
        \label{fig:multi_fr_20}
    \end{subfigure}

    \vskip\baselineskip
    \begin{subfigure}[b]{0.49\textwidth}
        \centering
        \includegraphics[width=\textwidth,trim=0 0 0 1, clip]{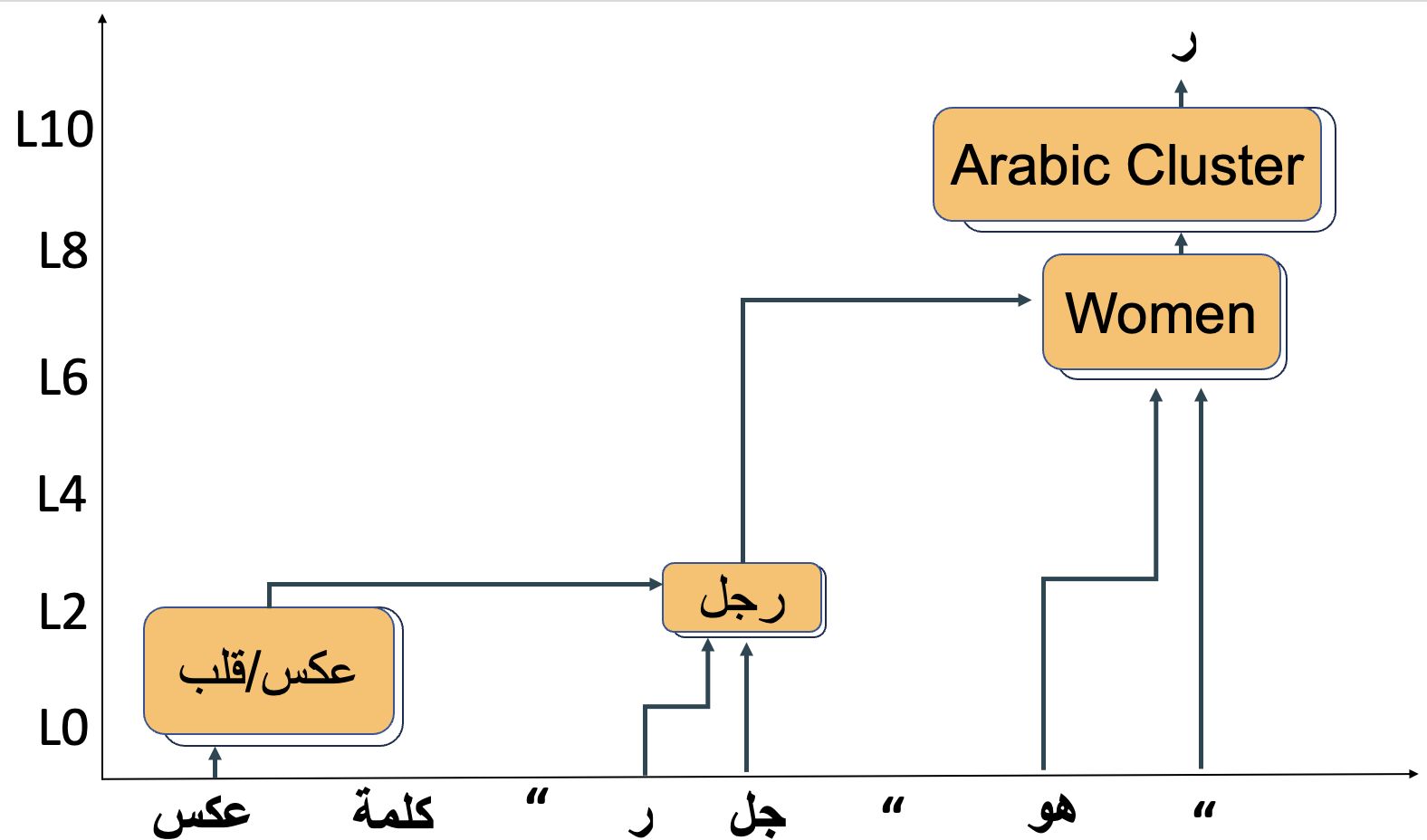}
        \caption{Arabic}
        \label{fig:multi_ar_20}
    \end{subfigure}
    \hfill
    \begin{subfigure}[b]{0.49\textwidth}
        \centering
        \includegraphics[width=\textwidth,trim=0 0.5 0 0, clip]{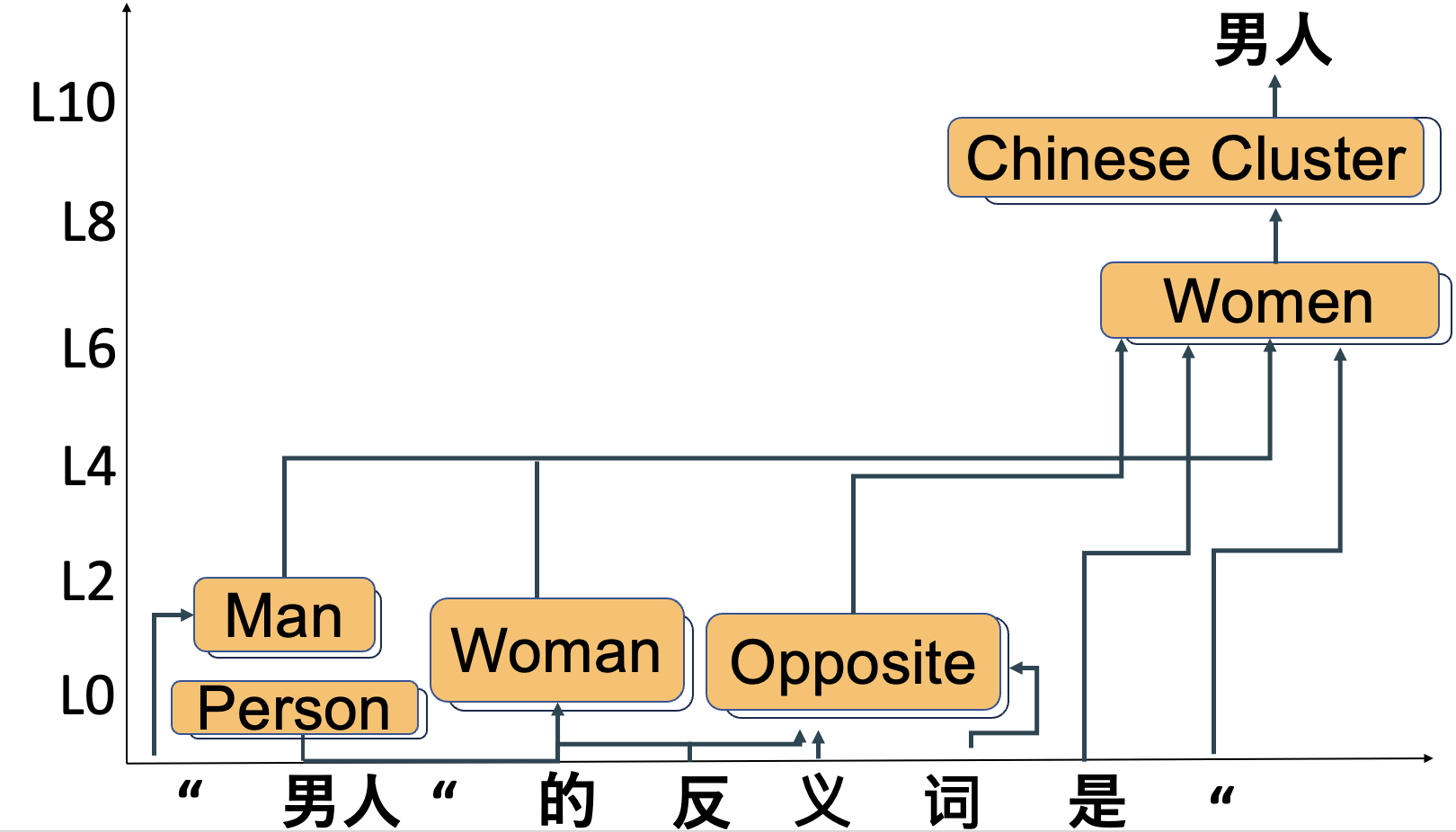}
        \caption{Chinese}
        \label{fig:multi_zh_20}
    \end{subfigure}
\end{minipage}
}
    \caption{Circuits identified across five languages at 20\% mixture.}
    \label{fig:multiling_evidence_20}
\end{figure*}

Figure~\ref{fig:multiling_evidence_20} shows more \texttt{Men\&Women} clusters and improved performance, especially for Latin-script languages. Even at this balanced mixture, non-Latin script languages still demonstrate weaker circuit formation.

The circuit patterns suggest two competing mechanisms: a reasoning circuit that correctly predicts \texttt{Women} and a copying circuit that simply reproduces \texttt{Men} from the input tokens.

To validate these hypotheses, we conduct four investigations:

\begin{enumerate}
    \item Intervention experiments: scale up reasoning circuit activation (\texttt{Men\&Women} cluster) while scaling down copying circuit activation (\texttt{Men} cluster)
    \item Comparative analysis across the three remaining mixtures to identify general trends
    \item Activation value analysis of reasoning and copying circuits to validate insufficient activation hypothesis
    \item Investigation of why English, German, and French demonstrate superior performance across all mixtures
\end{enumerate}

\subsubsection{Category Completion Task}
\label{section:category_task}

\begin{figure}
\centering
\includegraphics[width=0.6\linewidth, trim=0 0 0 0, clip]{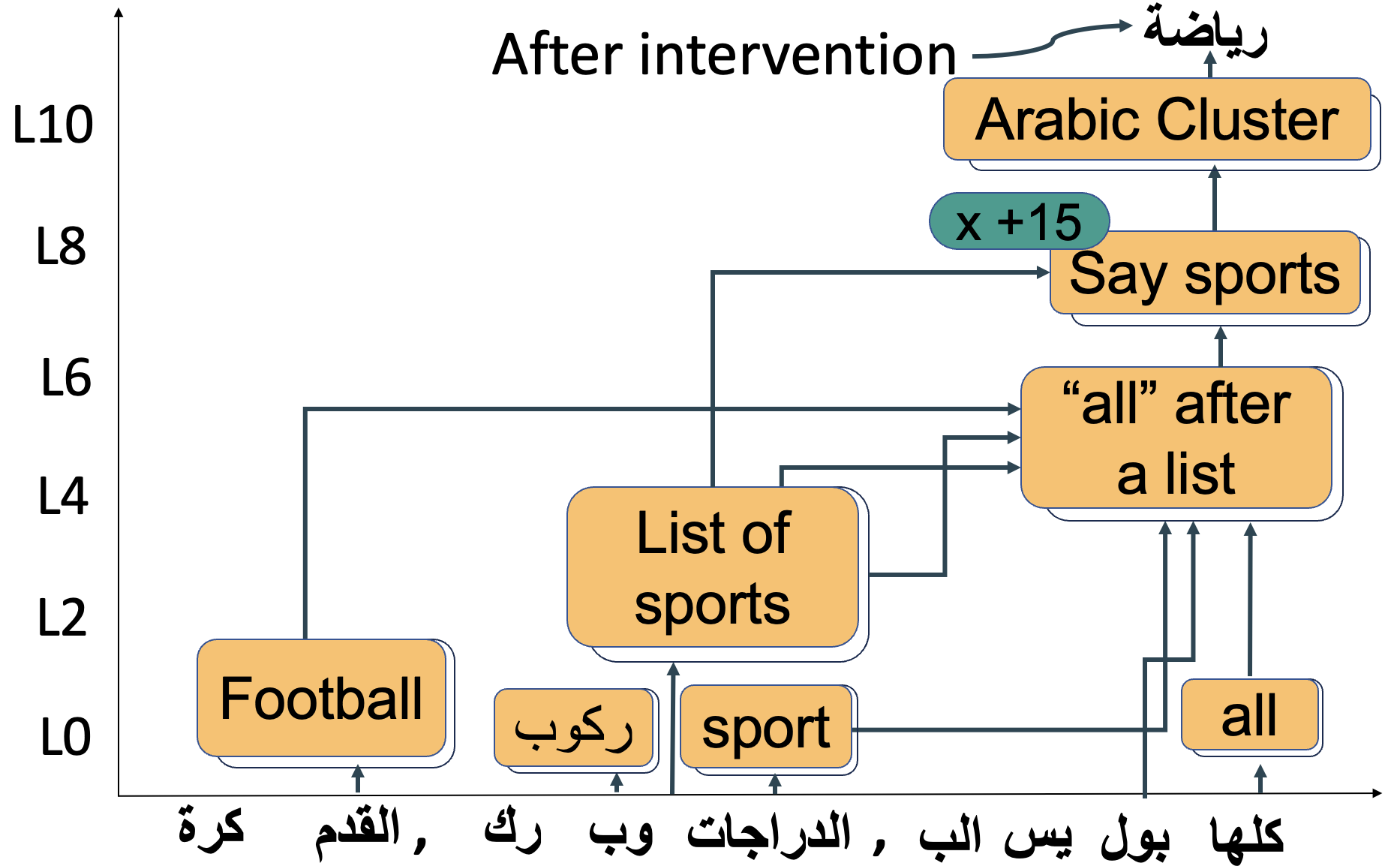}
\caption{Scaling up the \texttt{``Say Sports''} cluster makes the model predict the right answer.}
\label{figure:intervention_category}
\end{figure}

We analyze the \texttt{Category Completion} task using the 20\% mixture to understand performance in the balanced setting.  
Arabic performs poorly due to tokenization fragmentation, which forces early layers to reconstruct words rather than activate answer-related clusters.  
Intervention experiments (Figure~\ref{figure:intervention_category}) show that scaling these clusters restores correct predictions, emphasizing the importance of activation strength.

We present the circuits of the Category Completion Task presented in Figures~\ref{fig:circuits_calendar}.

\begin{figure*}
    \centering
\resizebox{0.8\textwidth}{!}{%
\begin{minipage}{\textwidth}
    \begin{subfigure}[b]{0.49\textwidth}
        \centering
        \includegraphics[width=\textwidth,trim=0 0.5 0 1, clip]{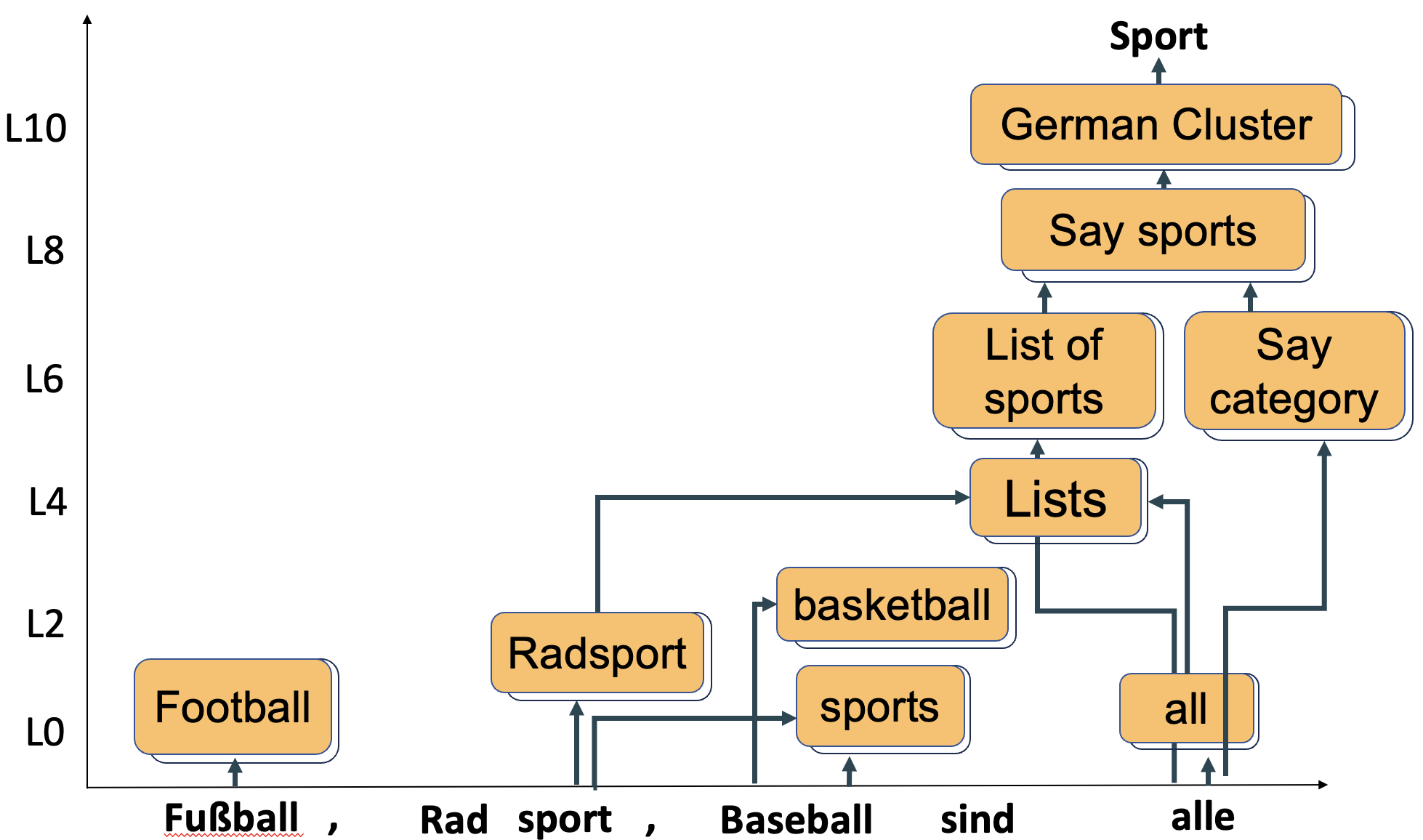}
        \caption{German}
    \end{subfigure}
    \hfill
    \begin{subfigure}[b]{0.49\textwidth}
        \centering
        \includegraphics[width=\textwidth]{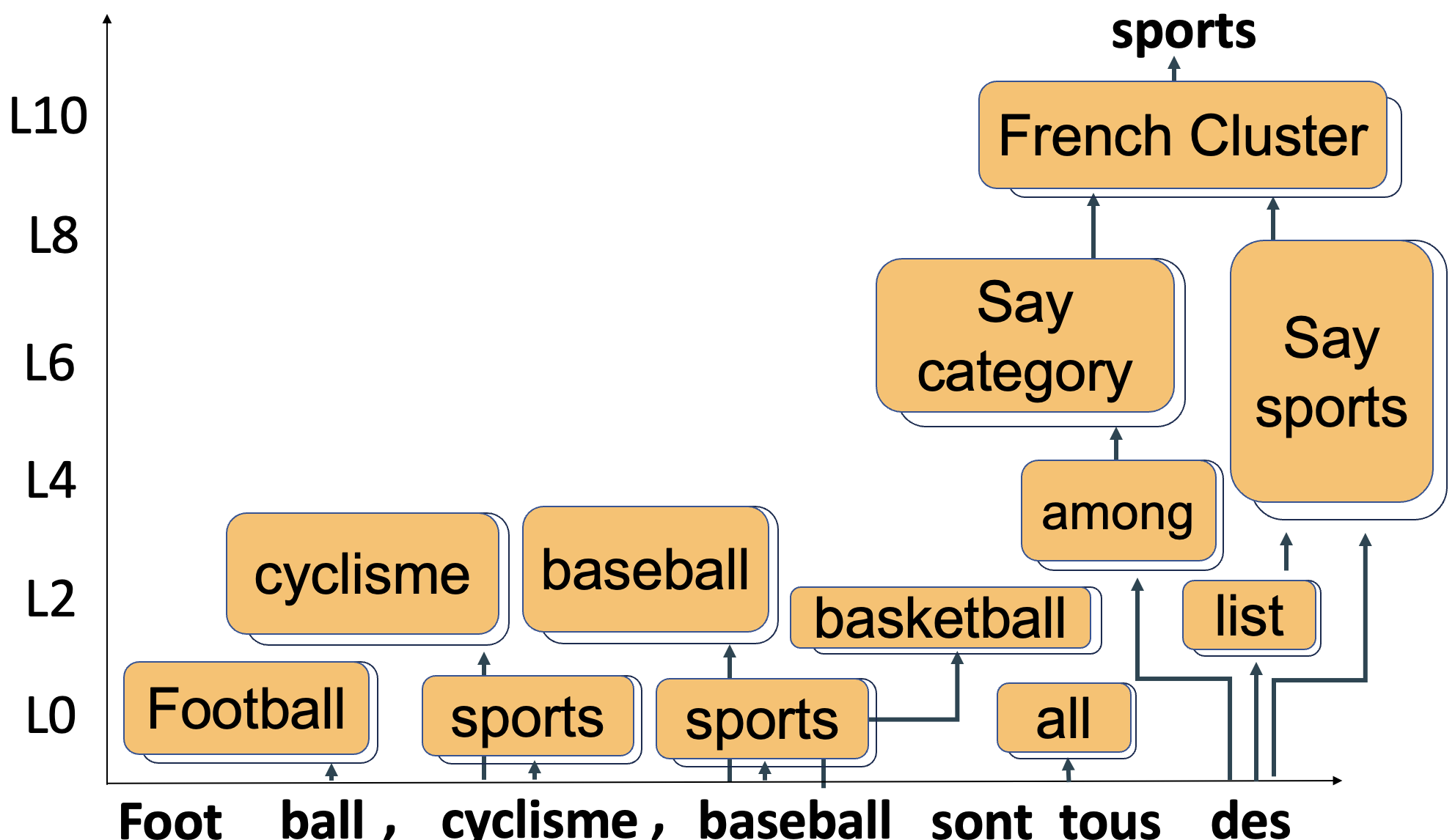}
        \caption{French}
    \end{subfigure}
    
    \vspace{0.3em}
    
    \begin{subfigure}[b]{0.49\textwidth}
        \centering
        \includegraphics[width=\textwidth,trim=0 0.5 0 0.5, clip]{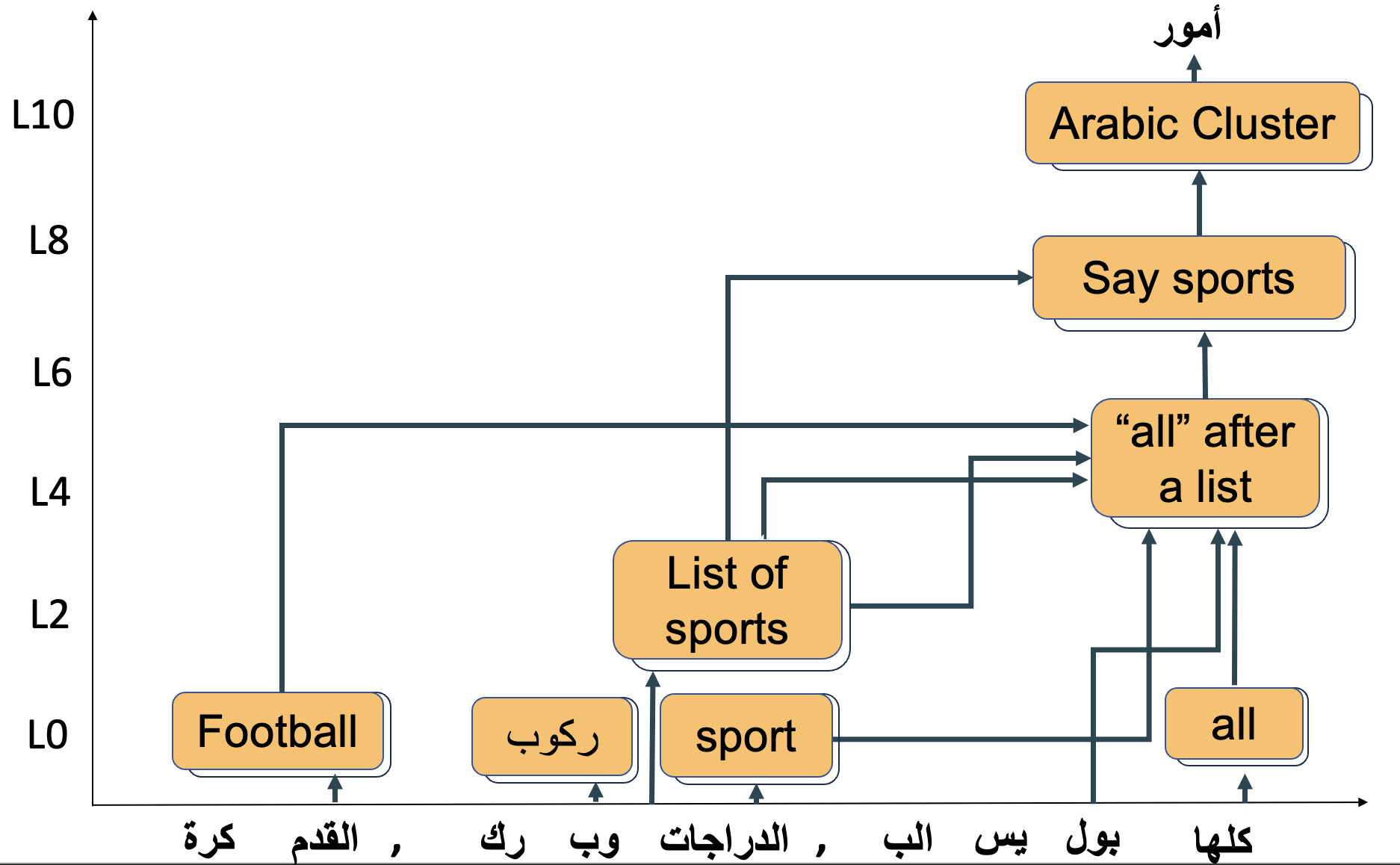}
        \caption{Arabic}
    \end{subfigure}
    \hfill
    \begin{subfigure}[b]{0.49\textwidth}
        \centering
        \includegraphics[width=\textwidth]{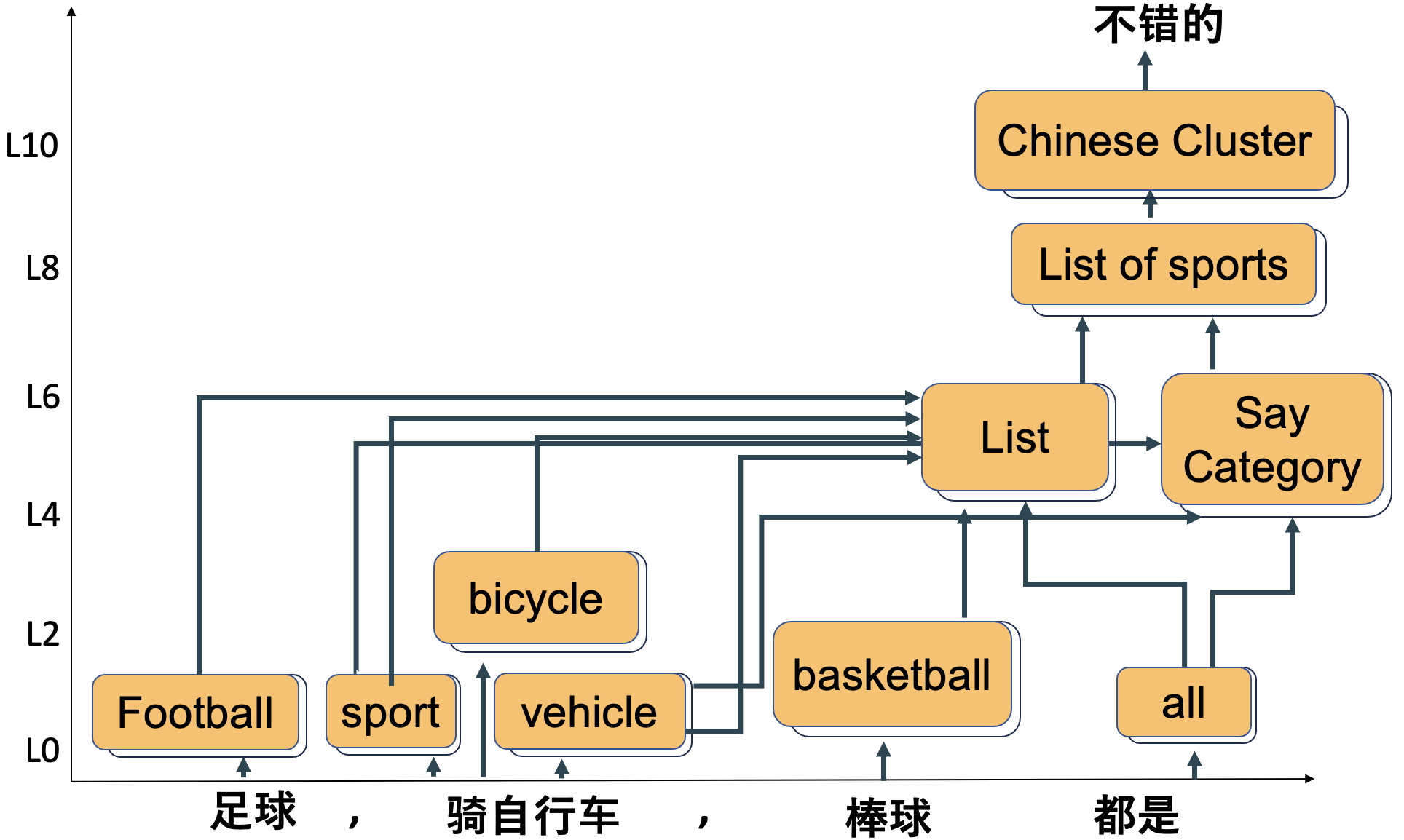}
        \caption{Chinese}
    \end{subfigure}
        \end{minipage}
}
    \caption{Circuits for category completion task.}
    \label{fig:circuits_calendar}
\end{figure*}

We also present the intervention results for Arabic in Figure~\ref{figure:intervention_category}, the only language where the model failed.

Comparatively, English consistently triggers stronger responses in task-specific circuits (\texttt{Men\&Women} for Antonym, \texttt{Say Sports} for Category Completion).  
This difference narrows under balanced mixtures but remains linked to sub-tokenization: languages with fragmented tokens show weaker edge strengths from embeddings to target clusters, partially explaining their reduced accuracy (Appendix~\ref{app:why_eng_better}).

We further validate these mechanisms through more complex tasks (translation and cultural 
prompts on LLaMA-1B), tokenizer analysis, and quantitative validation across the attribution graphs (see Appendix~\ref{app:complextasks}).

\subsection{Intervention Validation}
\label{app:intervention-validation}
We define an \textbf{intervention on a cluster of circuits} as steering the residual stream using a vector constructed from the cluster's features. 
Specifically, let $f_1, f_2, \dots, f_n$ denote the features in a cluster, and let $v(f_i)$ be the decoded vector of feature $f_i$. 
The intervention vector $v_\text{int}$ is then given by

\begin{equation}
    v_\text{int} = \alpha \sum_{i=1}^{n} v(f_i),
\end{equation}

where $\alpha$ is a scaling factor determined via sweeping to optimize the effect of the intervention. 
The residual stream is updated by adding $v_\text{int}$ at the relevant layer.

\begin{figure*}[h!]
    \centering
            \resizebox{0.8\textwidth}{!}{%
\begin{minipage}{\textwidth}
    \begin{subfigure}[b]{0.48\textwidth}
        \centering
        \includegraphics[width=\textwidth,trim=0 0.5 0 1, clip]{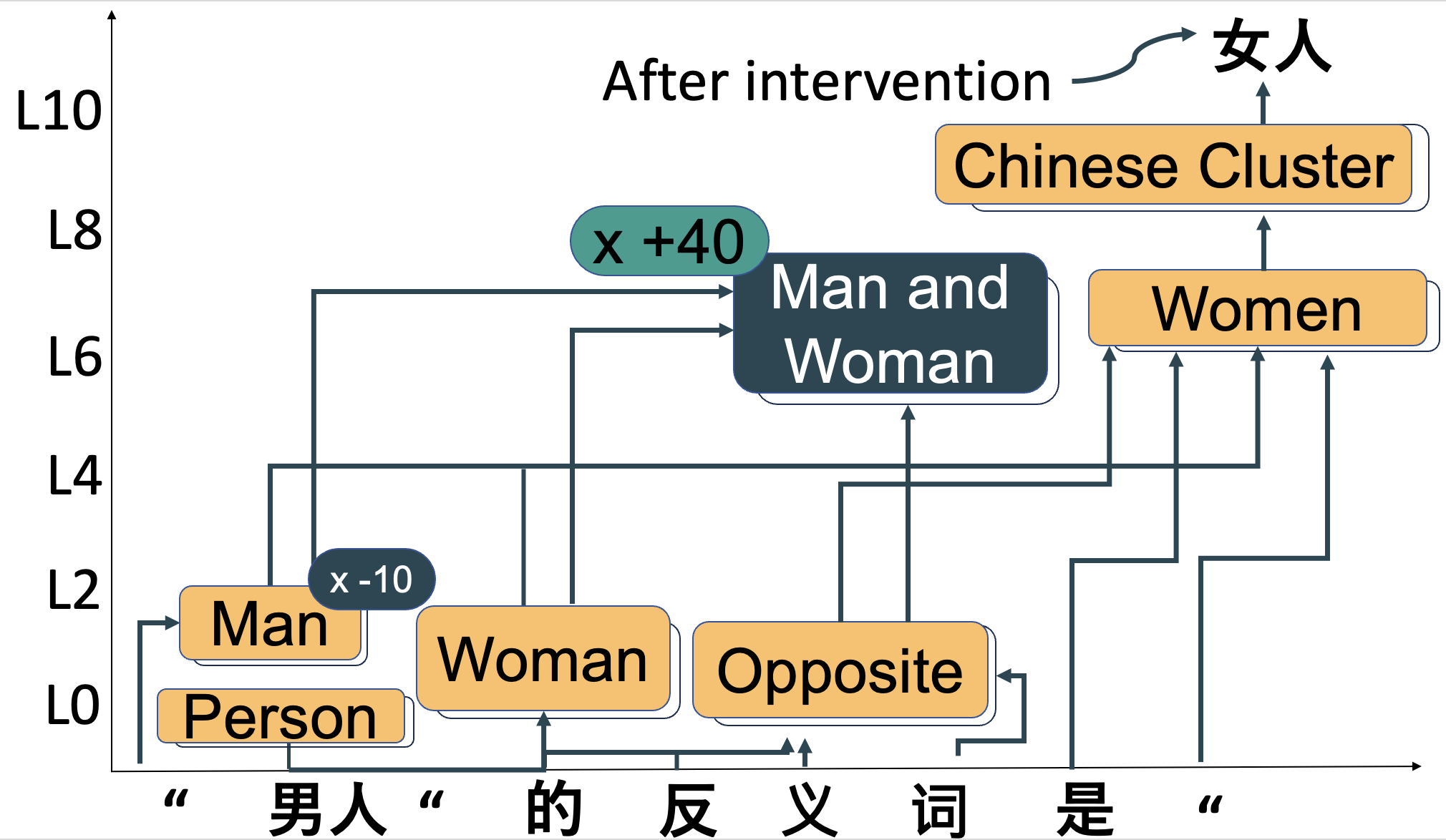}
        \caption{20\% mixture model intervention}
        \label{fig:antonym_de_intervene}
    \end{subfigure}
    \hfill
    \begin{subfigure}[b]{0.48\textwidth}
        \centering
        \includegraphics[width=\textwidth]{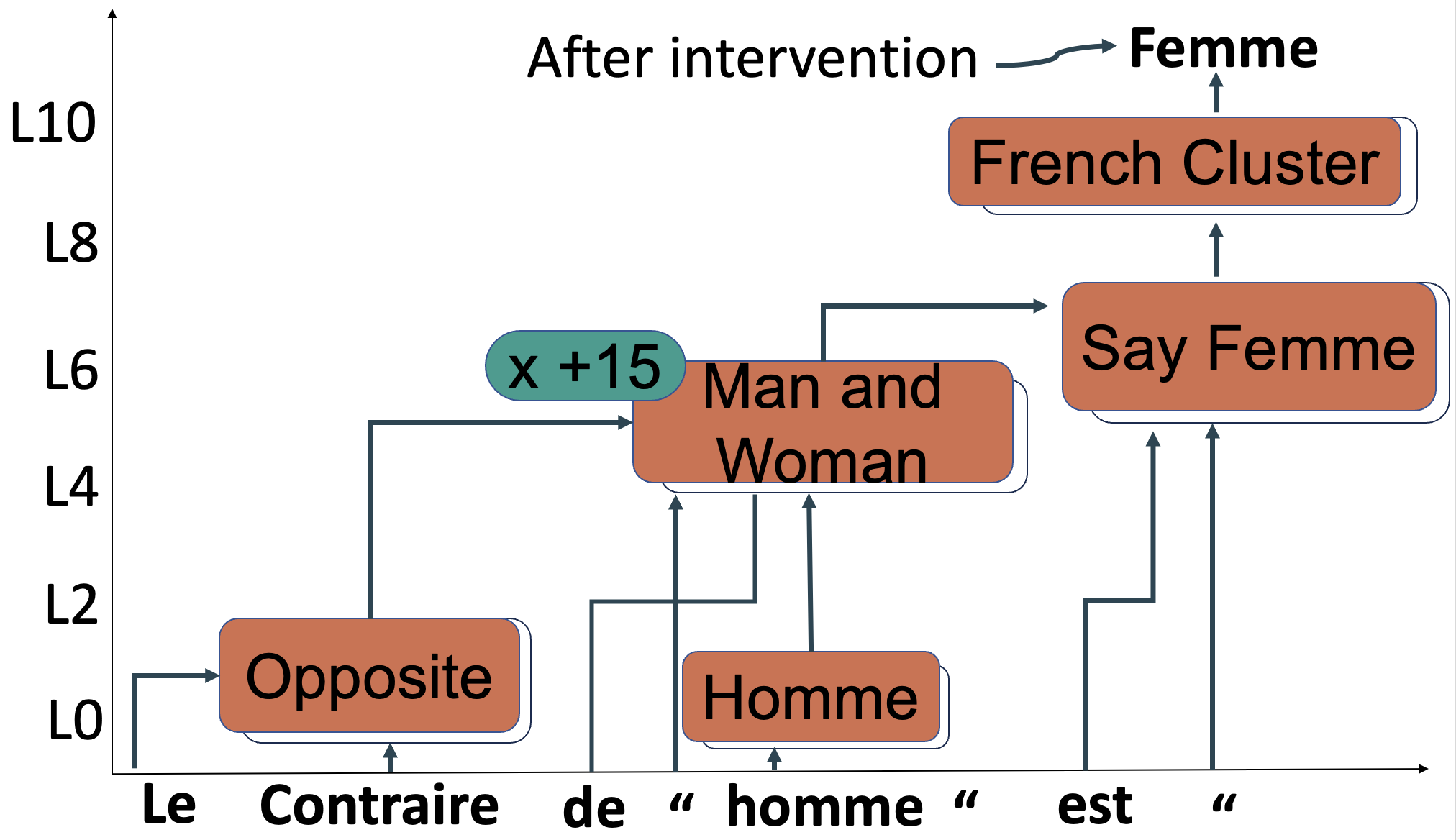}
        \caption{50\% mixture model intervention}
        \label{fig:antonym_ar_intervene}
    \end{subfigure}
    \hfill
    \begin{subfigure}[b]{0.48\textwidth}
        \centering
        \includegraphics[width=\textwidth]{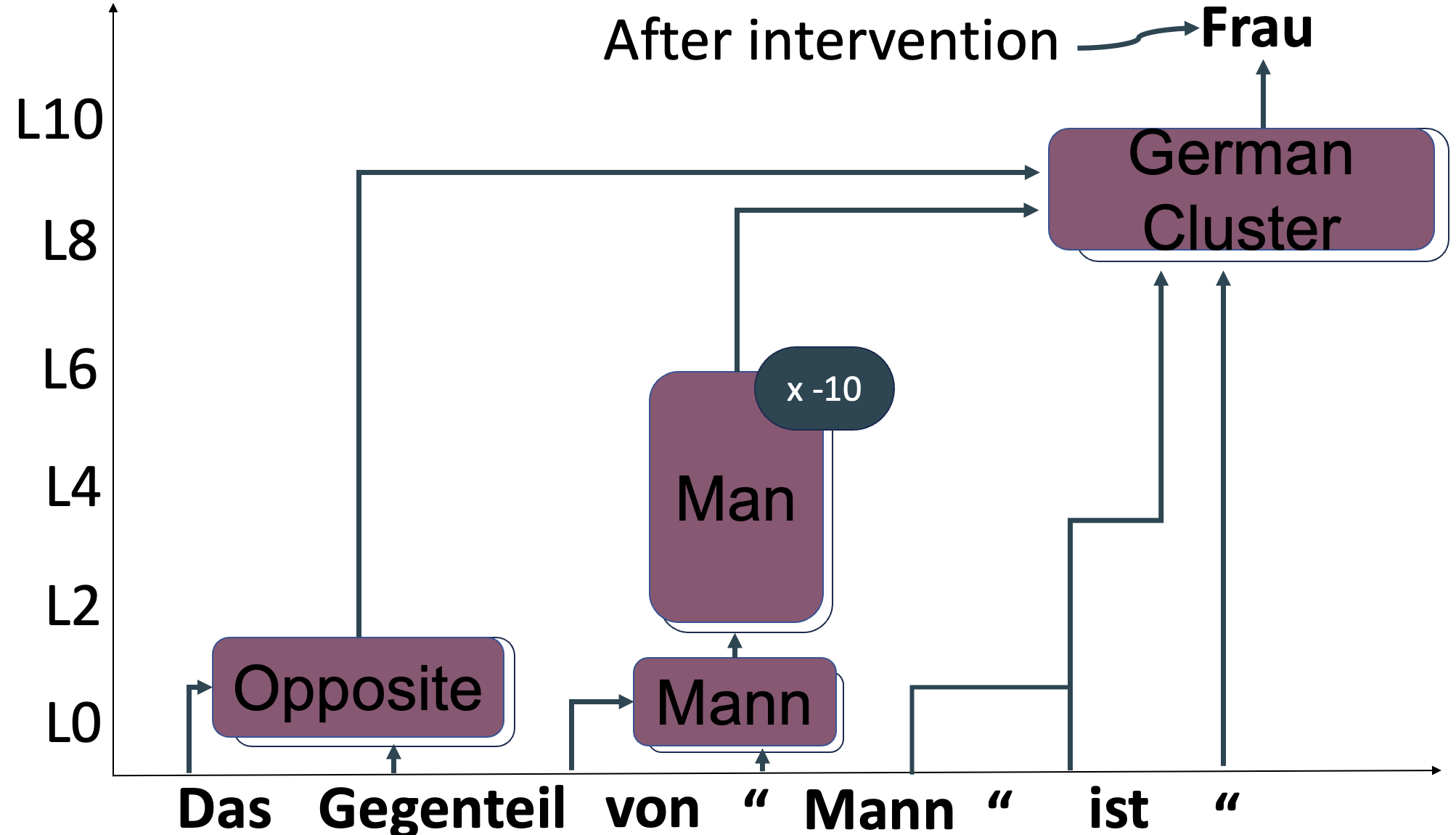}
        \caption{70\% mixture model intervention}
        \label{fig:antonym_ar_intervene}
    \end{subfigure}
    \hfill
    \begin{subfigure}[b]{0.48\textwidth}
        \centering
        \includegraphics[width=\textwidth,trim=0 0 0 1, clip]{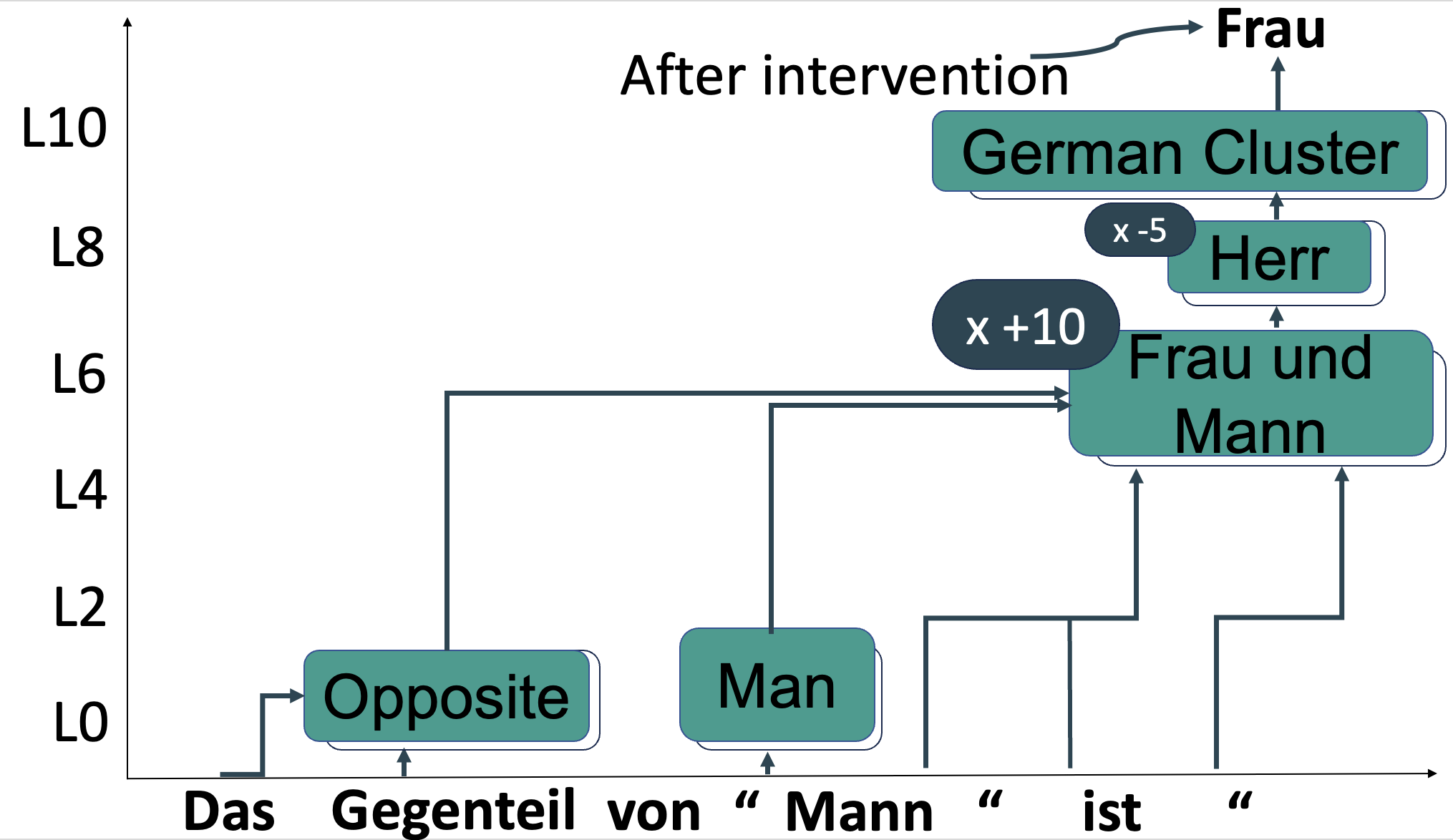}
        \caption{90\% mixture model intervention}
        \label{fig:antonym_ar_intervene}
    \end{subfigure}
    \end{minipage}
}
    \caption{Intervention examples where appropriate scaling switches the model's answer to the correct response.}
    \label{fig:intervention_figs}
\end{figure*}

\label{appendix:interventions}
Figures~\ref{fig:antonym_task}, \ref{fig:intervention_figs} and \ref{figure:intervention_category}   demonstrates successful interventions that validate the competing circuits hypothesis.

\textbf{In all the examples where the model fails to get the right answer}, scaling up the reasoning circuit while scaling down the copying circuit consistently produces correct predictions. We sweep positive coefficients for the reasoning circuit (1 to 30) and negative coefficients for the copying circuit (-30 to -1) to determine optimal steering values.

Figures ~\ref{fig:multiling_evidence_20}, ~\ref{fig:multiling_evidence_50}, ~\ref{fig:multiling_evidence_70}, and ~\ref{fig:multiling_evidence_90} further validate our hypothesis. In particular, most of the examples where the model fails to predict the correct answer show either an absence of the reasoning circuit or a low level of its activation, and this pattern becomes less pronounced as we move towards the 20\% mixture. Taken together with the intervention results, this supports our hypothesis about circuit competition.




\section{Extended Task Analysis: Translation and Cultural Prompts}
\label{app:complextasks}

We tested whether the shared multilingual space mechanisms generalize beyond simple lexical tasks (antonym, category completion) to more complex, context-dependent tasks on LLaMA-3.2-1B \cite{meta-llama32-1b}. Attribution analysis across all complex task circuits reveals the consistent three-phase structure: early layers extract language-specific features, middle layers form language-neutral semantic representations, and late layers activate language-specific output clusters.

\subsection{Translation Prompts}
\label{app:trans-prompts}
Translation tasks require cross-lingual semantic mapping without explicit language labels:
\begin{itemize}
    \item ``English: Flower, Fran\c{c}ais: \_\_''
    \item ``English: Flower, Arabic: \_\_''
\end{itemize}

\begin{figure*}[h]
    \centering
        \resizebox{0.8\textwidth}{!}{%
\begin{minipage}{\textwidth}
    \begin{subfigure}[b]{0.49\textwidth}
        \centering
        \includegraphics[width=\textwidth]{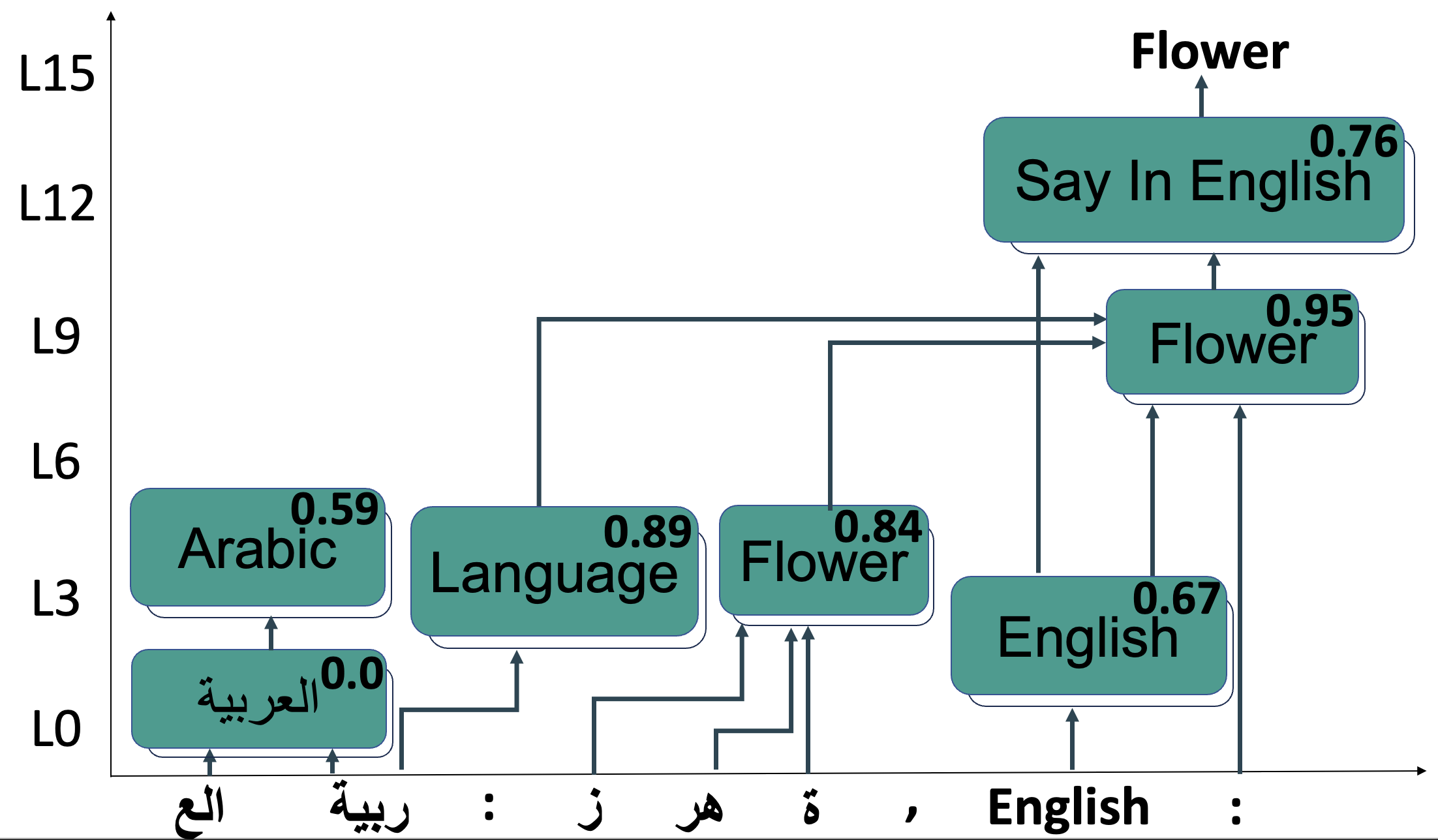}
        \caption{Arabic-English}
    \end{subfigure}
    \hfill
    \begin{subfigure}[b]{0.49\textwidth}
        \centering
        \includegraphics[width=\textwidth]{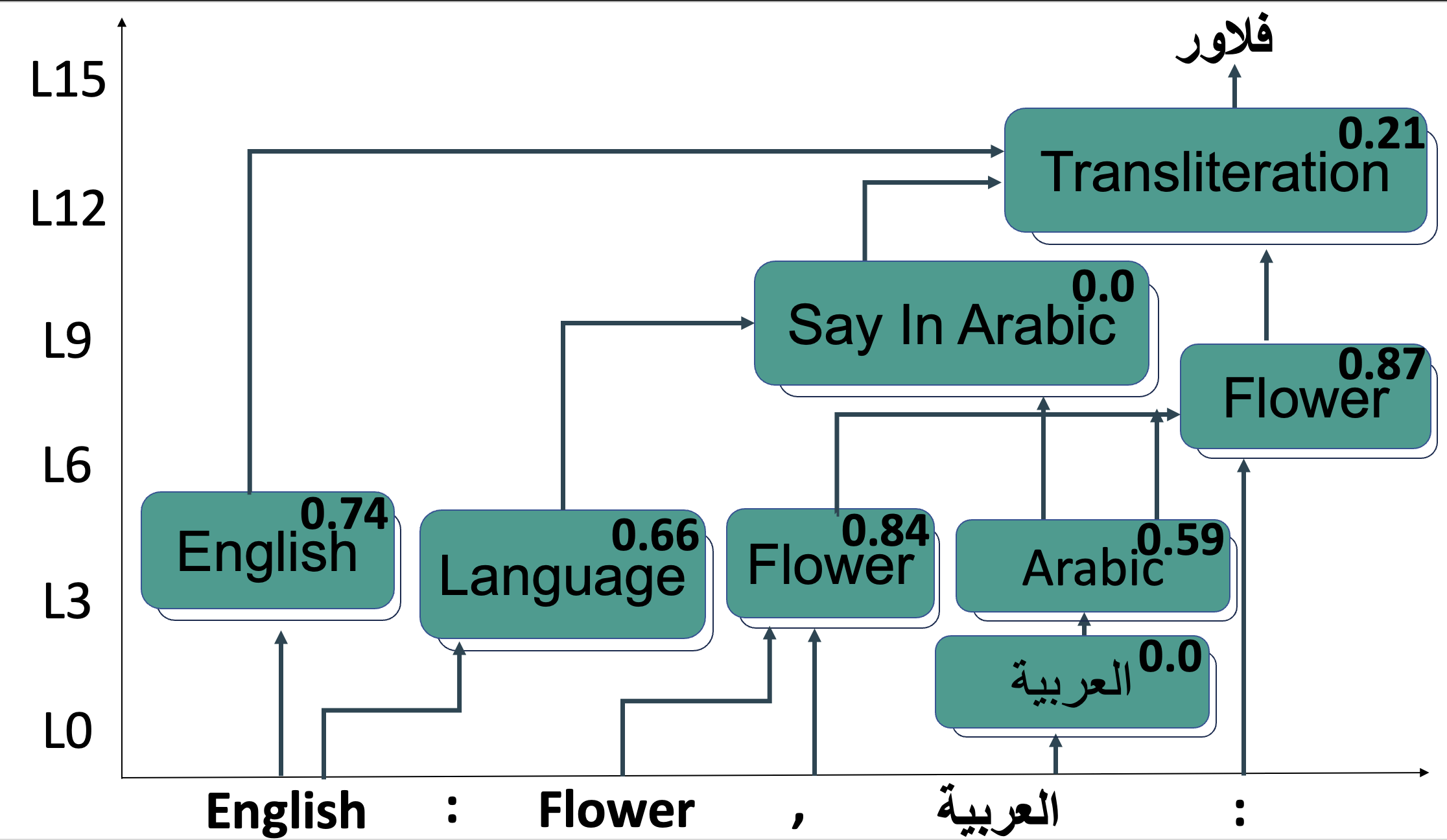}
        \caption{English-Arabic}
        \label{fig:trans-eng-ar}
    \end{subfigure}
    
    \vspace{0.3em}
    
    \begin{subfigure}[b]{0.49\textwidth}
        \centering
        \includegraphics[width=\textwidth,trim=0 1 0 1, clip]{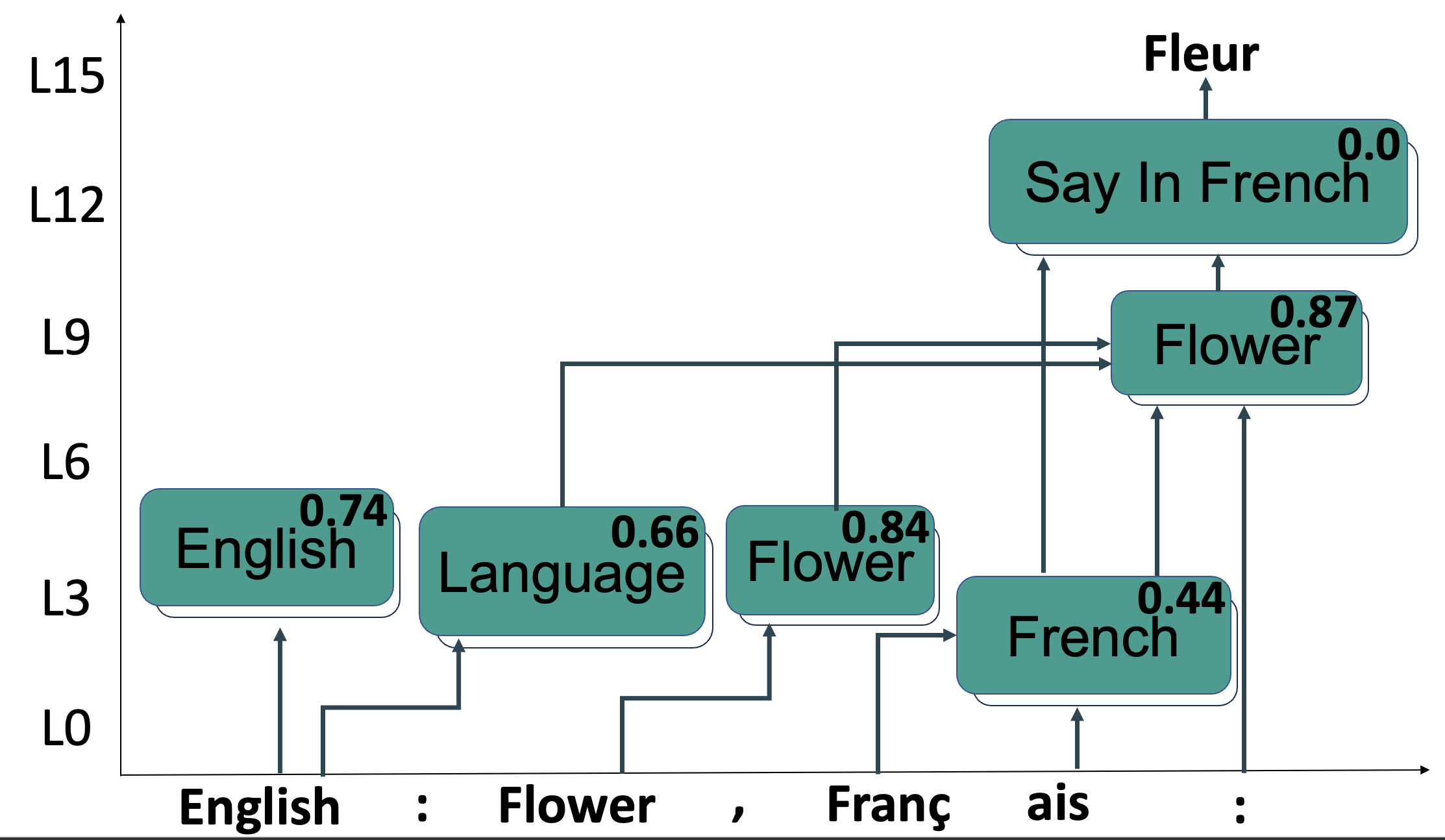}
        \caption{English-French}
    \end{subfigure}
\end{minipage}
}
    \caption{Circuits for the translation task.}
    \label{fig:circuits_seasons}
\end{figure*}

Circuits maintain the same rise-and-fall pattern across layers, despite task complexity.

\subsection{Cultural Context Prompts}
\label{app:culture-context-prompts}
Models complete: ``The best dish is...'' and generate language-appropriate food answers.

\begin{figure*}[h]
    \centering
    \begin{subfigure}[b]{0.49\textwidth}
        \centering
        \includegraphics[width=\textwidth]{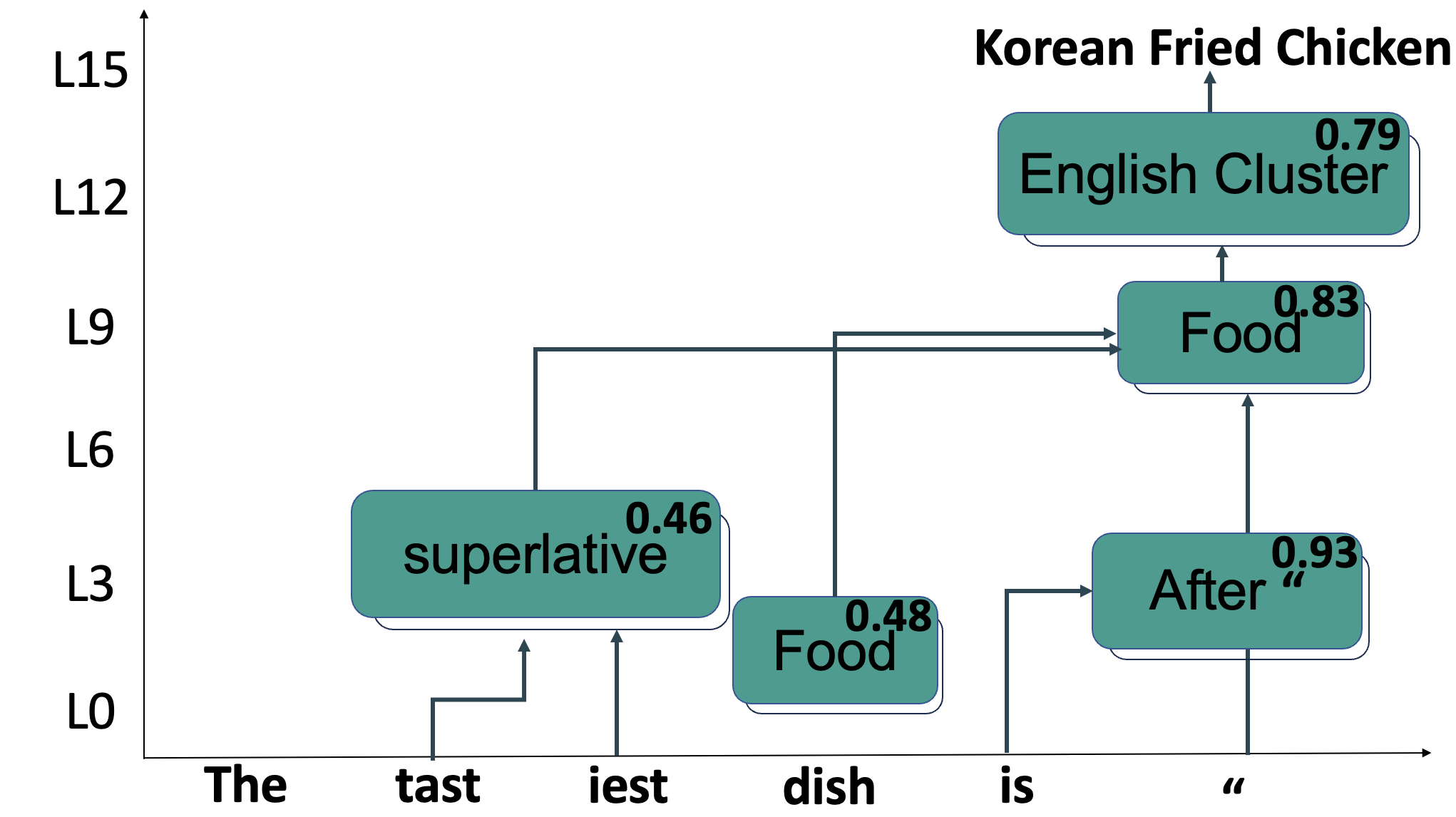}
        \caption{English Best Dish}
    \end{subfigure}
    \hfill
    \begin{subfigure}[b]{0.49\textwidth}
        \centering
        \includegraphics[width=\textwidth]{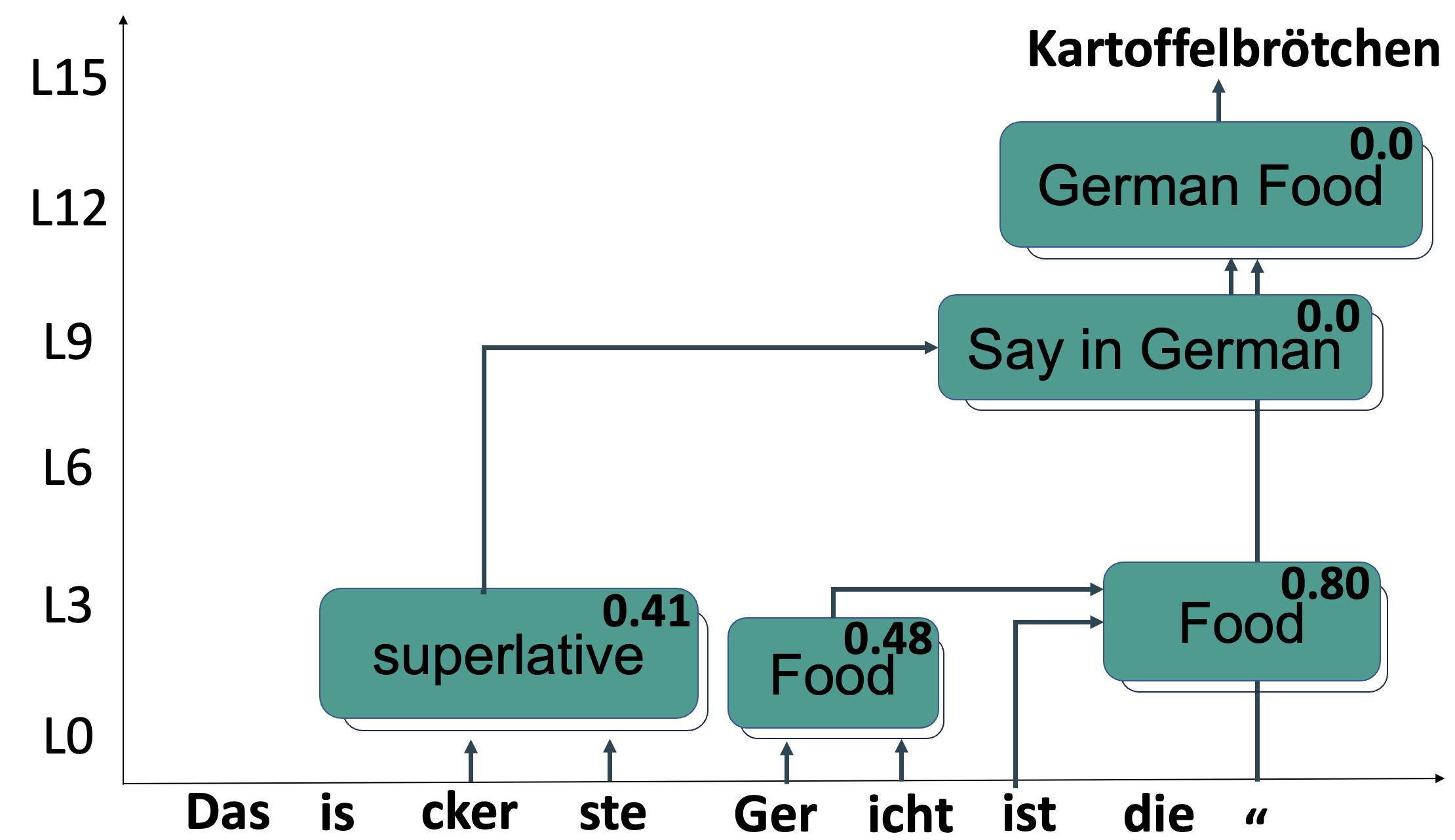}
        \caption{German Best Dish}
        \label{fig:trans-eng-ar}
    \end{subfigure}
    \caption{Circuits for the cultural task.}
    \label{fig:circuits_cultural}
\end{figure*}

The same multilinguality pattern emerges across these circuits (see Figure~\ref{fig:circuits_cultural}).

\section{Why English Performs Better: A Case Study}
\label{app:why_eng_better}

We present a case study examining the 20\% mixture to understand performance disparities in balanced training.

\subsection{Cluster Activation Analysis}
\label{app:eng-better-cluster-activ}
We measure activation strength of task-relevant clusters across languages. For the Antonym task, we track the \texttt{Man\&Woman} cluster; for Category Completion, the \texttt{Say Sports} cluster.

\begin{figure}
\centering
\includegraphics[width=0.6\linewidth, trim=0 0 0 0, clip]{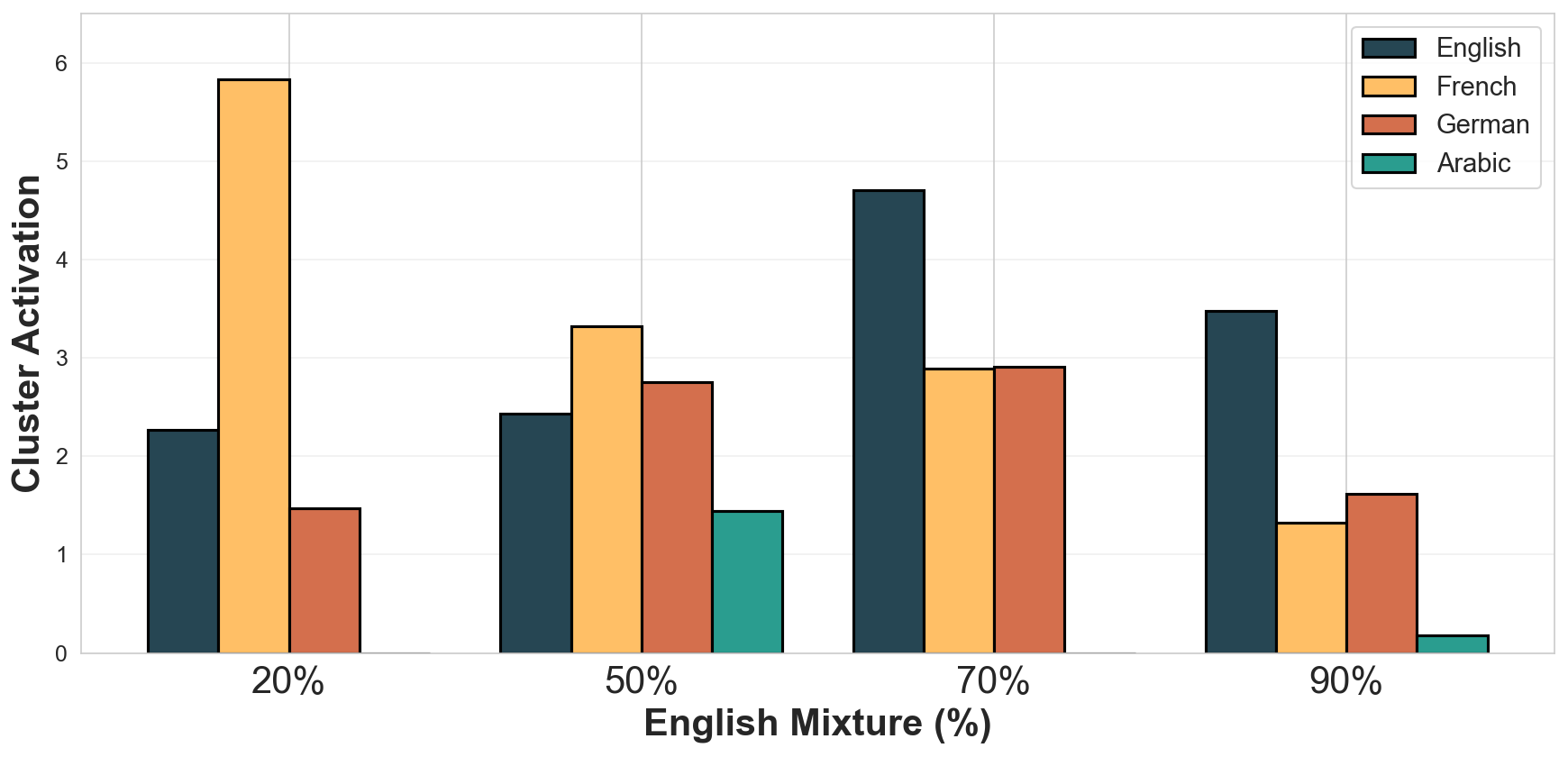}
\caption{Cluster activation strength across languages and mixture ratios. English shows consistently higher activations for task-relevant clusters (\texttt{Man\&Woman} for Antonym, \texttt{Say Sports} for Category Completion), with disparities diminishing at the 20\% mixture.}
\label{fig:cluster_activations}
\end{figure}

Figure~\ref{fig:cluster_activations} shows activation patterns. English consistently produces higher activations than other languages. This disparity decreases approaching the 20\% mixture, suggesting balanced training mitigates but does not eliminate the advantage.

\subsection{Sub-tokenization Effects}
\label{sec:subtokenization-effects}
We quantify sub-tokenization impact by measuring edge strength from token embeddings to target clusters. Edge strength is the dot product between the residual stream after token embedding and the cluster's input direction.

\begin{figure}
\centering
\includegraphics[width=0.5\linewidth, trim=0 0 0 0, clip]{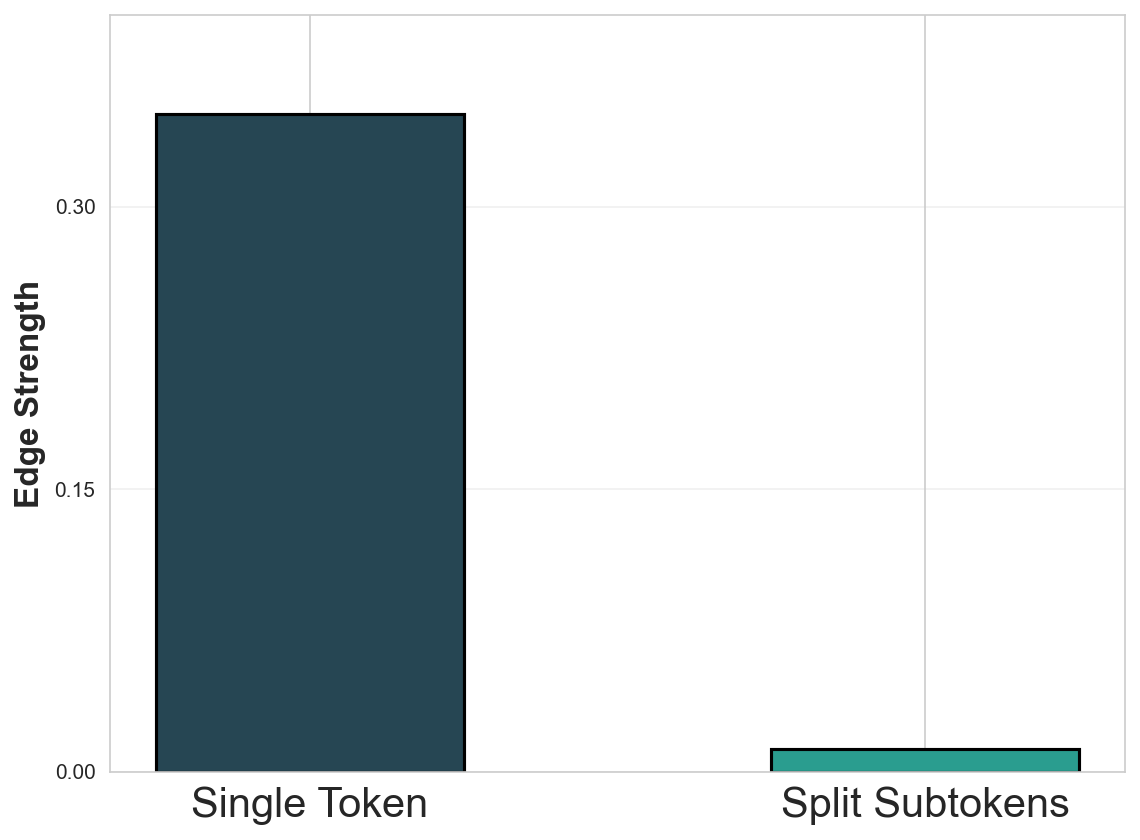}
\caption{Edge strength from token embeddings to target clusters versus sub-tokenization rate. Higher sub-tokenization correlates with weaker edge strength (r=-0.82), explaining reduced circuit activation in fragmented languages like Arabic.}
\label{fig:edge_strength}
\end{figure}

Figure~\ref{fig:edge_strength} presents results across languages with varying sub-tokenization rates. Highly sub-tokenized languages (e.g., Arabic) show significantly weaker edge strengths. Fragmented tokens fail to properly activate downstream circuits—semantic information distributes across multiple sub-tokens rather than concentrating in a single embedding.

\section{Model Diffing: Additional Material}
\label{app:model-diffing}
\subsection{Multilinguality of Repair Features}
\label{app:multiling-buckets}

\begin{figure}[t]
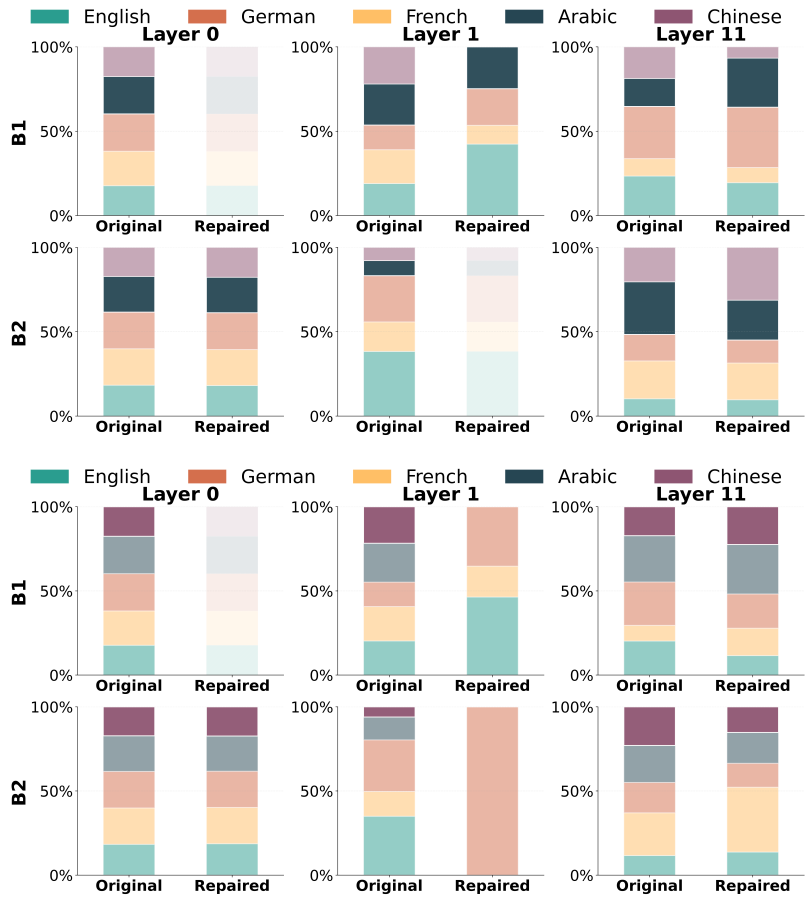

    \centering
    \resizebox{0.7\textwidth}{!}{%
\begin{minipage}{\textwidth}
    \begin{subfigure}{\linewidth}
        \centering
        \includegraphics[width=0.9\linewidth]{iclr2026/figures/fig2_lang_probs_ar_repair.png}
        \label{fig:a}
    \end{subfigure}

    \vspace{0.5em}

    \begin{subfigure}{\linewidth}
        \centering
        \includegraphics[width=0.9\linewidth]{iclr2026/figures/fig2_lang_probs_zh_repair.png}
        \label{fig:b}
    \end{subfigure}
    \end{minipage}
}
    \caption{Language distribution comparison across layers and buckets for \texttt{Arabic} (top) and \texttt{Chinese} (bottom).}
    \label{fig:bucket-figure}
\end{figure}

As mentionned in Section~\ref{subsec:feature_buckets}, we estimated the distribution of languages for features in each of our buckets. We evaluate on 10000 tokens and we report the scores in Figure~\ref{fig:bucket-figure}

The blurred bars indicate features for which the language distribution could not be reliably estimated. Although we use a large number of tokens, some features are rarely activated, resulting in insufficient samples to accurately characterize their language distribution.

\subsection{Token Assembly in Early Layers}
\label{app:causal_features}
\begin{figure}
    \centering
    \includegraphics[width=0.6\columnwidth, trim=0 0 0 1.5cm, clip]{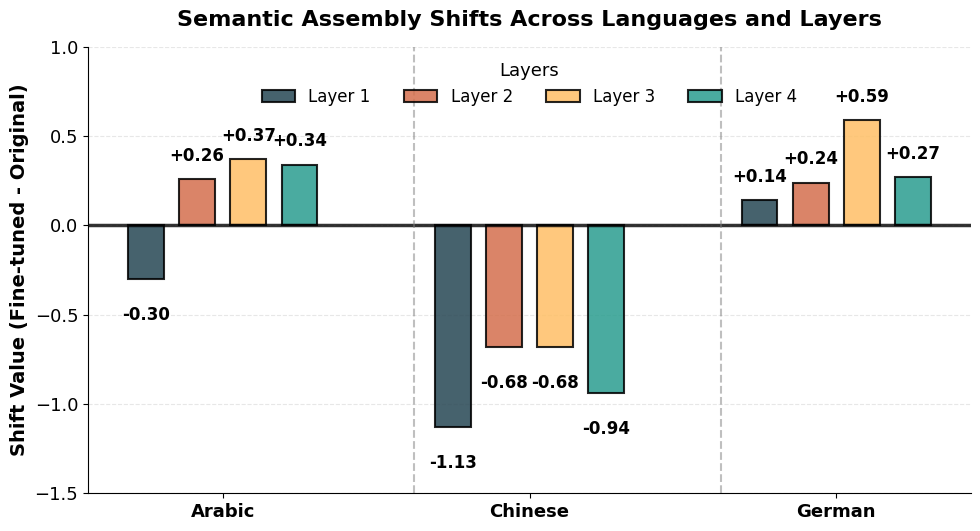}
    \caption{Logit Lens token assembly results. We find that models tend to repair early layers by making them better at assembling tokens, thus generalizing to abstractions earlier. This especially applies to \texttt{Arabic} and \texttt{German}. Surprisingly, \texttt{Chinese} finetuning results in less assembly across layers. This is in accordance with the higher-rank update of Chinese that shows that the model is possibly pattern-matching.}
    \label{fig:logitlens-features}
\end{figure}

Token assembly refers to the process by which early layers reconstruct sub-tokenized inputs (particularly fragmented words from morphologically rich languages like Arabic) into coherent higher-level representations before semantic processing.
Logit Lens results, as described in Section~\ref{subsec:causal_interventions}, are shown in Figure~\ref{fig:logitlens-features}. As observed, finetuning improves word assembly in \texttt{Arabic} and \texttt{German}. Counterintuitively, it increases fragmentation in \texttt{Chinese}, likely due to the comparatively high-rank updates required for \texttt{Chinese}. This suggests that the model relies more on memorization than on structured rules or heuristics for \texttt{Chinese}, as learning this language consumes a larger portion of the model’s capacity, leaving less capacity available for other languages.

\section{Tokenizer Analysis: BPE Constraints on Multilingual Processing}
\label{app:tokenizer}

Despite employing a debiased multilingual BPE tokenizer trained on balanced data, morphologically rich languages exhibit severe sub-tokenization compared to other languages. We isolate the source of this disparity through three complementary analyses.

\subsection{Vocabulary Allocation: Cannibalization Testing}
\label{app:vocab-allocation}
To rule out allocation bias, we computed fertility ratios (tokens per word) for language-specific tokenizers:

\begin{itemize}
    \item Arabic: 1.06
    \item Chinese: 1.03
    \item English: 0.99
    \item French: 0.96
    \item German: 1.00
\end{itemize}

All ratios cluster near 1.0, confirming that the debiased tokenizer allocates vocabulary fairly across languages. The disparity in fragmentation therefore stems from linguistic properties, not allocation bias.

\subsection{Morphological Coherence}
\label{app:morph-coherence}
Token-morpheme coherence (how well token boundaries align with morpheme boundaries) diverges dramatically:

\begin{table}[h]
\centering
\begin{tabular}{lcc}
\hline
Metric & Arabic & English \\
\hline
Token Fertility & 1.97 & 1.53 \\
Token-Morpheme Coherence & 0.10 & 0.42 \\
Meaningless Fragments (\%) & 89 & 58 \\
\hline
\end{tabular}
\caption{Arabic exhibits 4× lower morphological coherence than English despite fair vocabulary allocation. Meaningless fragments represent tokens that carry no semantic content.}
\end{table}

Concrete examples illustrate this fragmentation:
\begin{itemize}
    \item \textit{al-radar} (radar) fragments into [\textit{al-r}, \textit{ad}, \textit{ar}]: prefix, middle, and suffix separated
    \item \textit{tafaqad} (inspect) fragments into [\textit{taf}, \textit{qad}]: broken across morpheme boundaries with subtokens having no meaning.
\end{itemize}

English maintains coherence: ``reading'' $\rightarrow$ [``read'', ``ing''], preserving morpheme structure.

\subsection{Mechanistic Implications}
\label{app:tokenizer-mech-imp}
This tokenization constraint directly impacts early-layer processing. Fragmented tokens lack semantic content, forcing early layers (0--3) to first reconstruct word boundaries before processing meaning. This reconstruction cost weakens the initial activation patterns that propagate through multilingual circuits. 

Critically, despite this early-layer handicap, Arabic still develops identical shared multilingual representations in middle layers (5--6), demonstrating that the shared multilingual space mechanism is robust to tokenization disadvantage. However, the initial signal degradation explains why Arabic exhibits weaker circuit activation and lower downstream performance despite mechanistic equivalence.

\end{document}